\documentclass{article}

\PassOptionsToPackage{numbers, compress}{natbib}

 \usepackage[eandd, final]{neurips_2026}
\usepackage[utf8]{inputenc} %
\usepackage[T1]{fontenc}    %
\usepackage{hyperref}       %
\usepackage{url}            %
\usepackage{booktabs}       %
\usepackage{amsfonts}       %
\usepackage{nicefrac}       %
\usepackage{microtype}      %
\usepackage{xcolor}         %
\usepackage{caption}
\usepackage{graphicx}
\usepackage{booktabs}
\usepackage{pifont}
\usepackage{booktabs,tabularx}
\usepackage{multirow}
\usepackage{wrapfig}
\usepackage{amsmath}       %
\usepackage{xcolor}
\usepackage{amssymb} %
\usepackage{colortbl}
\newcommand{\cmark}{\textcolor{green!60!black}{\ding{51}}}
\newcommand{\xmark}{\textcolor{red}{\ding{55}}}

\renewcommand{\arraystretch}{0.8}
\newcommand{\upp}{$\uparrow$}
\newcommand{\downn}{$\downarrow$}
 \definecolor{metricaccuracy}{RGB}{105,56,135}
\definecolor{metriccontour}{RGB}{64,130,205}
\definecolor{metricregularity}{RGB}{210,70,70}
\definecolor{metriccolor}{RGB}{220,145,0}
\definecolor{metrictopology}{RGB}{0,155,145}

\definecolor{bestred}{RGB}{139,0,0}
\definecolor{secondblue}{RGB}{0,70,160}
\definecolor{thirdgreen}{RGB}{0,120,60}
\colorlet{aucaccuracy}{metricaccuracy!8}
\colorlet{auccontour}{metriccontour!8}
\colorlet{auccolor}{metriccolor!10}
\colorlet{aucregularity}{metricregularity!8}
\colorlet{auctopology}{metrictopology!9}

\newcommand{\SUPERVISE}{SUPERVISE}
\newcommand{\brbp}{\ensuremath{\sqrt{\mathrm{BR}\times\mathrm{BP}}}} %

\usepackage{tikz}
\title{SUPERVISE: A Unified Framework for Standardized and Reproducible Superpixel Evaluation}

\author{%
  Julien Walther \\
  Univ. Bordeaux, CNRS, Bordeaux INP,\\ IMS, UMR 5218, France\\
  \texttt{julien.walther@u-bordeaux.fr} \\
  \And
  Rémi Giraud \\
   Univ. Bordeaux, CNRS, Bordeaux INP,\\ IMS, UMR 5218, France \\
  \texttt{remi.giraud@u-bordeaux.fr} \\
  \AND
  Michaël Clément \\
  Univ. Bordeaux, CNRS, Bordeaux INP,\\ LaBRI, UMR 5800, France \\
  \texttt{michael.clement@labri.fr} \\
}

\begin{document}

\maketitle

\begin{abstract}
	Image segmentation into superpixels is a widely used technique in computer vision, with a large body of work developed over the years. Yet, its evaluation remains poorly standardized, as methods are compared under heterogeneous protocols, across inconsistent scale ranges, and with redundant or misaligned metric subsets.
	In this work, we introduce \textbf{SUPERVISE} (SUperpixel PERformance VISualization and Evaluation), a unified and reproducible benchmark that addresses these longstanding inconsistencies.
	Built on a scale-normalized protocol based on interpolation over the actual number of generated superpixels, SUPERVISE enables fair and consistent cross-method comparisons. We leverage this standardized framework to conduct the first large-scale statistical analysis of 20 evaluation metrics across 7 datasets and 31 methods, revealing strong inter-metric correlations and significant redundancy that challenge common evaluation practices. These findings provide empirical support  for reporting a compact set of representative metrics. SUPERVISE is released as a lightweight, modular framework built on precomputed shared label maps, making large-scale reproducible evaluation accessible to the community.
	All code, label maps, and evaluation results are publicly available at: \url{https://github.com/waldo-j/supervise}
\end{abstract}

\section{Introduction}

\begin{figure}[t]
	\centering
	\begin{minipage}[b]{0.54\linewidth}
		\centering
		\includegraphics[width=0.97\textwidth]{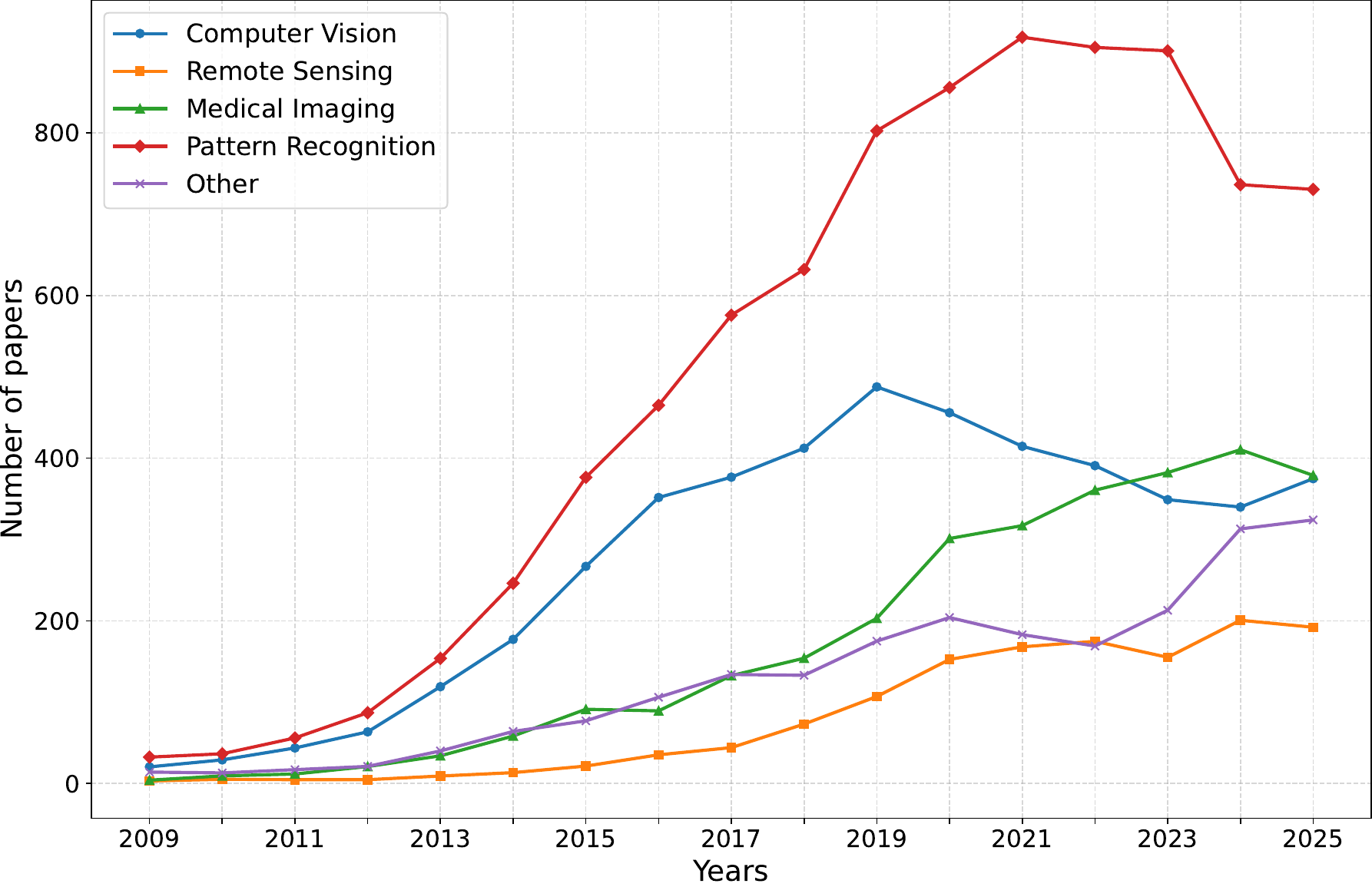}
	\end{minipage}
	\hfill
	\begin{minipage}[b]{0.43\textwidth}
		\centering
		{\scriptsize
			\begin{tabular}[b]{@{}c@{\hspace{2mm}}c@{}}
				\includegraphics[width=0.48\textwidth]{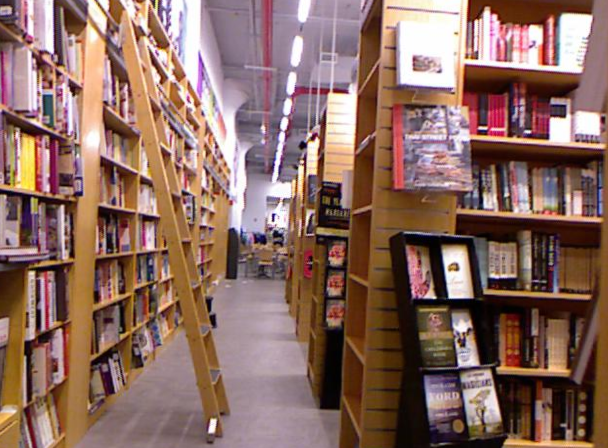}             &
				\includegraphics[width=0.48\textwidth]{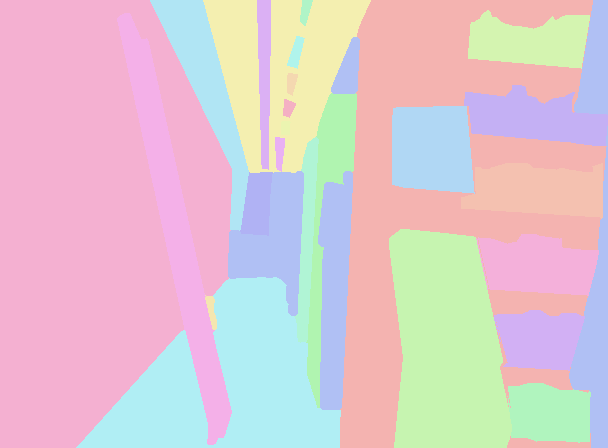}          \\[-0.5ex]
				a) Input image                                                                                 &
				b) Object groundtruth                                                                            \\[0.5ex]
				\includegraphics[width=0.48\textwidth]{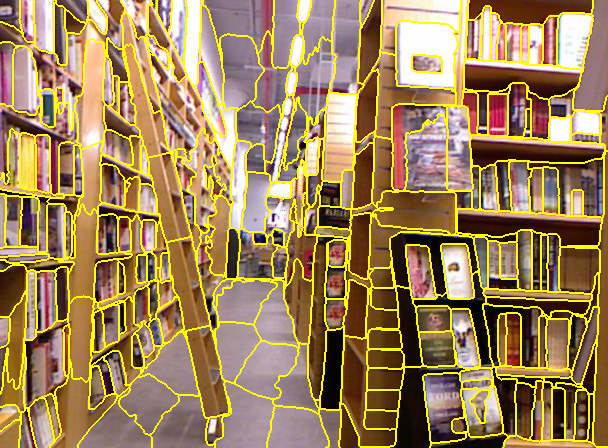} &
				\includegraphics[width=0.48\textwidth]{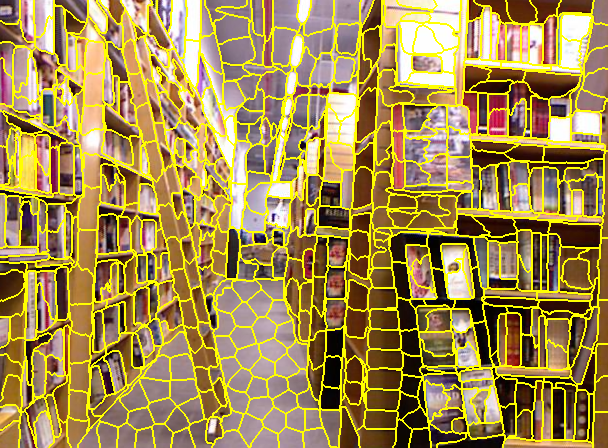}   \\[-0.5ex]
				c) Superpixels ($k=150$)                                                                       &
				d) Superpixels ($k=650$)                                                                         \\
			\end{tabular}}
	\end{minipage}
	\par\nointerlineskip\vspace{7pt}
	\begin{minipage}[t]{0.54\linewidth}
		\centering
		\captionof{figure}{\textbf{Domains of superpixel research papers.}
			Average number of superpixel-related research papers per domain, computed using the OpenAlex API~\cite{openalex}}
		\label{fig:occurrences}
	\end{minipage}
	\hfill
	\begin{minipage}[t]{0.43\textwidth}
		\centering
		\caption{
			\textbf{Superpixel segmentation} is
			a fine over-segmentation aiming to capture the borders of image objects and subparts.
		}%
		\label{fig:superpixel_decomposition}
	\end{minipage}
	\vspace{-0.4cm}
\end{figure}

Superpixels have been a central representation
in pattern recognition and computer vision for over two decades (Fig.~\ref{fig:occurrences}), providing an efficient over-segmentation that reduces image complexity while preserving local structure.
As such,
they have played a key role %
as an oversegmentation for annotation purposes
\cite{long2018efficient,berg2019ilastik}, or as
structural primitives for many tasks such as
stereo matching \cite{li2016pmsc},
optical flow \cite{liu2019selflow}
or saliency estimation \cite{wang2015saliency}.
With the advent of end-to-end deep learning, 
hierarchical features could be learned directly from raw pixels,
causing the use of superpixels to evolve. 
They remain particularly valuable
in high-resolution domains such as medical imaging and remote sensing,
where dense processing is computationally expensive.
Their recent use as adaptive tokenization mechanisms in transformer-based architectures
for representation learning~\cite{ke2024cast,lew2024superpixel},
semantic segmentation~\cite{mei2025spformer},
hierarchical image classification~\cite{park2025},
tokenization~\cite{aasan2025spit},
image restoration~\cite{aasan2025dht},
or
super-resolution~\cite{zhang2023spin}
further highlights their continued relevance,
notably in conjunction with foundation models~\cite{lyu2024superpixel,yang2024foundation}.%

Superpixels
can be seen as an irregular, content-aware alternative to a regular grid,
capturing fine structures, object subparts, or any semantically agnostic content, while reducing the number of primitives.
This dual objective naturally induces specific evaluation criteria including:
object adherence, contour quality, regularity, color homogeneity, and topology.
Because these objectives are partly conflicting, superpixel segmentation is inherently governed by trade-offs: improving contour adherence often leads to irregular regions
and vice versa.
Hence, no decomposition is uniquely optimal, and evaluation is fundamentally multi-criteria, unlike standard semantic  segmentation.
Meaningful comparison requires jointly considering complementary properties spanning the objective space %
\cite{stutz2018}.

However, in practice, some existing works report very heterogeneous subsets of metrics drawn from a large and evolving pool, which may lead to inconsistent and insufficient evaluations
missing some superpixel properties
\cite{giraud2024}.
Moreover, several commonly used metrics appear to be strongly correlated and may provide little additional information \cite{giraud2017,stutz2018}.
This may create an appearance of comprehensive evaluation while obscuring the few underlying dimensions that truly differentiate methods.
Another difficulty in the evaluation process
arises from the multi-scale nature of superpixels.
As illustrated in Fig.~\ref{fig:superpixel_decomposition},
unlike most segmentation tasks,
superpixels do not correspond to a single final partition, but to a family of potential decompositions at different scales,
making fair evaluation particularly sensitive to how scale is handled to avoid bias.
However, many evaluations and benchmarks report metric values for fixed requested numbers of superpixels, even though the actual number produced
can vary substantially between methods and affect performance \cite{giraud2024}.

\begin{wrapfigure}{r}{0.44\textwidth}
	\vspace{-0.45cm}
	\centering
	\captionof{table}{\textbf{Variance in reported results.} Like other deterministic metrics,
		the reported achievable segmentation accuracy (ASA) may vary across papers, although methods share code, default settings and weights.
	}%
	{\scriptsize
		\begin{tabular}{lcc}
			                            & \multicolumn{2}{@{\hspace{0mm}}c@{\hspace{0mm}}}{\textbf{Reported ASA 500 SP for}}                            \\
			\textbf{Papers}             & {SLIC \cite{achanta2012}}                                                          & {SSN~\cite{jampani2018}} \\
			\midrule
			SSN~\cite{jampani2018}      & 0.957                                                                              & 0.969                    \\
			AINET~\cite{wang2021ainet}  & 0.946                                                                              & 0.962                    \\
			CDS~\cite{xu2024cds}        & 0.957                                                                              & 0.965                    \\
			SPAM~\cite{walther2025spam} & 0.959                                                                              & 0.967                    \\
		\end{tabular}
	}
	\label{tab:bias}
\end{wrapfigure}

For these reasons,
despite the maturity of the field and existing open-source benchmarks, the evaluation of superpixel methods remains poorly standardized.
Overall, performance is reported under heterogeneous conditions: across different scale ranges, datasets, and subsets of metrics.
Moreover, unlike many vision tasks where results are reported from fixed outputs,
superpixel methods are often re-run to provide results for each scale,
leading to potential inconsistencies.
As shown in Tab.~\ref{tab:bias}, even classical %
algorithms such as SLIC~\cite{achanta2012} or deep learning-based methods such as SSN~\cite{jampani2018} are reported with notably different segmentation accuracy across papers,
exposing a systematic comparison bias that cannot be attributed to the method but instead to implementation differences, parameters, requested scales, or reporting protocol.

In this work, we introduce SUPERVISE (SUperpixel PERformance VISualization and Evaluation),
an open-source, unified and reproducible framework for standardized superpixel benchmarking.
We formalize superpixel evaluation as a two-dimensional problem governed by scale and metric selection, reflecting the ill-posed nature of the task~\cite{giraud2024}.
SUPERVISE addresses these two axes through a scale-normalized protocol based on interpolation over the actual number of generated superpixels, and through a large-scale analysis of evaluation metrics across methods and datasets.
This analysis reveals a structured and partially redundant metric space, in which several metrics capture closely related behaviors while complementary evaluation dimensions remain.

Furthermore, by decoupling method execution from metric computation,
via precomputed label maps,
SUPERVISE provides a lightweight, extensible, and verifiable benchmarking pipeline.

\textbf{Contributions.} Our main contributions are as follows: \\[-4ex]
\begin{itemize}
	\setlength\itemsep{0.05em}
	\item We identify and formalize key limitations of existing superpixel evaluation protocols, including scale bias, metric redundancy, protocol heterogeneity, and limited visualizations, and relate them to the ill-posed and multi-objective nature of the task.

	\item We propose SUPERVISE, a unified and reproducible benchmarking framework built on a scale-normalized evaluation protocol based on interpolation over the actual number of generated superpixels, enabling fair and consistent comparison across methods.

	\item
	      To demonstrate the capabilities of SUPERVISE, we conduct a large-scale benchmark covering 31 superpixel methods, 20 evaluation metrics, and 7 datasets,
	      and publicly release all label maps and results to foster reproducible comparison.

	\item We perform a comprehensive statistical analysis of the evaluation space, revealing strong inter-metric and cross-dataset correlations that point to an intrinsic low-dimensional structure. From this, we identify compact and complementary subsets of metrics and datasets for more efficient evaluation.
\end{itemize}

\section{Related Work}

\paragraph{Superpixel methods.}
Superpixel segmentation has produced a large and diverse literature.
Early algorithmic approaches were mainly based on low-level image cues and can be
broadly grouped into graph-based, clustering-based, energy-based,
watershed-based, and hierarchical methods.
Representative examples include
SLIC~\cite{achanta2012},
SEEDS~\cite{vandenbergh2012}, ERS~\cite{liu2011}, LSC~\cite{chen2017},
and hierarchical or tree-based decompositions~\cite{wei2018sh}.
More recent work has extended this landscape with learning-based and
foundation-model-based superpixel methods.
Deep approaches such as
SSN~\cite{jampani2018}, SFCN~\cite{yang2020},
AINET~\cite{wang2021ainet}, and CDS~\cite{xu2024cds} rely on learned
features or differentiable formulations to improve alignment with image
structures and object boundaries, generally at the expense of regularity.

More recently, methods such as SPAM~\cite{walther2025spam} have leveraged segmentation foundation models to guide superpixel generation, highlighting the complementarity between high-level semantic priors and low-level representations.
Foundation models provide strong object-level cues but remain biased toward specific scales and categories, while superpixels offer a flexible and fine-grained decomposition, enabling the capture of smaller or less semantically defined regions.
Hierarchical methods, \emph{e.g.}, \cite{belem2022fast,yan2022hierarchical,xie2025hierarchical,walther2026hspam}, emphasize multi-scale capacity by providing perfectly nested superpixels across multiple scales, which may be a topological feature of interest for some downstream tasks.

\smallskip
\textbf{Evaluation metrics.}
The multi-objective nature of superpixel decomposition has led to a large and diverse set of evaluation metrics, each probing a different dimension of the solution space. These metrics are commonly grouped into five main categories: object segmentation, contour adherence, regularity, color homogeneity, and topology.
For instance, contour adherence can be assessed using boundary precision-recall measures, while regularity is often characterized through shape-based criteria such as circularity or convexity.
Accordingly, evaluation metrics aim to capture these complementary properties and are commonly assumed to be predictive of downstream task performance.
The comprehensive list of considered metrics is provided in Appendix~\ref{sec:appendix_metrics}.

However, reporting a large number of metrics does not necessarily yield more informative evaluation, and metric selection can significantly affect method ranking.
Many metrics are indeed correlated, partially redundant or lack robustness, leading to biased assessments \cite{stutz2018,giraud2024,wu2026survey}.
Moreover, they emphasize different and often conflicting properties: boundary-based metrics may favor fragmented decompositions, while regularity-based ones penalize accurate contour adherence or color homogeneity.
Existing evaluations often provide only a partial view of this objective space, relying on subsets of metrics that are redundant or misaligned with the underlying trade-offs.
	{Similar metric-selection issues were studied in medical image segmentation by~\cite{TA2015}, who analyzed metric properties and ranking correlations to identify non-redundant criteria.}

\begin{table*}[t]
	\centering
	\caption{\textbf{Comparison of superpixel benchmark frameworks.}
		We report here the main benchmarks that evaluate state-of-the-art superpixel methods, and also provide evaluation codes.
		We indicate code properties and reported results
		in the paper:  \cmark \text{ }, \xmark \text{ } and {\color{orange}$\sim$} respectively indicate that the feature is present, absent or partially addressed.
		(?) indicates that the code is not provided or no longer active.
		SUPERVISE is designed as a collaborative and evolving framework, intended to be regularly updated to incorporate new metrics, methods, and datasets (+).
	}
	\label{tab:benchmarks}
	\resizebox{\textwidth}{!}{%
		\begin{tabular}{l@{\hspace{0mm}}llcccccc}
			       &                             &
			       & \textbf{Neubert \& Protzel} & \textbf{Wang et al.}                       & \textbf{Stutz et al.}   & \textbf{Barcelos et al.} & \textbf{Wu et al.}     & \textbf{SUPERVISE}                                                         \\
			       &                             &                                            & 2012 \cite{neubert2012} & 2017 \cite{wang2017}     & 2018 \cite{stutz2018}  & 2024 \cite{barcelos2024} & 2026 \cite{wu2026survey} & 2026 \textbf{(Ours)} \\
			\midrule
			\multicolumn{2}{c}{\multirow{9}{*}{\rotatebox{90}{Code properties}}}
			       & Open-source code            &
			\cmark & Not active                  & \cmark                                     & \cmark                  & \xmark                   & \cmark                                                                                              \\
			       &                             & Programming language                       & {Matlab}                & {Matlab}                 & {C/C++}                & C/Bash                   & ?                        & Python               \\
			       &                             & Number of metrics                          & 11                      & 13                       & 5                      & 5                        & 5                        & 20+                  \\
			       &                             & Unified evaluation protocol                & \xmark                  & \xmark                   & {\color{orange}$\sim$} &
			{\color{orange}$\sim$}
			       & \xmark                      & \cmark                                                                                                                                                                                                \\
			       &                             & Correction of disconnected components      & \xmark                  & ?                        & \cmark                 & \cmark                   & ?                        & \cmark               \\
			       &                             & Interpolation and AUC over actual SP count & \xmark                  & ?                        & \xmark                 & \xmark                   & ?                        & \cmark               \\
			       &                             & Cross-metric robust visualizations         & \xmark                  & \xmark                   & \xmark                 & \xmark                   & \xmark                   & \cmark               \\
			       &                             & Efficient compressed processing            & \xmark                  & ?                        & \xmark                 & \xmark                   & ?                        & \cmark               \\
			       &                             & Multi-threading                            & \xmark                  & ?                        & \xmark                 & \xmark                   & ?                        & \cmark               \\
			\midrule
			\multicolumn{2}{c}{\multirow{6}{*}{\rotatebox[origin=c]{90}{\begin{tabular}{@{}c@{}}Reported\\results\end{tabular}}}}
			       & Number of datasets          & 3                                          & 1                       & 5                        & 5                      & 2                        & 7+                                              \\
			       &                             & Number of methods                          & 14                      & 15                       & 30                     & 24                       & 25                       & 31+                  \\
			       &                             & Inclusion of deep learning-based methods   & \xmark                  & \xmark                   & \xmark                 & \cmark                   & \cmark                   & \cmark               \\
			       &                             & Shared precomputed results                 & \cmark                  & \xmark                   & \xmark                 & \xmark                   & \xmark                   & \cmark               \\
			       &                             & Shared precomputed label maps              & \xmark                  & \xmark                   & \xmark                 & \xmark                   & \xmark                   & \cmark               \\
			       &                             & Statistical analysis                       & \xmark                  & \xmark                   & \cmark                  & \xmark                   & \xmark                   & \cmark               \\
			\bottomrule
		\end{tabular}%
	}%
\end{table*}

\smallskip
\textbf{Benchmarks and evaluation protocols.}
The need for common evaluation tools has long been recognized in the
superpixel literature.
In Tab.~\ref{tab:benchmarks}, we
report the properties of the main superpixel benchmarks,
mostly provided with open source codes.
One of the first dedicated superpixel benchmarks
combined datasets, error metrics, usage guidelines, and a Matlab toolbox~\cite{neubert2012}.
A later integrated benchmark library (no longer active) compared
several methods with a broader collection of metrics, including measures of
regularity~\cite{wang2017}.
This benchmarking effort was then expanded to a
large-scale evaluation covering 30 methods on five datasets, with attention
to fair parameter optimization, strict connectivity enforcement, %
implementation details, visual quality,
and a more critical assessment of the state-of-the-art metrics~\cite{stutz2018}.
More recent surveys such as \cite{barcelos2024, wu2026survey} proposed a clear taxonomy, grouping methods into methodological families, and
evaluating them according to criteria such as connectivity,
control over the number of superpixels, color
homogeneity,
and visual quality.
In \cite{wu2026survey}, the authors proposed a short review of superpixel-based applications.
These open-source codes aim to unify evaluation by also providing the source codes of methods.
Yet, these methods are implemented
in heterogeneous environments
and need to be re-run for new validation.

These benchmarks have progressively introduced strategies to improve the reproducibility and comparability of superpixel evaluation. Nevertheless, several key limitations remain.
First, most benchmarks rely on a limited subset of metrics, often without assessing their potential biases or redundancy.
Second, evaluation is typically performed at fixed nominal operating points, leading to scale-dependent biases and hindering fair comparison across methods.
Third, reproducibility remains challenging, as methods implemented in very heterogeneous environments must be re-run for evaluation.
Finally, the choice of both metrics and datasets is rarely questioned: datasets are often inconsistent, and little effort is devoted to systematically assessing their complementarity or representativeness, or to designing compact, well-founded metric subsets.

\smallskip
\textbf{Positioning of our work.}
The SUPERVISE framework intends to address all the mentioned issues. %
First, it provides an open source, efficient, robust and comprehensive scale-corrected comparison protocol,
allowing robust cross-metric visualizations (see Sec.~\ref{sec:scale_normalized_protocol}). %
Second, its unified design allows perfect reproducibility and fast benchmarking of new methods by
sharing results and label maps from state-of-the-art methods (see Sec.~\ref{sec:unified_pipeline}). %
Similarly to other computer vision applications, results of previous methods on standard datasets can be directly reported.
We propose in Sec. \ref{sec:benchmark_scope},
performance-visualization examples completed by a more detailed report on method performance in Appendix~\ref{sec:app_bsd_result}.
Third, it leverages these features to
provide a large-scale statistical analysis of metric and dataset redundancy (see Sec.~\ref{sec:statistics}).
This analysis
notably provides empirical support for using a compact, representative, and weakly redundant set of metrics for standardized superpixel evaluation.

\begin{figure}[t]
	\centering
	\includegraphics[width=0.975\textwidth]{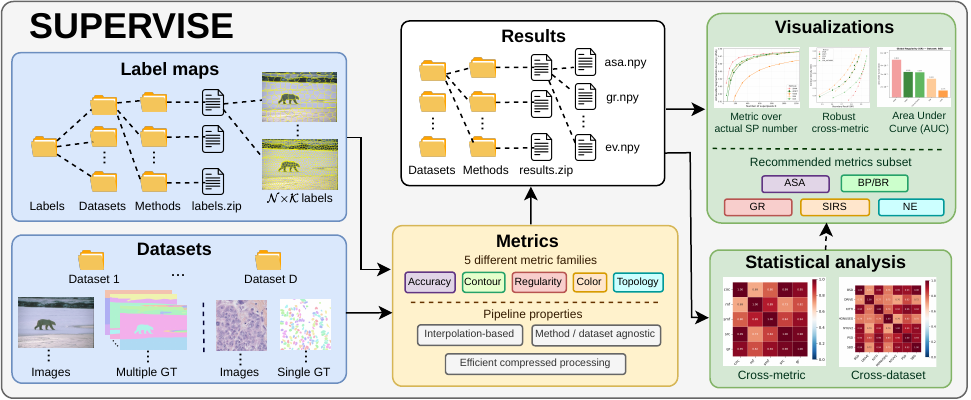}
	\caption{\textbf{Overview of the SUPERVISE benchmark}.
		The pipeline takes as input an image set, ground-truth annotations, and superpixel label maps produced by different methods at different scales.
		It computes a comprehensive range of metrics that
		feed robust visualizations over the
		actual number of generated superpixels and interpolated values over a specific range for cross-metric and AUC results.
		These stored and publicly shared results also feed a
		higher-level statistical analysis.
		By decoupling method execution from evaluation, SUPERVISE provides a unified, reproducible, dataset and method-agnostic framework for comparing superpixel methods.
	}
	\label{fig:schema_supervise}
\end{figure}

\section{SUPERVISE evaluation protocol}
\label{sec:benchmark}

This section presents the proposed SUPERVISE evaluation protocol, %
illustrated in Fig.~\ref{fig:schema_supervise}.
Our goal is to provide the first unified, high-level, fair, and entirely reproducible framework for comparing superpixel methods across datasets, metrics, and operating scales.

\subsection{Problem statement}
\label{subsec:problem_statement}

We model superpixel evaluation as a function defined over a multi-indexed space
$(d, m, i, k, c)$, where $d \in \mathcal{D}$ denotes a dataset, $m \in \mathcal{M}$ a method,
$i \in \mathcal{I}_d$ an image, $k \in \mathcal{K}_d$ the requested number of superpixels
(scale), and $c \in \mathcal{C}$ an evaluation metric.
Each evaluation produces a score:
\begin{equation}
	s(d,m,i,k,c) \in \mathbb{R}.
\end{equation}

A fundamental difficulty is that the requested scale $k$ does not reflect the actual generated scale
$\hat{k} \in \hat{\mathcal{K}}_{d,m,i}$, which depends on the dataset, method and image.
This prevents direct comparison of methods at fixed operating points.
We address this issue in Sec. \ref{sec:scale_normalized_protocol}.

Another challenge stems from the large number of existing  metrics ($\sim$20), which has led to inconsistent benchmarks.
As exhaustive reporting is impractical, this calls for a principled approach to identify a compact and representative subset.
SUPERVISE integrates a comprehensive set of metrics within a unified framework (Sec.~\ref{sec:unified_pipeline}) and provides a systematic analysis of their correlations to support such selection (Sec.~\ref{sec:statistics}).

\subsection{Scale-corrected report of performance}
\label{sec:scale_normalized_protocol}

\begin{figure*}[t!]
	\centering
	{\footnotesize
		\begin{tabular}{@{}c@{\hspace{2mm}}c@{\hspace{2mm}}c@{}}
			\includegraphics[width=0.31\textwidth]{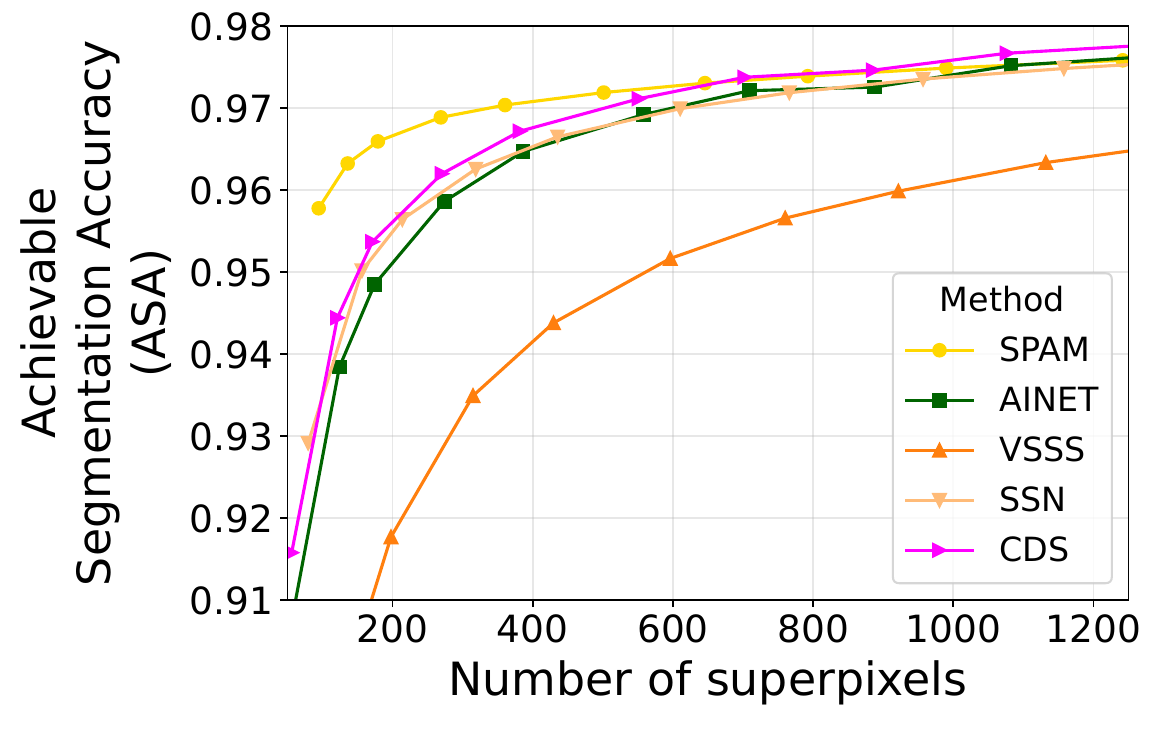}               &
			\includegraphics[width=0.31\textwidth]{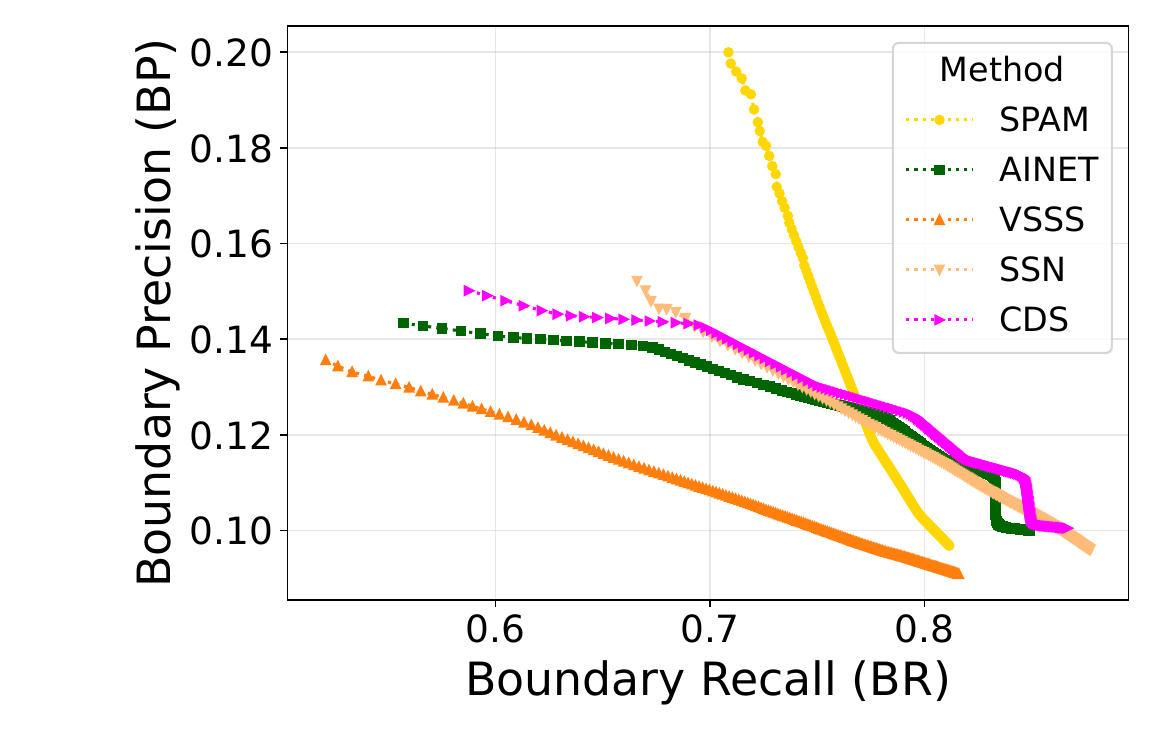} &
			\includegraphics[width=0.31\textwidth]{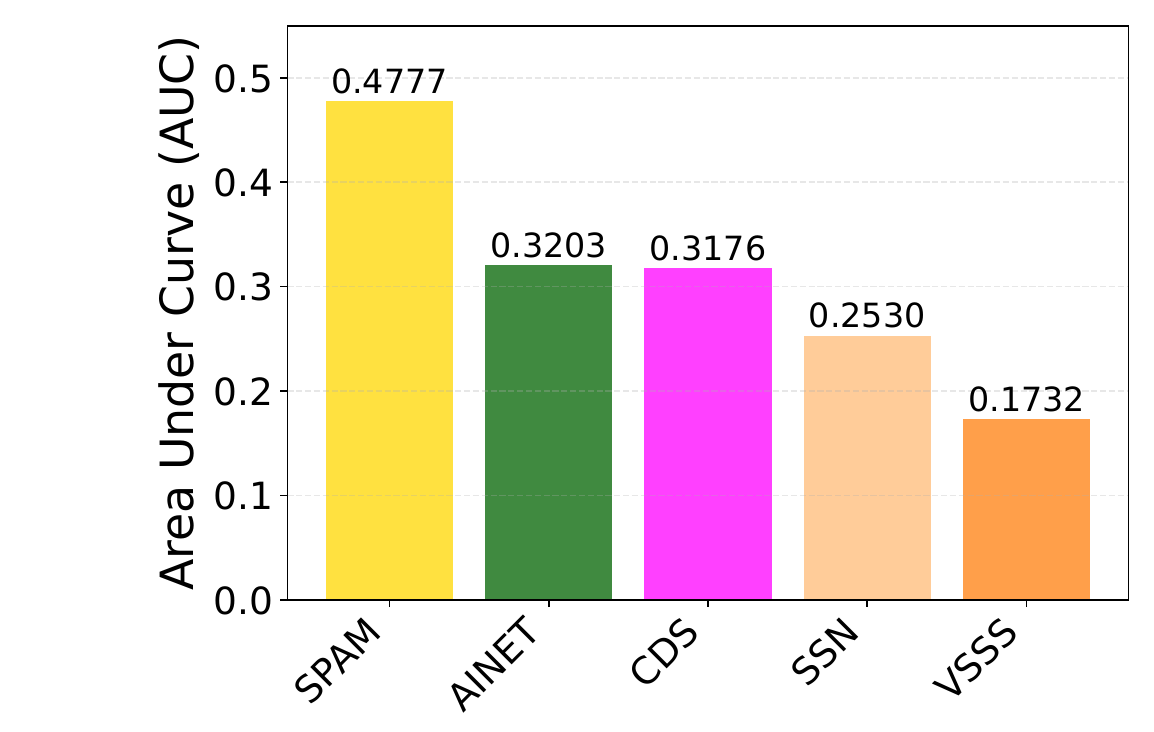}       \\
			(a) Standard display:                                                                         &
			(b) Interpolated cross-metric over range:                                                     &
			(c) Robust AUC over range:                                                                      \\[-0.5ex]
			ASA vs Actual SP number $\hat{k}$                                                             &
			BR vs BP for k in [100, 1000]                                                                   &
			GR for k in [100, 1000]
		\end{tabular}}
	\caption{\textbf{Example of performance visualizations.} Scale-normalized metric curves for a subset of evaluated methods on BSD. Each curve shows the interpolated metric value as a function of the actual number of generated superpixels, aggregated over the dataset. Panel (c) reports the normalized AUC used as the benchmark summary score.}
	\label{fig:visu_examples}
\end{figure*}

A central difficulty in superpixel benchmarking is that methods do not generally produce the requested number of superpixels, so comparing metric values at a fixed target scale introduces a systematic bias.
Fig.~\ref{fig:requested_vs_real} illustrates this issue with two methods producing a different number of superpixels for the same request.
When segmentation accuracy (ASA) is plotted against the requested number (biased),
Method 2 appears better,
simply because it operates at a finer effective scale.
Aligning results on the actual superpixel count (corrected) removes this artificial advantage and enables fair %
comparison.
This example directly motivates the SUPERVISE protocol and the robust visualizations
with such corrected display, robust interpolated cross-metrics and AUC score over a common range (see Fig.~\ref{fig:visu_examples}).

\textbf{Connected components.}
Some methods may produce non-connected regions, although connectivity is part of the superpixel definition, which can favorably bias evaluation.
Hence, following recent benchmarks \cite{stutz2018,barcelos2024}, we first treat each connected component as an individual superpixel.

\textbf{Scale correction.}
Initial scores $s(d,m,i,k,c)$ are defined at %
operating points $k \in \mathcal{K}_{d}$, i.e., the requested superpixel counts.
As in other benchmarks reviewed in Tab.~\ref{tab:benchmarks}, we first remap them onto the actual scale $\hat{k}\in \hat{\mathcal{K}}_{d,m,i}$, yielding $s(d,m,i,\hat{k},c)$ (Fig.~\ref{fig:requested_vs_real}), a correction adopted by only 40\% of the literature \cite{giraud2024}.
For visualization, scores are averaged over images at each scale $\hat{k}$ (see Fig.~\ref{fig:visu_examples}(a)).

\textbf{Interpolation and common scale range.}
Since actual scales $\hat{\mathcal{K}}_{d,m,i}$ differ across methods,
we approximate $s$ by a piecewise-linear interpolation $\tilde{s}$ of the discrete scores $s(d,m,i,\hat{k},c)$ over a common domain $\hat{k} \in [k_{\min}, k_{\max}]$.
This interpolation bridges method-specific discrete scales and enables consistent cross-method comparisons by sampling all methods over a shared scale range (see Fig.~\ref{fig:visu_examples}(b)). %
Beyond fair comparison, this formulation also facilitates the analysis of cross-metric trade-offs without bias due to operating-point mismatch.
The choice of the range $[k_{\min}, k_{\max}]$
naturally depends on the targeted application.
In Sec.~\ref{sec:statistics}, we consider a standard range corresponding roughly to superpixel sizes between $15{\times}15$ and $40{\times}40$ pixels that fall within the range of requested scales.
Moderate variations of this range have limited impact on the overall conclusions.

\textbf{Aggregation (AUC).}
To provide relevant average scores (Fig.~\ref{fig:visu_examples}(c)), we first average over images:
\begin{equation}
	\bar{s}(d,m,\hat{k},c) = \frac{1}{|\mathcal{I}_d|} \sum_{i \in \mathcal{I}_d} \tilde{s}(d,m,i,\hat{k},c),
\end{equation}
and then compute a scale-aggregated score for a given dataset $d$, method $m$ and metric $c$ as:
\begin{equation}
	\mathrm{AUC}(d,m,c) = \frac{1}{k_{\max}-k_{\min}} \int_{k_{\min}}^{k_{\max}} \bar{s}(d,m,k,c)\, dk.
\end{equation}

\definecolor{mred1}{RGB}{253,121,126}
\definecolor{mred2}{RGB}{245,27,24}
\definecolor{mblue1}{RGB}{112,168,253}
\definecolor{mblue2}{RGB}{2,103,209}
\definecolor{mlightblue}{RGB}{120,175,255}

\newcommand{\reddiamond}{%
	\raisebox{0.1ex}{\textcolor{mred2}{
			\begin{tikzpicture}
				\fill[mred2] (0,0.085) -- (0.085,0) -- (0,-0.085) -- (-0.085,0) -- cycle;
			\end{tikzpicture}
		}}%
}

\newcommand{\redcircle}{%
	\raisebox{0.1ex}{
		\begin{tikzpicture}
			\fill[mred1] (0,0) circle (0.075);
		\end{tikzpicture}
	}%
}

\newcommand{\bluesquare}{%
	\raisebox{0.1ex}{
		\begin{tikzpicture}
			\fill[mblue2] (0,0) rectangle (0.15,0.15);
		\end{tikzpicture}
	}%
}

\newcommand{\lightbluetriangle}{%
	\raisebox{0.1ex}{\textcolor{mblue1}{
			\begin{tikzpicture}
				\fill[mblue1] (0,0) -- (0.17,0) -- (0.085,0.15) -- cycle;
			\end{tikzpicture}
		}}%
}

\definecolor{mred}{RGB}{220,40,40}
\definecolor{mblue}{RGB}{40,90,220}
\begin{figure*}[t!]
	\centering
	{\footnotesize
		\begin{tabular}{@{\hspace{0mm}}c@{\hspace{2mm}}c@{\hspace{5mm}}c@{\hspace{2mm}}c@{\hspace{2mm}}}
			Method 1: $\text{ASA}=0.970$                                                                        & Method 2: $\text{ASA}=0.971$ \\[0.2ex]
			\includegraphics[width=0.31\textwidth]{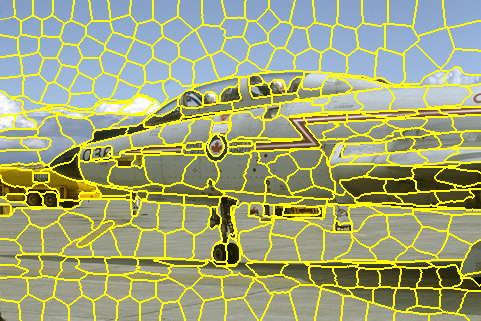} &
			\includegraphics[width=0.31\textwidth]{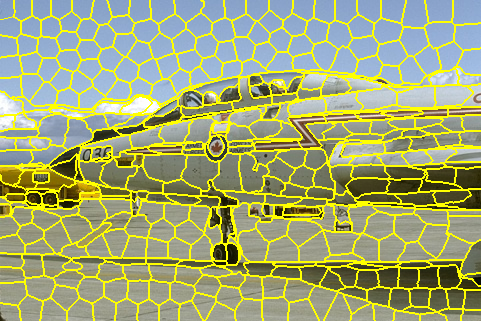}   &
			\includegraphics[width=0.31\textwidth]{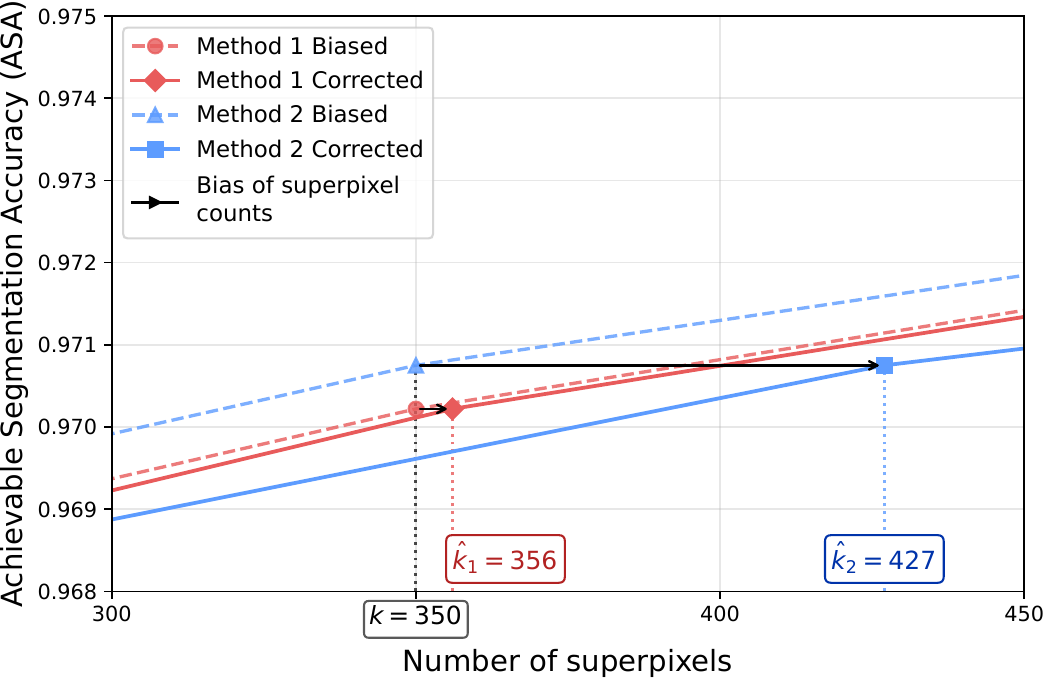}                                    \\
			\redcircle{}{\color{mred1}$k=350$} / {\color{mred2}${\hat{k}_{1}}=356$ \reddiamond{}} %
			                                                                                                    &
			\lightbluetriangle{}{\color{mblue1}$k=350$} / ${\color{mblue2}{\hat{k}_{2}}=427}$ \bluesquare{}
			                                                                                                    &
			Biased vs Corrected performance shift
		\end{tabular}}
	\caption{\textbf{Scale bias induced by reported superpixel number.}
		Two decompositions generated with the same requested number of superpixels
		($k=350$) can produce different counts ($\hat{k}_{1}=356$ and
		$\hat{k}{_2}=427$).
		Since many metrics improve at finer scales,
		aligning results on the actual superpixel count (corrected) removes a potential artificial advantage and enables fair scale-corrected comparison.}
	\label{fig:requested_vs_real}
\end{figure*}

\subsection{Unified evaluation framework}
\label{sec:unified_pipeline}

Fig.~\ref{fig:schema_supervise} provides an overview of the proposed framework.
SUPERVISE is designed as a modular evaluation pipeline that separates method execution from benchmark computation.
Each method is represented only through its output superpixel label maps, independently of the implementation language or training/testing procedure, ensuring identical evaluation conditions.
For each image, a method provides superpixel decompositions at several scales.
The benchmark then computes a comprehensive set of complementary metrics %
among the five superpixel properties.

\smallskip
\textbf{Large number of metrics and methods.}
Superpixel evaluation has accumulated a large and heterogeneous set of metrics over the years, often leading to inconsistent practices.
SUPERVISE consolidates most existing metrics and provides efficient,
high-level Python implementations, making them easy to use and interpret.
The framework is designed to be easily extensible, allowing new metrics to be added with minimal effort.
It also includes tools for statistical analysis and correlation studies, supporting principled comparison and metric selection.
We also include a broad set of methods with
open-source implementations, notably those compiled by \cite{barcelos2024}, using their default parameters, and aim to continuously extend the benchmark by evaluating new  methods.

\smallskip
\textbf{Built-in reproducibility.}
SUPERVISE relies on a lightweight representation based on shared precomputed label maps, avoiding the need to re-run methods for evaluation.
All metrics are computed from these label maps, %
ensuring a fully reproducible and implementation-independent pipeline.
By decoupling method execution from evaluation, this design removes variability due to software environments, hardware settings, or unavailable implementations, while enabling efficient and parallel processing.
This design choice also makes the benchmark naturally extensible and easy to use: new methods and datasets can be integrated by simply providing label and groundtruth maps in the standardized format. Importantly, adding a new method only extends the benchmark: previously computed metric values and scores remain unchanged, since evaluation is performed independently from method execution.
Finally, in contrast to existing benchmarks (see Tab.~\ref{tab:benchmarks}), the Python implementation relies exclusively on standard libraries, ensuring broad cross-platform compatibility.

\smallskip
\textbf{Label map generation.
 All benchmark label maps are generated using the public implementations released by the method authors, with their default parameters and, for learning-based approaches, the released pretrained weights. No additional training or parameter tuning is performed. Each method is evaluated at ten requested scales. We provide references to the exact implementations and pretrained models used to generate the released maps.}

\smallskip
\textbf{Efficient compressed processing.}
Superpixel label maps are stored in a compressed format (\emph{e.g.}, 4\,GB raw results reduced to 40\,MB for 10 scale decompositions on the 200 images of the BSD dataset), making them easy to share and distribute.
They are accessed directly without full decompression, enabling efficient I/O and memory usage.
Our pipeline is fully parallelized, allowing fast large-scale processing: \emph{e.g.}, evaluating a method on 200 images, with 4--8 groundtruth annotations per image, 10 scales, and 20 metrics, takes approximately 25 minutes on a standard 14-core laptop.
Such efficiency and ease of use are intended to facilitate rapid development and evaluation of superpixel methods.

\smallskip
\textbf{Limitations.}
While \SUPERVISE{} aims to provide a comprehensive and reproducible benchmark, it relies on precomputed label maps, which requires trust in the correctness of the provided decompositions. In addition, as with any large-scale benchmark, its relevance depends on continuous updates, including the integration of new methods, datasets, and evaluation protocols.
Furthermore, this work focuses on intrinsic evaluation, and the relationship between these metrics and downstream task performance remains to be systematically investigated. Despite these limitations, \SUPERVISE{} offers, to the best of our knowledge, the simplest, most comprehensive, and most reproducible benchmark to date for addressing the inherent complexity of superpixel evaluation at scale.

\subsection{Benchmark of superpixel methods}
\label{sec:benchmark_scope}

To demonstrate the capabilities of SUPERVISE, we conduct a large-scale benchmark covering 31 superpixel methods, 20 evaluation metrics, and 7 datasets (listed in Appendix~\ref{sec:appendix_dataset}).
This extensive setup highlights the robustness, flexibility, scalability, and reproducibility of the proposed framework.
Still, we emphasize that the primary goal of this study is to illustrate how a unified and statistically grounded evaluation protocol can reveal meaningful trends, inconsistencies, and redundancies across metrics and datasets, rather than to establish a definitive performance ranking.

For each superpixel method, we collect decompositions at ten requested operating levels, covering both coarse and fine partitions. For scale-aggregated evaluation, the common interpolation range is selected independently for each dataset so that it remains within the evaluated scales and approximately corresponds to superpixel sizes between \(15\times15\) and \(40\times40\) pixels. We use \([100,1000]\) for BSD, NYUv2 and SBD, \([150,1250]\) for KITTI, DRIVE and PSD, and \([150,1500]\) for MONUSEG. Within a dataset, the same interval is used for every method and metric and is never selected according to the resulting ranking. These bounds are user-defined and can be adapted to any task-specific scale regime. Their influence on method rankings is analyzed in Appendix \ref{sec:ranking_stability}. All results reported in the benchmark follow the protocol defined in Secs. \ref{sec:scale_normalized_protocol} and \ref{sec:unified_pipeline}.

Tab.~\ref{tab:auc-category-ranks-recommended} provides a compact overview of the benchmark results. Instead of reporting every individual metric, it ranks methods on the recommended compact subset from Sec.~\ref{sec:statistics}.
In Appendix~\ref{sec:app_bsd_result}, we report the comprehensive evaluation and
ranking on the standard BSD dataset.
For the other datasets, we report a compact evaluation on this subset, derived from prior literature and further supported by our correlation analysis.
Overall, our results are consistent with recent benchmarks \cite{barcelos2024,wu2026survey}: deep learning-based methods---usually trained on the BSD dataset---largely dominate in terms of accuracy and contour adherence, with lower performance in color homogeneity and regularity.
The proposed cross-metric visualizations, such as ASA vs GR, provide a robust way to capture these trade-offs.
In particular, recent foundation-based approaches such as \cite{walther2025spam} demonstrate an improved balance between
these competing objectives.

\section{Statistical analysis of metrics and datasets}
\label{sec:statistics}

Due to the ill-posed and multi-objective nature of superpixel segmentation,
many evaluation metrics have been proposed, often capturing overlapping behaviors,
as highlighted in~\cite{stutz2018,giraud2024,wu2026survey}.
As a result, reporting numerous metrics may give the appearance of comprehensive evaluation while effectively measuring a smaller set of latent criteria.

We leverage the standardized, scale-aligned outputs of SUPERVISE to analyze in depth the correlations between metrics and in Appendix~\ref{sec:cross_dataset_compact_panel}, between datasets.
All analyses are based on the scale-normalized interpolation protocol introduced in Sec.~\ref{sec:scale_normalized_protocol}.
Each observation corresponds to an interpolated value of the metric on the common range of actual superpixel counts.

\begin{table*}[t]
\centering\tiny
\setlength{\tabcolsep}{1.6pt}
\renewcommand{\arraystretch}{0.9}
\caption{\textbf{Method rankings by AUC for the recommended metric subset across datasets}. The five column groups correspond to ASA, \brbp{}, SIRS, GR, and NE, respectively. Ranks are reported for NYUv2 (N), BSD (B), SBD (S), KITTI (K), DRIVE (D), MONUSEG (M), and PSD (P). Rank 1 is best.}
\label{tab:auc-category-ranks-recommended}
\resizebox{\textwidth}{!}{%
\begin{tabular}{l>{\columncolor{aucaccuracy}}r>{\columncolor{aucaccuracy}}r>{\columncolor{aucaccuracy}}r>{\columncolor{aucaccuracy}}r>{\columncolor{aucaccuracy}}r>{\columncolor{aucaccuracy}}r>{\columncolor{aucaccuracy}}r>{\columncolor{auccontour}}r>{\columncolor{auccontour}}r>{\columncolor{auccontour}}r>{\columncolor{auccontour}}r>{\columncolor{auccontour}}r>{\columncolor{auccontour}}r>{\columncolor{auccontour}}r>{\columncolor{auccolor}}r>{\columncolor{auccolor}}r>{\columncolor{auccolor}}r>{\columncolor{auccolor}}r>{\columncolor{auccolor}}r>{\columncolor{auccolor}}r>{\columncolor{auccolor}}r>{\columncolor{aucregularity}}r>{\columncolor{aucregularity}}r>{\columncolor{aucregularity}}r>{\columncolor{aucregularity}}r>{\columncolor{aucregularity}}r>{\columncolor{aucregularity}}r>{\columncolor{aucregularity}}r>{\columncolor{auctopology}}r>{\columncolor{auctopology}}r>{\columncolor{auctopology}}r>{\columncolor{auctopology}}r>{\columncolor{auctopology}}r>{\columncolor{auctopology}}r>{\columncolor{auctopology}}r}
\toprule
 & \multicolumn{7}{c}{\textcolor{metricaccuracy}{Accuracy}} & \multicolumn{7}{c}{\textcolor{metriccontour}{Contour}} & \multicolumn{7}{c}{\textcolor{metriccolor}{Color}} & \multicolumn{7}{c}{\textcolor{metricregularity}{Regularity}} & \multicolumn{7}{c}{\textcolor{metrictopology}{Topology}} \\
\cmidrule(lr){2-8} \cmidrule(lr){9-15} \cmidrule(lr){16-22} \cmidrule(lr){23-29} \cmidrule(lr){30-36}
Method & N & B & S & K & D & M & P & N & B & S & K & D & M & P & N & B & S & K & D & M & P & N & B & S & K & D & M & P & N & B & S & K & D & M & P \\
\midrule
AINET \cite{wang2021ainet} & 9 & 6 & 6 & 6 & 20 & 24 & 15 & 6 & 3 & 4 & 10 & 22 & 26 & 25 & 28 & 28 & 27 & 23 & 12 & 23 & 28 & 13 & 10 & 12 & 12 & 15 & 4 & 11 & 27 & 27 & 27 & 27 & 27 & 29 & 28 \\
BASS \cite{uziel2019bass} & 22 & 25 & 22 & 25 & 24 & 22 & 28 & 4 & 7 & 2 & 1 & 21 & 5 & 1 & 2 & 4 & 4 & 1 & 8 & 1 & 3 & 18 & 19 & 19 & 20 & 12 & 31 & 19 & 14 & 17 & 16 & 18 & 10 & 25 & 23 \\
BINET \cite{zhao2024rethinking} & 3 & 8 & 4 & 10 & 12 & 29 & 23 & 2 & 11 & 7 & 13 & 12 & 29 & 28 & 24 & 29 & 25 & 25 & 1 & 29 & 29 & 17 & 13 & 16 & 14 & 23 & 6 & 17 & 29 & 28 & 29 & 28 & 29 & 28 & 27 \\
CDS \cite{xu2024cds} & 4 & 3 & 2 & 3 & 4 & 27 & 7 & 3 & 2 & 5 & 4 & 7 & 28 & 23 & 25 & 25 & 23 & 21 & 18 & 26 & 24 & 15 & 11 & 13 & 11 & 19 & 5 & 13 & 30 & 30 & 28 & 30 & 28 & 31 & 29 \\
CRTREES \cite{yan2022hierarchical} & 24 & 27 & 19 & 23 & 17 & 23 & 11 & 13 & 23 & 18 & 20 & 10 & 16 & 9 & 27 & 24 & 17 & 26 & 30 & 27 & 23 & 25 & 24 & 25 & 25 & 26 & 17 & 25 & 1 & 1 & 1 & 1 & 1 & 1 & 1 \\
DISF \cite{belem2020superpixel} & 11 & 9 & 9 & 8 & 1 & 4 & 3 & 9 & 10 & 8 & 8 & 1 & 4 & 3 & 1 & 1 & 1 & 3 & 31 & 4 & 1 & 29 & 29 & 29 & 28 & 30 & 27 & 29 & 8 & 8 & 7 & 9 & 8 & 10 & 9 \\
ERS \cite{liu2011} & 8 & 13 & 12 & 9 & 7 & 21 & 2 & 16 & 21 & 20 & 3 & 8 & 9 & 7 & 10 & 16 & 16 & 16 & 2 & 20 & 17 & 22 & 21 & 23 & 21 & 24 & 10 & 23 & 7 & 7 & 8 & 7 & 9 & 8 & 8 \\
ETPS \cite{yao2015} & 18 & 15 & 16 & 21 & 29 & 19 & 26 & 17 & 12 & 11 & 11 & 27 & 14 & 5 & 3 & 2 & 2 & 2 & 14 & 3 & 2 & 6 & 12 & 15 & 9 & 5 & 15 & 7 & 21 & 23 & 22 & 22 & 20 & 23 & 21 \\
FUZZYSLIC \cite{wu2020fuzzy} & 17 & 18 & 17 & 18 & 23 & 14 & 25 & 27 & 26 & 27 & 27 & 26 & 20 & 26 & 18 & 17 & 19 & 18 & 9 & 18 & 20 & 3 & 4 & 3 & 4 & 2 & 13 & 4 & 25 & 25 & 25 & 24 & 21 & 24 & 25 \\
GMMSP \cite{Ban18} & 10 & 11 & 11 & 13 & 16 & 9 & 10 & 19 & 15 & 19 & 9 & 18 & 13 & 6 & 12 & 8 & 7 & 17 & 23 & 15 & 12 & 14 & 17 & 10 & 18 & 13 & 23 & 15 & 23 & 22 & 13 & 20 & 18 & 15 & 19 \\
GRID & 30 & 31 & 31 & 30 & 30 & 31 & 31 & 31 & 31 & 31 & 31 & 31 & 31 & 31 & 31 & 31 & 31 & 31 & 26 & 31 & 31 & 1 & 1 & 1 & 1 & 1 & 1 & 1 & 12 & 13 & 21 & 12 & 12 & 22 & 14 \\
HHTS \cite{chang2024hierarchical} & 31 & 28 & 30 & 31 & 27 & 15 & 30 & 24 & 25 & 22 & 5 & 11 & 19 & 10 & 21 & 6 & 14 & 12 & 4 & 2 & 16 & 31 & 31 & 30 & 31 & 31 & 29 & 31 & 1 & 1 & 1 & 1 & 1 & 1 & 1 \\
LNSNET \cite{zhu2021learning} & 29 & 26 & 26 & 29 & 19 & 20 & 17 & 29 & 27 & 24 & 29 & 19 & 24 & 27 & 30 & 23 & 22 & 30 & 28 & 28 & 25 & 24 & 22 & 21 & 26 & 14 & 19 & 22 & 13 & 14 & 15 & 16 & 24 & 19 & 16 \\
LSC \cite{chen2017} & 13 & 14 & 14 & 16 & 13 & 7 & 12 & 14 & 20 & 14 & 24 & 16 & 11 & 8 & 4 & 11 & 12 & 15 & 22 & 5 & 9 & 11 & 14 & 9 & 17 & 10 & 21 & 12 & 20 & 18 & 19 & 26 & 19 & 16 & 22 \\
NNSC \cite{giraud2019texture} & 15 & 19 & 24 & 17 & 26 & 10 & 19 & 25 & 28 & 29 & 28 & 29 & 8 & 24 & 14 & 18 & 26 & 14 & 16 & 11 & 14 & 8 & 6 & 8 & 6 & 9 & 12 & 6 & 31 & 31 & 31 & 31 & 31 & 26 & 31 \\
RISF \cite{galvao2020image} & 20 & 21 & 21 & 19 & 8 & 6 & 13 & 18 & 17 & 21 & 19 & 5 & 7 & 12 & 8 & 5 & 6 & 7 & 20 & 10 & 6 & 23 & 23 & 22 & 24 & 21 & 26 & 21 & 1 & 1 & 1 & 1 & 1 & 1 & 1 \\
RSS \cite{chai2020rooted} & 25 & 23 & 27 & 22 & 6 & 13 & 24 & 12 & 14 & 15 & 16 & 3 & 18 & 16 & 7 & 7 & 8 & 6 & 17 & 13 & 7 & 27 & 27 & 26 & 27 & 25 & 28 & 27 & 15 & 12 & 12 & 13 & 17 & 13 & 13 \\
SAMO \cite{giraud2024} & 2 & 2 & 8 & 5 & 25 & 2 & 8 & 7 & 6 & 16 & 14 & 25 & 2 & 22 & 20 & 26 & 28 & 28 & 27 & 21 & 22 & 9 & 7 & 7 & 7 & 8 & 14 & 9 & 10 & 10 & 10 & 10 & 13 & 12 & 11 \\
SCALP \cite{giraud2018_scalp} & 12 & 12 & 13 & 15 & 18 & 16 & 14 & 23 & 18 & 26 & 26 & 24 & 15 & 20 & 9 & 10 & 13 & 10 & 25 & 9 & 10 & 4 & 5 & 5 & 5 & 3 & 9 & 5 & 19 & 16 & 17 & 19 & 23 & 21 & 20 \\
SEAL \cite{tu2018learning} & 5 & 4 & 3 & 1 & 31 & 28 & 1 & 20 & 13 & 17 & 7 & 20 & 23 & 18 & 22 & 22 & 21 & 20 & 3 & 24 & 18 & 26 & 26 & 27 & 23 & 28 & 11 & 26 & 9 & 9 & 9 & 8 & 11 & 9 & 10 \\
SEEDS \cite{vandenbergh2012} & 23 & 22 & 29 & 14 & 15 & 17 & 20 & 28 & 29 & 28 & 12 & 15 & 10 & 13 & 19 & 21 & 29 & 4 & 5 & 6 & 5 & 21 & 20 & 20 & 19 & 18 & 20 & 20 & 22 & 20 & 18 & 21 & 7 & 6 & 7 \\
SFCN \cite{yang2020} & 6 & 7 & 5 & 7 & 14 & 25 & 16 & 8 & 5 & 10 & 18 & 17 & 27 & 29 & 26 & 27 & 24 & 22 & 7 & 22 & 27 & 12 & 9 & 11 & 10 & 17 & 3 & 10 & 28 & 29 & 30 & 29 & 30 & 30 & 30 \\
SH \cite{wei2018sh} & 14 & 10 & 10 & 11 & 3 & 8 & 4 & 15 & 19 & 13 & 15 & 6 & 6 & 11 & 15 & 9 & 5 & 13 & 24 & 12 & 8 & 28 & 28 & 28 & 30 & 29 & 25 & 28 & 1 & 1 & 5 & 1 & 1 & 1 & 1 \\
SICLE \cite{belem2022fast} & 26 & 17 & 15 & 12 & 2 & 5 & 9 & 10 & 8 & 3 & 17 & 2 & 3 & 2 & 6 & 3 & 3 & 5 & 13 & 8 & 4 & 30 & 30 & 31 & 29 & 27 & 30 & 30 & 6 & 6 & 6 & 6 & 6 & 7 & 6 \\
SIT-HSS \cite{xie2025hierarchical} & 21 & 16 & 20 & 26 & 5 & 18 & 22 & 26 & 9 & 1 & 21 & 4 & 21 & 21 & 23 & 13 & 9 & 27 & 15 & 25 & 26 & 16 & 16 & 18 & 15 & 22 & 8 & 16 & 1 & 1 & 1 & 5 & 1 & 1 & 1 \\
SLIC \cite{achanta2012} & 16 & 20 & 18 & 24 & 22 & 11 & 21 & 22 & 22 & 25 & 25 & 23 & 17 & 19 & 17 & 20 & 18 & 24 & 29 & 17 & 21 & 7 & 8 & 6 & 8 & 6 & 16 & 8 & 24 & 24 & 20 & 25 & 22 & 20 & 24 \\
SNIC \cite{achanta2017superpixels} & 19 & 24 & 23 & 20 & 11 & 12 & 18 & 21 & 24 & 23 & 23 & 13 & 12 & 14 & 11 & 12 & 10 & 11 & 6 & 14 & 11 & 10 & 15 & 14 & 13 & 11 & 18 & 14 & 16 & 15 & 14 & 14 & 16 & 14 & 15 \\
SPAM \cite{walther2025spam} & 1 & 1 & 1 & 2 & 28 & 1 & 5 & 1 & 1 & 12 & 6 & 30 & 1 & 17 & 16 & 19 & 20 & 19 & 10 & 7 & 19 & 5 & 3 & 4 & 3 & 7 & 7 & 3 & 11 & 11 & 11 & 11 & 15 & 11 & 12 \\
SSN \cite{jampani2018} & 7 & 5 & 7 & 4 & 9 & 3 & 6 & 5 & 4 & 9 & 2 & 14 & 22 & 15 & 13 & 15 & 15 & 9 & 11 & 19 & 15 & 19 & 18 & 17 & 16 & 20 & 22 & 18 & 18 & 21 & 24 & 17 & 26 & 17 & 17 \\
TP \cite{levinshtein2009turbopixels} & 28 & 30 & 28 & 28 & 21 & 30 & 29 & 30 & 30 & 30 & 30 & 28 & 30 & 30 & 29 & 30 & 30 & 29 & 21 & 30 & 30 & 2 & 2 & 2 & 2 & 4 & 2 & 2 & 26 & 26 & 26 & 23 & 25 & 27 & 26 \\
VSSS \cite{zhou2023vine} & 27 & 29 & 25 & 27 & 10 & 26 & 27 & 11 & 16 & 6 & 22 & 9 & 25 & 4 & 5 & 14 & 11 & 8 & 19 & 16 & 13 & 20 & 25 & 24 & 22 & 16 & 24 & 24 & 17 & 19 & 23 & 15 & 14 & 18 & 18 \\
\bottomrule
\end{tabular}%
}
\end{table*}

\begin{figure*}[t!]
	\centering
	\definecolor{metricaccuracy}{RGB}{105,56,135}
	\definecolor{metriccontour}{RGB}{64,130,205}
	\definecolor{metricregularity}{RGB}{210,70,70}
	\definecolor{metriccolor}{RGB}{220,145,0}
	\definecolor{metrictopology}{RGB}{0,155,145}
	\setlength{\tabcolsep}{2pt}
	\renewcommand{\arraystretch}{0.8}

	{\scriptsize
		\begin{tabular}{c@{\hspace{0mm}}c@{\hspace{0mm}}c@{\hspace{0mm}}c@{\hspace{2mm}}c@{\hspace{0mm}}}
			\includegraphics[width=0.19\linewidth]{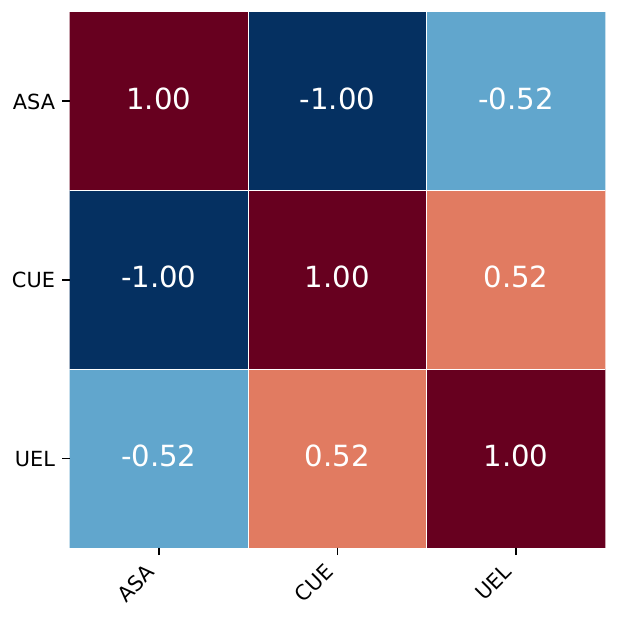}   &
			\includegraphics[width=0.19\linewidth]{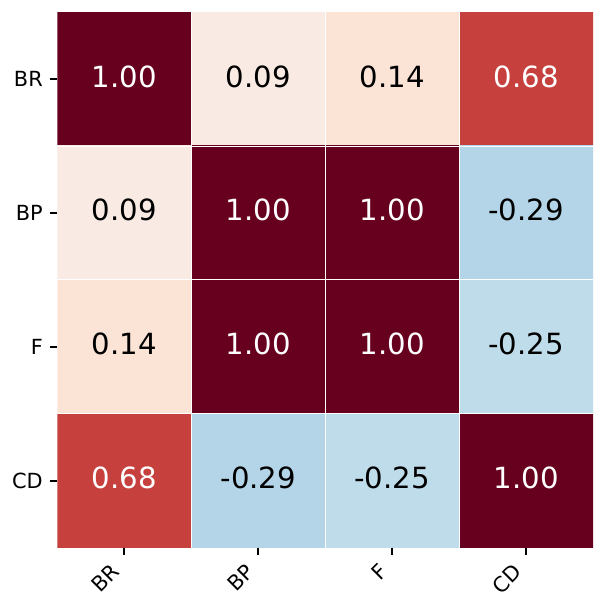}    &
			\includegraphics[width=0.19\linewidth]{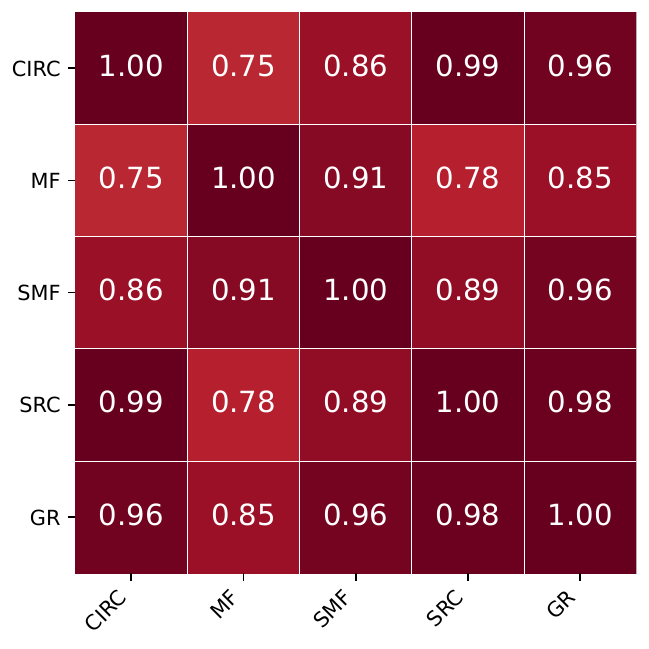} &
			\includegraphics[width=0.19\linewidth]{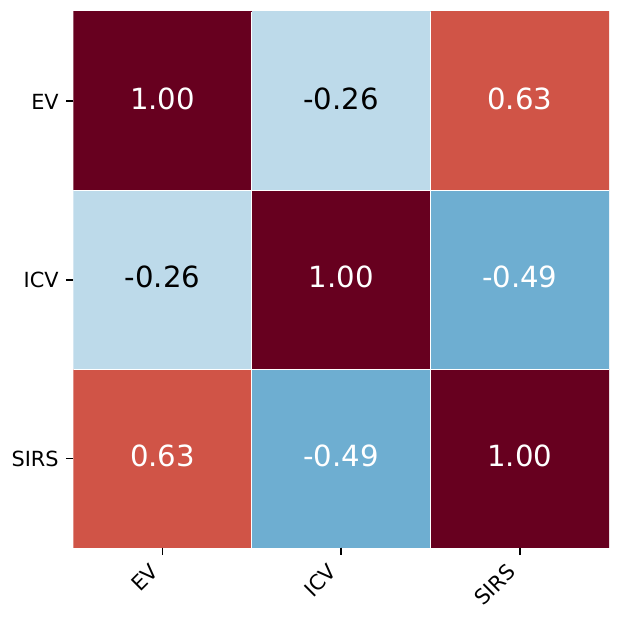}      &
			\includegraphics[width=0.19\linewidth]{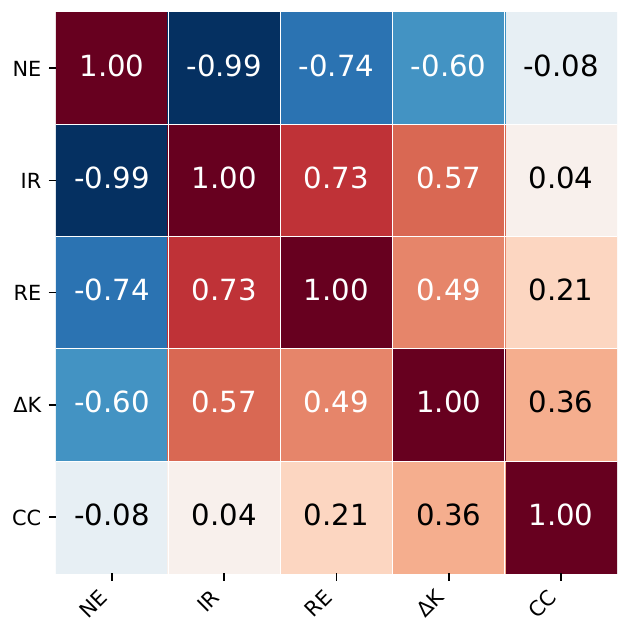}   \\
			\scriptsize \textcolor{metricaccuracy}{Accuracy}                    &
			\scriptsize \textcolor{metriccontour}{Contour}                      &
			\scriptsize \textcolor{metricregularity}{Regularity}                &
			\scriptsize \textcolor{metriccolor}{Color}                          &
			\scriptsize \textcolor{metrictopology}{Topology}                    \\
		\end{tabular}
	}

	\caption{\textbf{Metric statistical correlation analysis.}
		Experiments are performed on the BSD dataset. Spearman correlations by metric family.
	}
	\label{fig:correlation_families_bsd}
\end{figure*}

\begin{figure*}[t!]
	\centering
	{\scriptsize
		\begin{tabular}{c@{\hspace{2mm}}c@{\hspace{2mm}}c@{\hspace{0mm}}}
			\includegraphics[width=0.3000\textwidth]{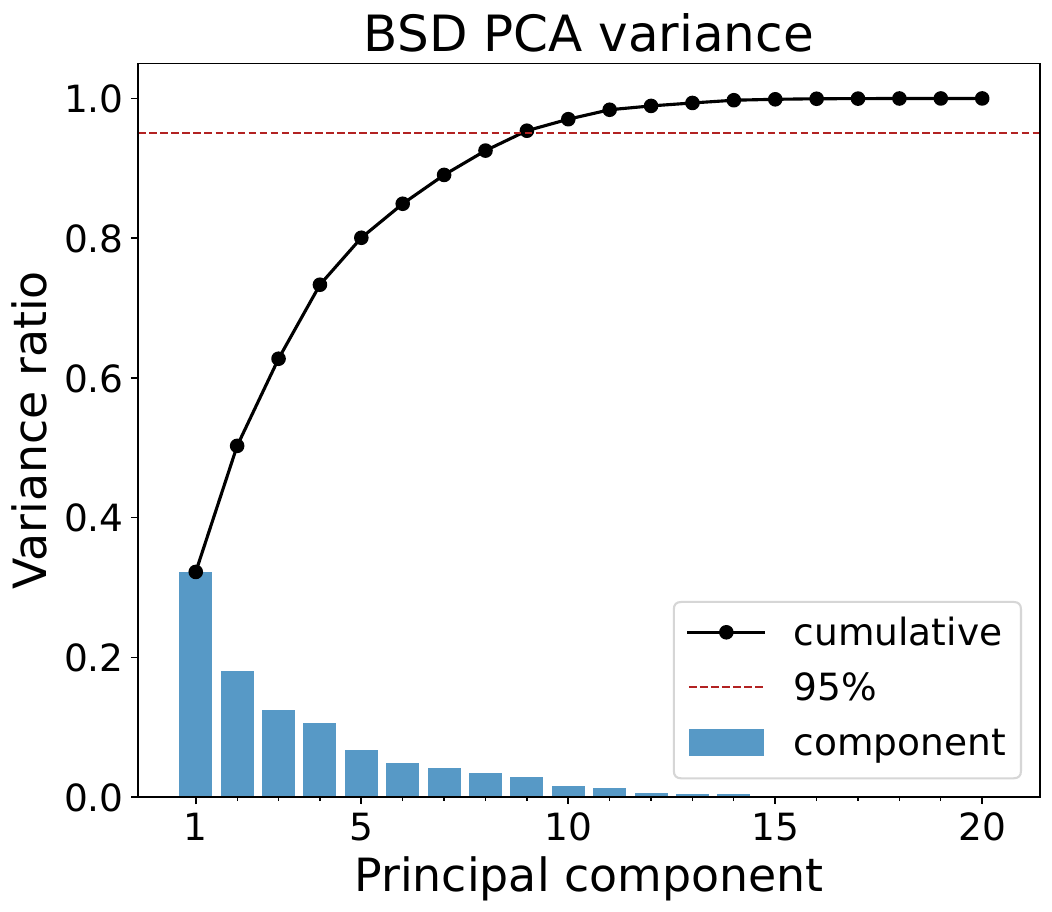} &
			\includegraphics[width=0.3735\textwidth]{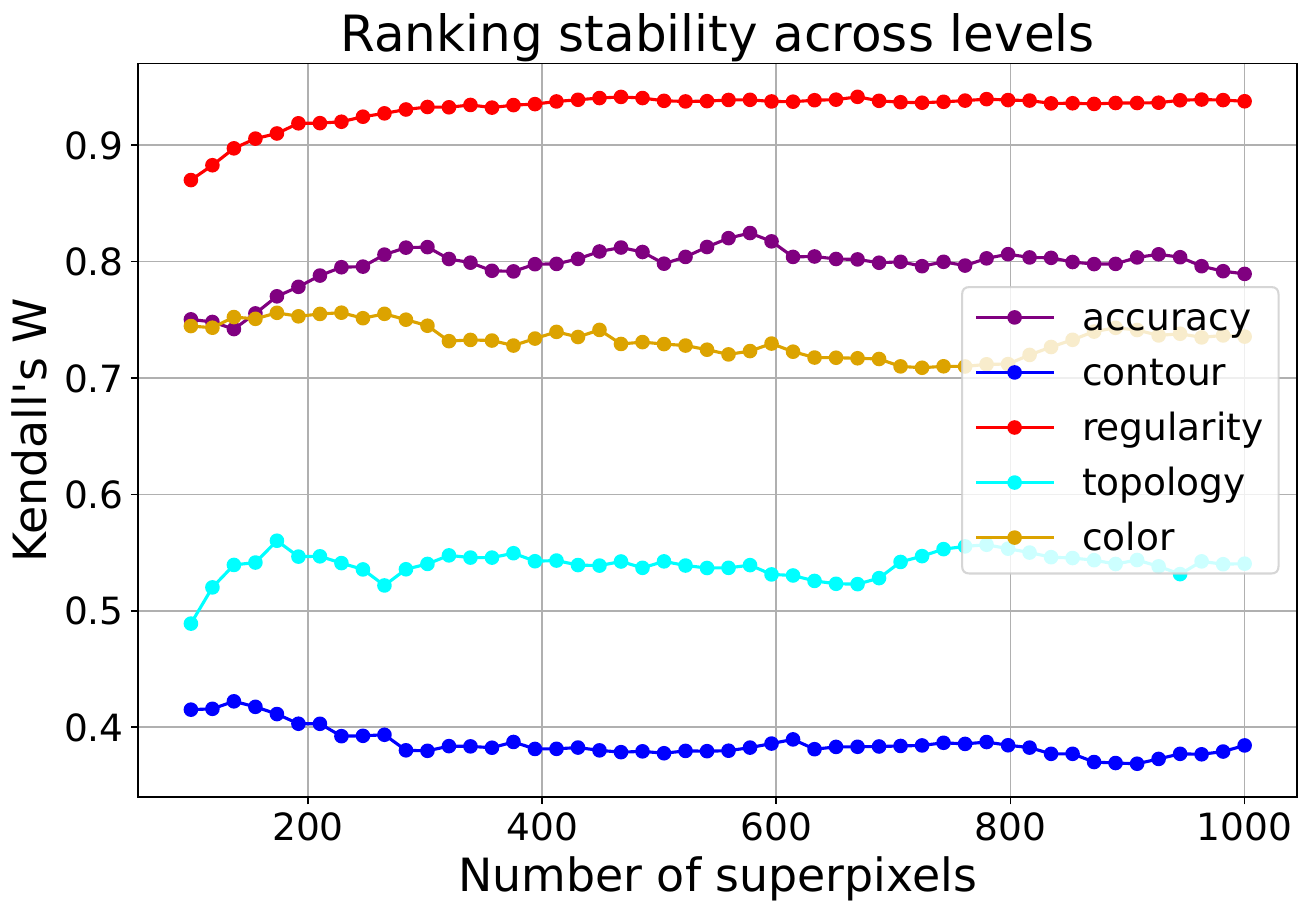} &
			\includegraphics[width=0.2723\textwidth]{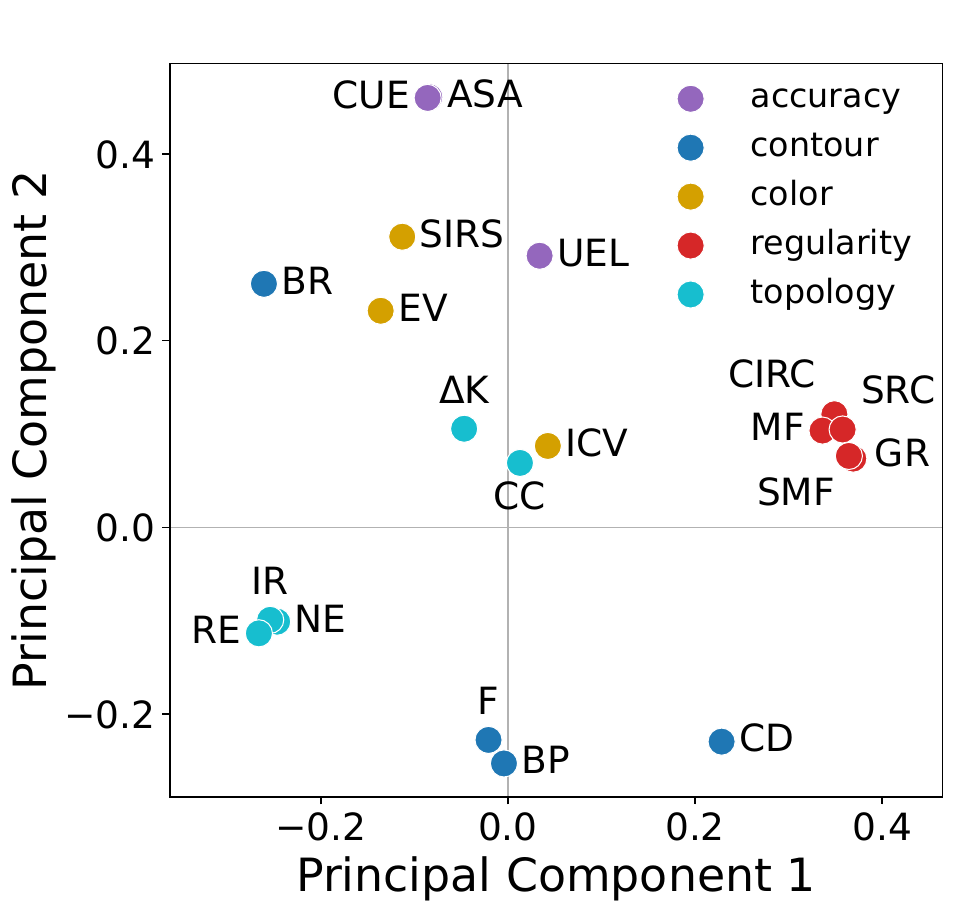} \\
		\end{tabular}
	}
	\caption{\textbf{Metric statistical correlation analysis.}
		Experiments are performed on the BSD dataset.
		(Left) Principal Component Analysis (PCA) explained variance on BSD. (Middle) Kendall's concordance W by categories. (Right) 2D PCA projection.
	}
	\label{fig:correlation_cat_bsd}
\end{figure*}

\subsection{Cross-metric correlation}

In this section, we present a detailed per-family analysis of the correlation and dependencies between
the large set of considered metrics in Fig.~\ref{fig:correlation_families_bsd}, following the protocol of \cite{TA2015}.
All experiments are reported on the standard BSD dataset.
In Appendix \ref{sec:appendix_metrics}, we report a full description of the considered metrics, with recommendations of use.
This study aims to identify structural redundancies within the current evaluation landscape, while providing large-scale empirical support for metric interpretations and recommendations that have so far remained mostly qualitative in the literature.

As a result, as illustrated in Fig.~\ref{fig:schema_supervise},
these correlations empirically confirm some prior recommendations from \cite{stutz2018,barcelos2024,giraud2024} and support the following  practical compact subset of
one metric/score per family, \emph{i.e.},
segmentation accuracy: ASA;
contour adherence: \brbp{};
regularity: GR;
color homogeneity: SIRS; %
topology: NE.
All these metrics are bounded in [0,1] and higher values indicate better performance.
Yet, %
the choice of metrics may depend on the targeted evaluation.

\textbf{Pairwise and multivariate dependencies.}
We first use pairwise Spearman correlations between metrics,
as it captures monotonic dependencies independently of metric scale.
The accuracy metrics appear to be strongly correlated,
with ASA and CUE showing an almost identical behavior,
as suggested by \cite{stutz2018}, while UEL remains less coupled.
This demonstrates the redundancy of reporting both ASA and CUE,
as several recent papers do \cite{giraud2024}.
Overall, most works report ASA, which---similarly to other metrics---is defined in [0,1], where higher values correspond to better performance.

The contour family also contains clear dependencies related to standard Boundary Precision (BP)-Recall (BR) metrics.
Firstly, BP and F (F-measure), computed as $\text{F}=2\frac{\text{BR}\times\text{BP}}{\text{BR}+\text{BP}}$, appear almost perfectly correlated.
This surprising observation can be explained by the oversegmentation nature of the superpixel segmentation task: boundary precision is generally low, due to the large number of superpixel boundaries that do not correspond to groundtruth contours,
whereas boundary recall is usually high, leading to $\text{F}\approx 2\text{BP}$.
This observation suggests discarding the \text{F}-measure and favoring robust \text{BR} over \text{BP} cross-metric visualization from common interpolation ranges using our framework.

Regularity metrics evaluate shape regularity and homogeneity in different ways.
This family exhibits the strongest internal correlation,
seemingly indicating that superpixel methods are not prone to produce superpixels of very specific shapes that would lead to inconsistent measures.
In the absence of clear discrepancies, metric definitions and potential robustness prevail and
 GR is advocated by the recent benchmark \cite{wu2026survey}.

The color metrics show a different structure.
EV and SIRS are moderately correlated, while ICV is only weakly related to the two others
but probably due to its non-robustness to scale and illumination \cite{giraud2017}.
This indicates that color homogeneity is not captured by a single behavior in the metric set.
Yet, SIRS \cite{barcelos2022improving} was introduced to
address the non-robustness to slight color variations of EV.

Finally, the topology-related metrics form a more diverse family
containing  hierarchy-based and superpixel number and components related metrics.
Hierarchical metrics NE and IR exhibit strong correlation,
while RE remains correlated with them to a lesser extent.
By having a deeper look into these metrics,
NE and IR appear to form an almost degenerate metric pair.
Since they quantify complementary partitions of the same overlap space,
we empirically observe $\mathrm{NE} + \mathrm{IR} \approx 1$,
leading to near-perfect anti-correlation across datasets and methods
(see Appendix \ref{sec:appendix_metrics}).
The count-deviation metric $\Delta K$ and the connectivity-related criterion are less
strongly correlated, indicating that they describe
different aspects
also worth evaluating regarding the targeted validation.
This larger internal diversity may also partly explain why the effective dimensionality of the complete metric space slightly exceeds the five conceptual metric families considered in our taxonomy.

\paragraph{Multivariate structure of the metric space.}
We analyze in Fig.~\ref{fig:correlation_cat_bsd} the intrinsic structure of the metric space through principal component analysis (PCA), treating metrics as random variables defined over the joint distribution of methods, images, and scales on BSD.
The PCA participation-ratio effective dimension
$d_{\mathrm{eff}}=(\sum_k\lambda_k)^2/\sum_k\lambda_k^2$
estimates how many directions carry substantial variance.
The first components account for a large fraction of the total variance, but the curve increases progressively.
On BSD, the 20 metrics have an effective dimension
of $5.74$, and 9 principal components are required to explain 95\% of the variance, motivating a compact metric subset.

This result provides a conservative interpretation of metric redundancy.
The evaluation space is clearly structured and lower-dimensional than the raw number of reported metrics suggests.
This supports the view of superpixel evaluation as a structured multi-objective problem with partially correlated objectives,
as illustrated by Appendix \ref{sec:appendix_pca} on the seven datasets.

\paragraph{Value-space dependence and ranking consistency.}
We next assess in Fig.~\ref{fig:correlation_cat_bsd} the robustness of method rankings across scales using Kendall’s coefficient of concordance W, which measures agreement between multiple orderings \cite{teles2012concordance}.
The results reveal markedly different behaviors across criteria.
Regularity is highly stable across levels, with $W$ close to one over most of
the range, and accuracy also yields consistently stable rankings.
Color exhibits a moderate but relatively steady agreement, whereas topology
and especially contour show substantially lower stability.
This suggests that changes in scale do not affect all evaluation criteria in
the same way.
For some metrics, the relative ranking is largely preserved when
the number of superpixels varies, meaning that benchmark conclusions are robust
to the chosen level.
For others, however, rankings can change noticeably across scales, indicating
that a method may appear favorable only at a given resolution.
These results highlight that method comparison is inherently scale-dependent for several criteria, and advocate for multi-scale evaluation protocols rather than single operating-point assessments.

\section{Conclusion}

In this work, we introduced \SUPERVISE{}, a unified framework for standardized and reproducible superpixel benchmarking. Building on the lack of consistent evaluation protocols and the inherently ill-posed nature of superpixel segmentation, we argue that fair comparison requires jointly controlling two key axes: the evaluation scale and the choice of metrics.
By aligning methods using the actual number of generated superpixels, our protocol enables robust, scale-consistent comparisons beyond fixed operating points, while our large-scale analysis reveals strong metric dependencies, exposing redundancy and limitations of current evaluation practices.

These results advocate for a compact and representative metric subset. Beyond this, \SUPERVISE{} provides a scalable and fully reproducible benchmark through shared label maps and decoupled evaluation, currently covering 20 metrics, 31 methods, and 7 datasets, and designed to grow over time. We believe it offers a principled foundation for consistent and interpretable superpixel evaluation.
Extending the framework to spatio-temporal settings, including supervoxels and video segmentation, constitutes a natural direction for future work.

\begin{ack}
This work has been funded by the French Ministry of Higher Education and Research (MESR).
The authors declare no competing interests.
\end{ack}

\bibliography{main}
\newpage

\definecolor{metricaccuracy}{RGB}{105,56,135}
\definecolor{metriccontour}{RGB}{64,130,205}
\definecolor{metricregularity}{RGB}{210,70,70}
\definecolor{metriccolor}{RGB}{220,145,0}
\definecolor{metrictopology}{RGB}{0,155,145}

\definecolor{bestred}{RGB}{139,0,0}
\definecolor{secondblue}{RGB}{0,70,160}
\definecolor{thirdgreen}{RGB}{0,120,60}
\colorlet{aucaccuracy}{metricaccuracy!8}
\colorlet{auccontour}{metriccontour!8}
\colorlet{auccolor}{metriccolor!10}
\colorlet{aucregularity}{metricregularity!8}
\colorlet{auctopology}{metrictopology!9}

\appendix

\section{Technical appendices and supplementary material}

\subsection{Datasets and annotations}
\label{sec:appendix_dataset}
\begin{figure*}[t!]
	\centering
	\setlength{\tabcolsep}{1pt}
	\renewcommand{\arraystretch}{0.9}
	\begin{tabular}{@{}c@{\hspace{3pt}}c@{\hspace{2pt}}c@{\hspace{5pt}}c@{\hspace{2pt}}c@{}}
		\rotatebox{90}{\hspace{1cm}\textbf{BSD}}                                                                        &
		\includegraphics[width=0.215\textwidth]{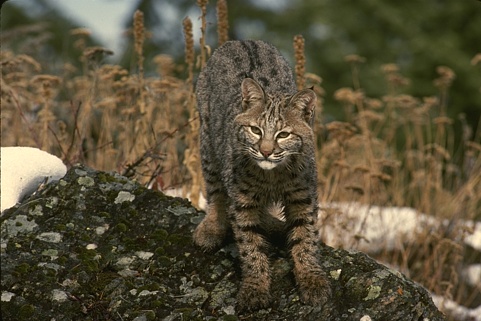}                                                         & \includegraphics[width=0.215\textwidth]{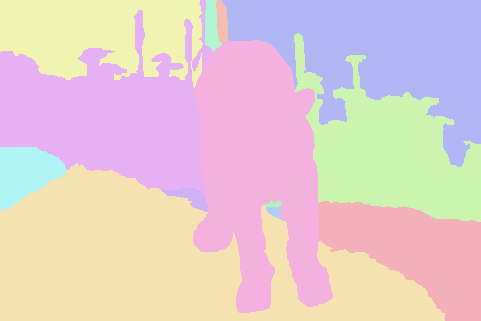}                                                 & \includegraphics[width=0.215\textwidth]{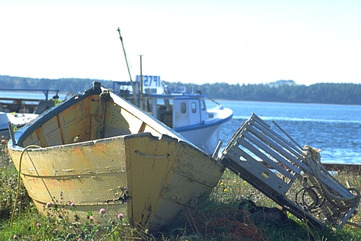}                          & \includegraphics[width=0.215\textwidth]{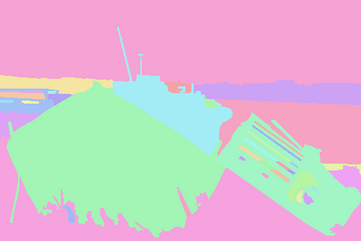}
		\\[0.5ex]
		\rotatebox{90}{\hspace{0.75cm}\textbf{SBD}}                                                                     &
		\includegraphics[width=0.215\textwidth]{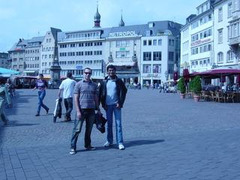}                     & \includegraphics[width=0.215\textwidth]{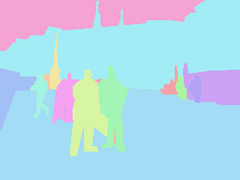}                     & \includegraphics[width=0.215\textwidth]{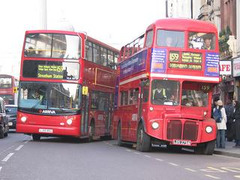}                     & \includegraphics[width=0.215\textwidth]{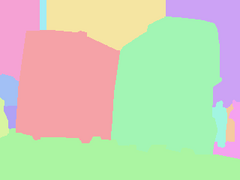}
		\\[0.5ex]
		\rotatebox{90}{\hspace{0.6cm}\textbf{NYUv2}}                                                                    &
		\includegraphics[width=0.215\textwidth]{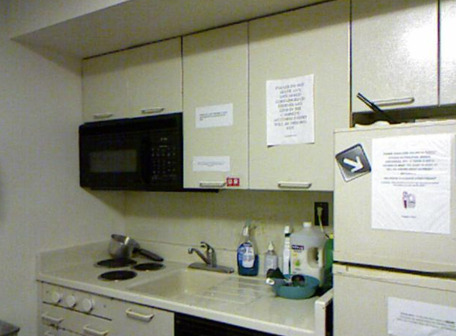}                  & \includegraphics[width=0.215\textwidth]{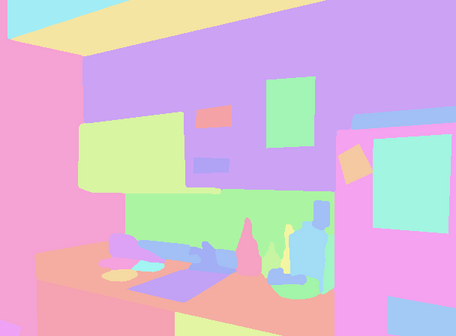}                  & \includegraphics[width=0.215\textwidth]{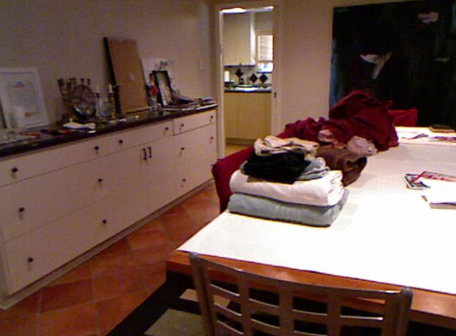}                  & \includegraphics[width=0.215\textwidth]{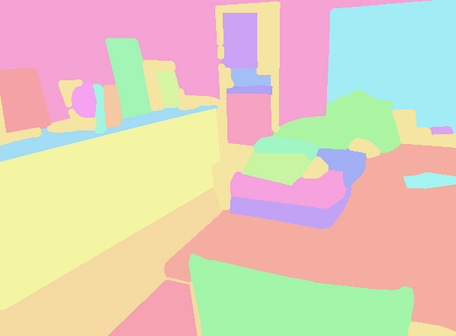}
		\\[0.5ex]
		\rotatebox{90}{\textbf{KITTI}}                                                                                  &
		\includegraphics[width=0.215\textwidth]{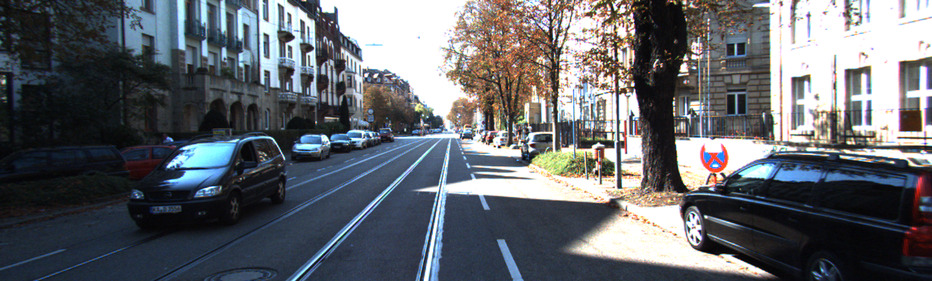}                 & \includegraphics[width=0.215\textwidth]{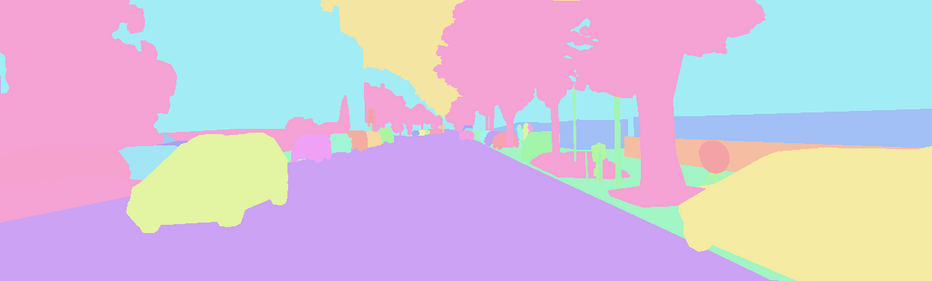}                 & \includegraphics[width=0.215\textwidth]{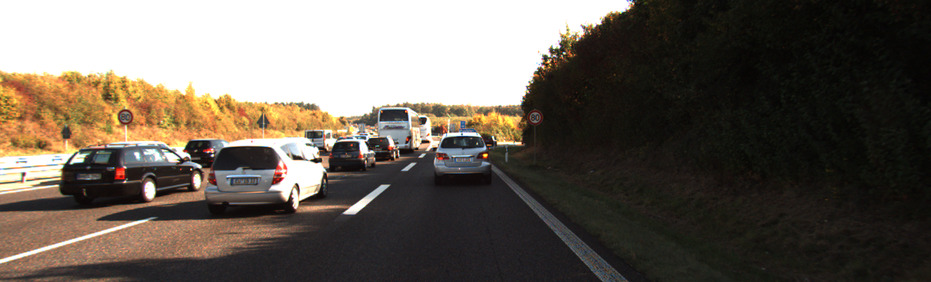}                 & \includegraphics[width=0.215\textwidth]{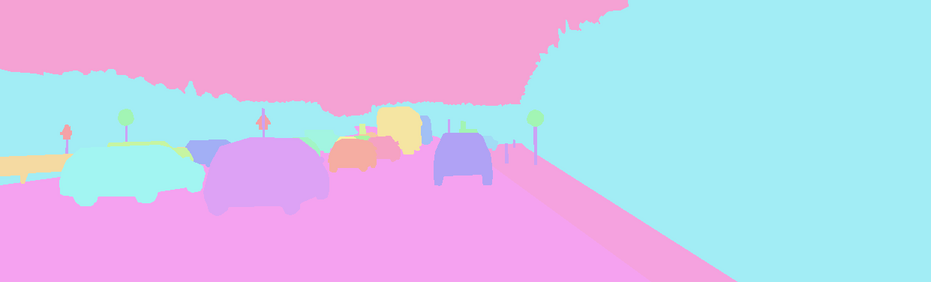}
		\\[0.5ex]
		\rotatebox{90}{\hspace{0.5cm}\textbf{PSD}}                                                                      &
		\includegraphics[width=0.215\textwidth]{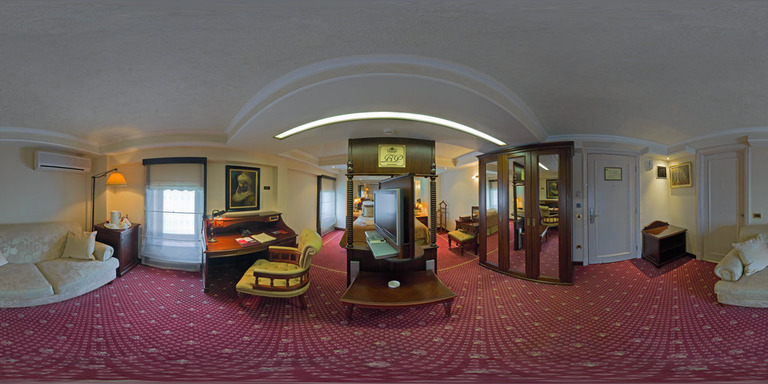}                           & \includegraphics[width=0.215\textwidth]{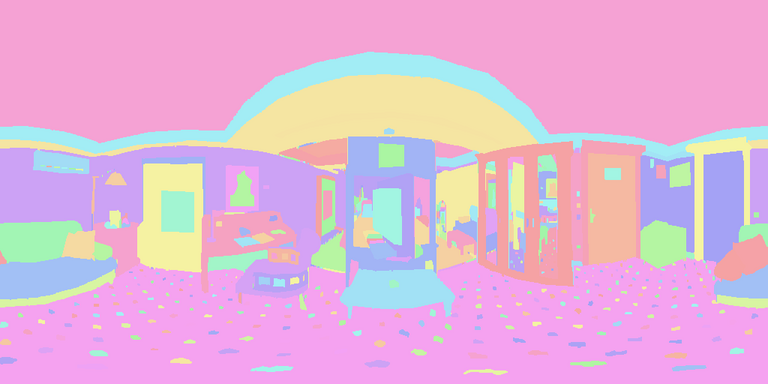}                           & \includegraphics[width=0.215\textwidth]{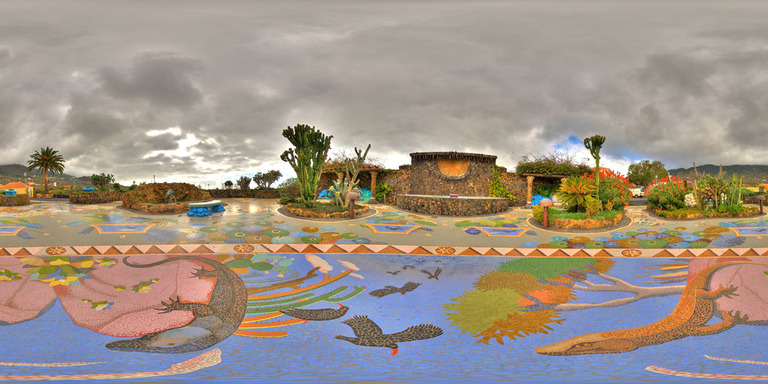}                          & \includegraphics[width=0.215\textwidth]{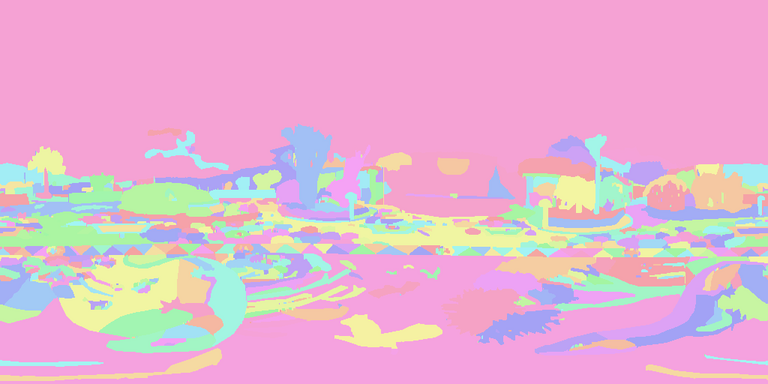}
		\\[0.5ex]
		\rotatebox{90}{\hspace{1cm}\textbf{DRIVE}}                                                                      &
		\includegraphics[width=0.215\textwidth]{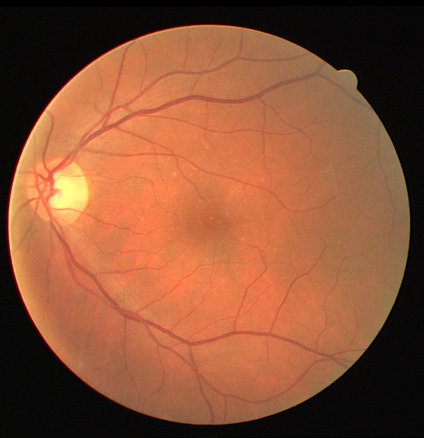}                        & \includegraphics[width=0.215\textwidth]{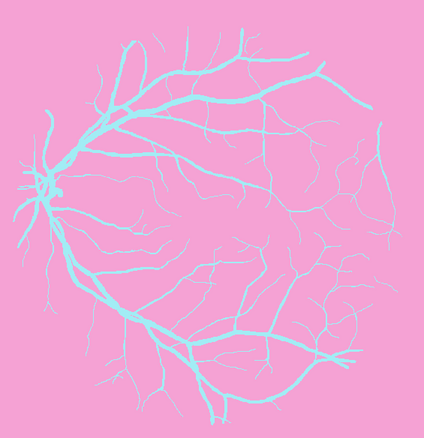}                        & \includegraphics[width=0.215\textwidth]{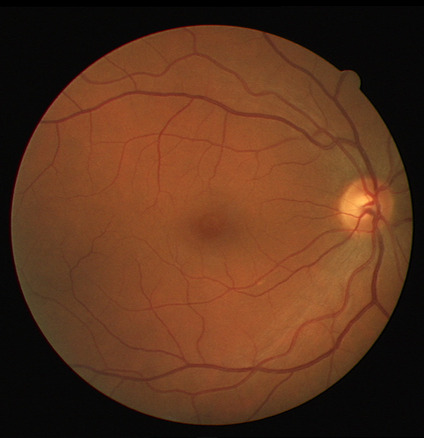}                        & \includegraphics[width=0.215\textwidth]{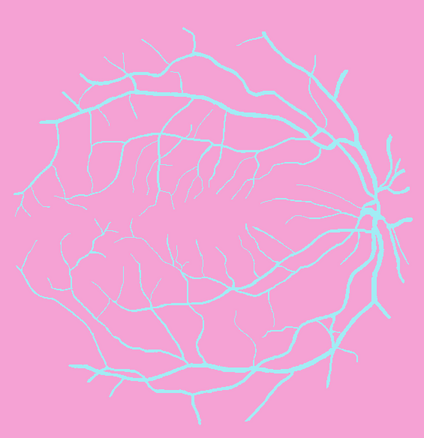}
		\\[0.5ex]
		\rotatebox{90}{\hspace{0.5cm}\textbf{MONUSEG}}                                                                  &
		\includegraphics[width=0.215\textwidth]{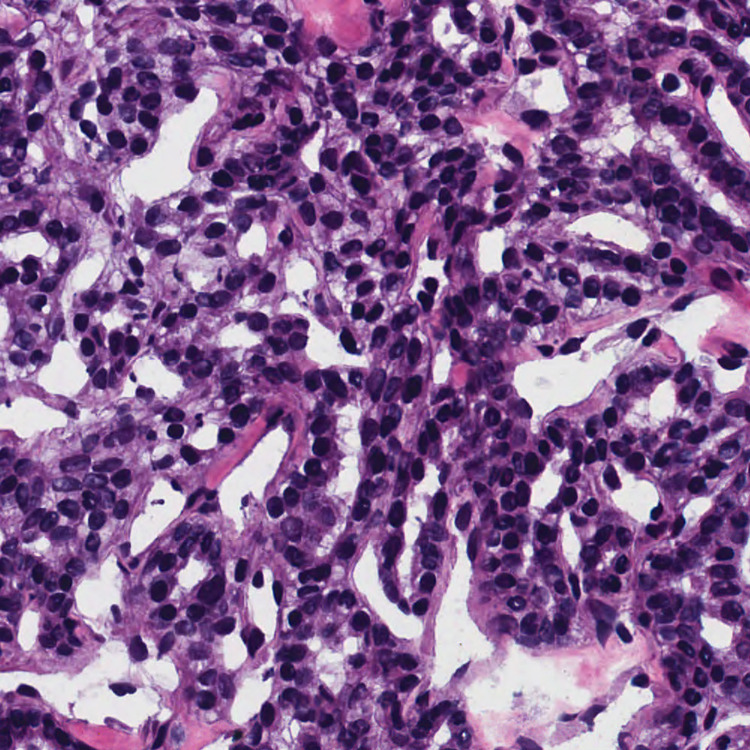} & \includegraphics[width=0.215\textwidth]{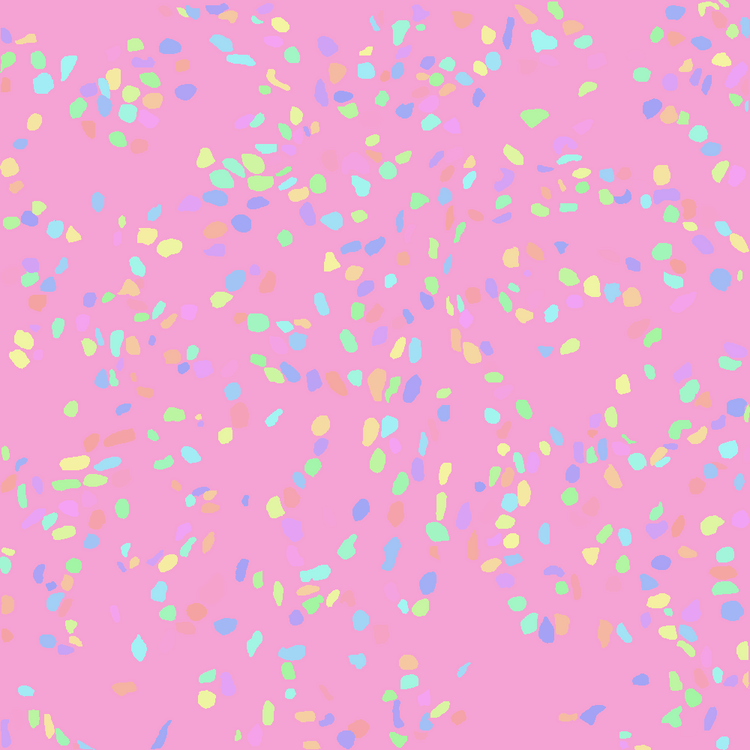} & \includegraphics[width=0.215\textwidth]{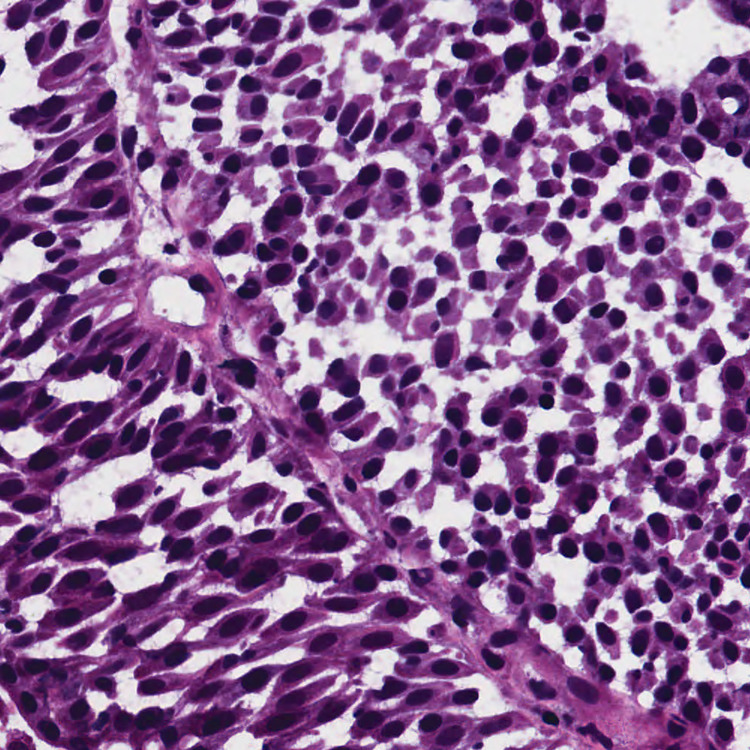} & \includegraphics[width=0.215\textwidth]{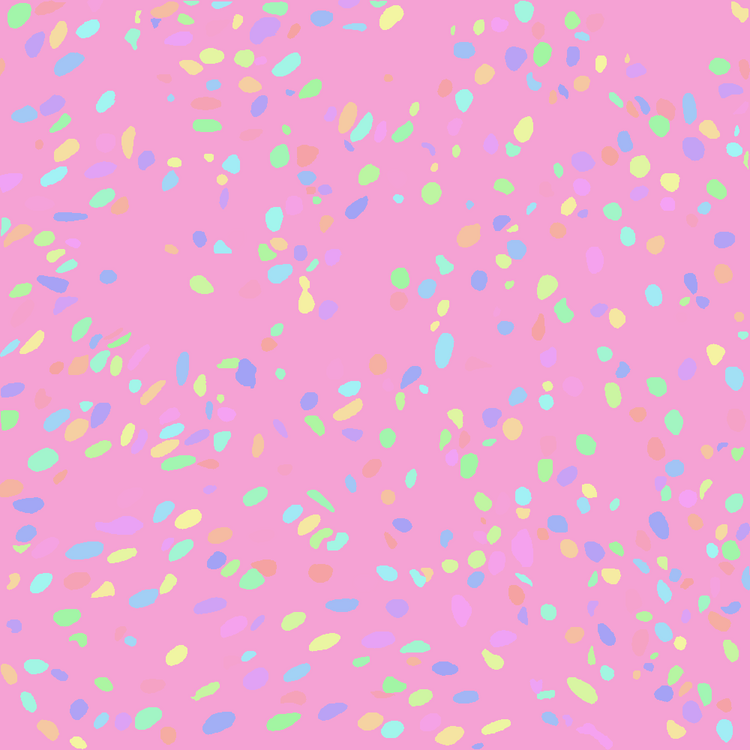}                              \\[-0.5ex]
		                                                                                                                & {\scriptsize Image }                                                                                         & {\scriptsize Groundtruth }                                                                                      & {\scriptsize Image }                                                                                         & {\scriptsize Groundtruth } \\
	\end{tabular}
	\caption{\textbf{Examples from the datasets used in the benchmark:} BSD, SBD, NYUv2, KITTI, PSD, DRIVE, MONUSEG.}
	\label{fig:supp_dataset_examples_2}
\end{figure*}

The proposed benchmark covers natural, indoor, driving, panoramic, and biomedical image
domains. Fig.~\ref{fig:supp_dataset_examples_2} shows representative images
and annotations, while Tab.~\ref{tab:supp_dataset_summary} summarizes the
number of test images, typical image sizes, and annotation density for each
dataset.

The dataset selection is
heterogeneous. BSD, SBD, and NYUv2 cover
generic object and scene understanding settings; KITTI focuses on road scenes
with strong perspective effects; PSD stresses wide panoramic geometry; DRIVE
contains thin retinal structures; and MONUSEG contains dense biomedical
instances. This diversity is important because a superpixel method that behaves
well on natural images may fail to preserve thin structures, dense object
boundaries, or the requested number of regions in another imaging domain.

We use the annotations originally distributed with each dataset without modifying their content, and only convert them to a common label map representation. When multiple annotations are available, they are all retained. In particular, BSD provides 4–8 annotations per image, for which metric scores are computed for each annotation and averaged across annotators, following standard superpixel evaluation practice.

\begin{table*}[h]
	\centering
	\caption{\textbf{Summary of the datasets used in the benchmark.} }
	\label{tab:supp_dataset_summary}
	\resizebox{\textwidth}{!}{%
		\begin{tabular}{llccc@{\hspace{2mm}}c@{\hspace{2mm}}ccl}
			\toprule
			                                 &                                   &                      &                       & \multicolumn{3}{c}{\textbf{Number of objects}} &                             &                                                                                    \\
			\textbf{Dataset}                 & \textbf{Domain}                   & \textbf{Test images} & \textbf{Average size} & \textbf{Mean}                                  & \textbf{Min}                & \textbf{Max} & \textbf{GT / image} & \textbf{Notes}                                \\
			\midrule
			BSD \cite{martin2001database}    & Outdoor scenes                    & 200                  & 321x481                                                                                                                                                                                   %
			                                 & 22                                & 3                    & 69                    & 4-8                                            & Fine multi-user annotations                                                                                      \\
			SBD \cite{gould2009decomposing}  & Outdoor scenes                    & 477                  & 320x240               & 20                                             & 2                           & 135          & 1                   & PASCAL-style segmentation masks               \\
			NYUv2 \cite{silberman2012indoor} & Indoor scenes                     & 399                  & 608x448               & 25                                             & 5                           & 79           & 1                   & Depth files available                         \\
			KITTI \cite{Alhaija2018IJCV}     & Driving                           & 200                  & 1242x375              & 21                                             & 6                           & 40           & 1                   & Driving scenes                                \\
			PSD \cite{wan2018spherical}      & Panoramic 360° scenes & 75                   & 1024x512              & 510                                            & 115                         & 1085         & 1                   & Wide panoramic images                         \\
			DRIVE \cite{staal2004}           & Medical images of retina          & 20                   & 565x584               & 2                                              & 2                           & 2            & 1                   & Binary medical annotations of thin structures \\
			MONUSEG \cite{kumar2017dataset}  & Medical images of cells           & 14                   & 1000x1000             & 479                                            & 250                         & 819          & 1                   & Large images with many objects                \\
			\bottomrule
		\end{tabular}%
	}
\end{table*}
\subsection{Metric families and evaluation criteria}
\label{sec:appendix_metrics}

\begin{figure}[t]
	\centering
	\setlength{\tabcolsep}{1pt}
	{\scriptsize
		\begin{tabular}{@{}cccc@{}}
			                                                                 & SLIC \cite{achanta2012}
			                                                                 & BASS~\cite{uziel2019bass}
			                                                                 & CDS~\cite{xu2024cds}              \\
			\includegraphics[width=0.23\textwidth]{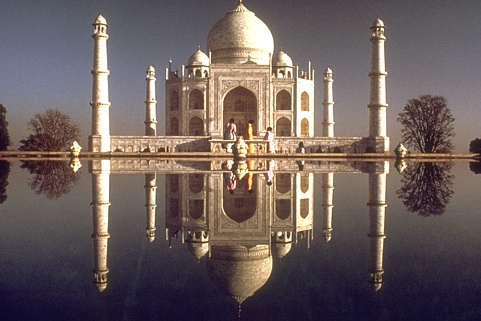}        &
			\includegraphics[width=0.23\textwidth]{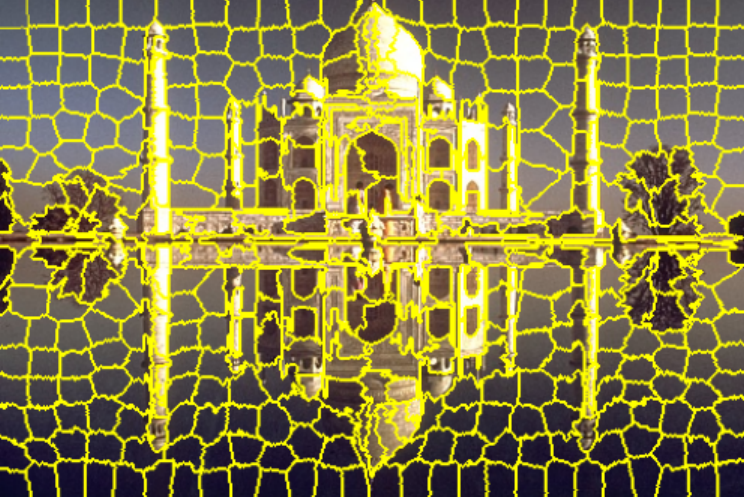} &
			\includegraphics[width=0.23\textwidth]{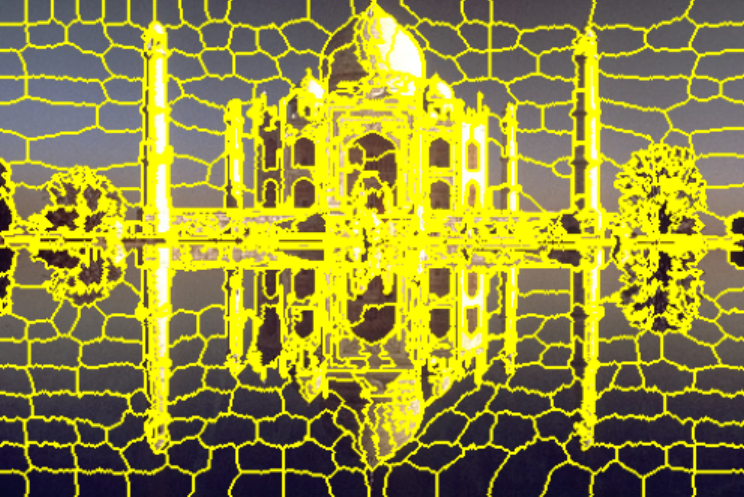} &
			\includegraphics[width=0.23\textwidth]{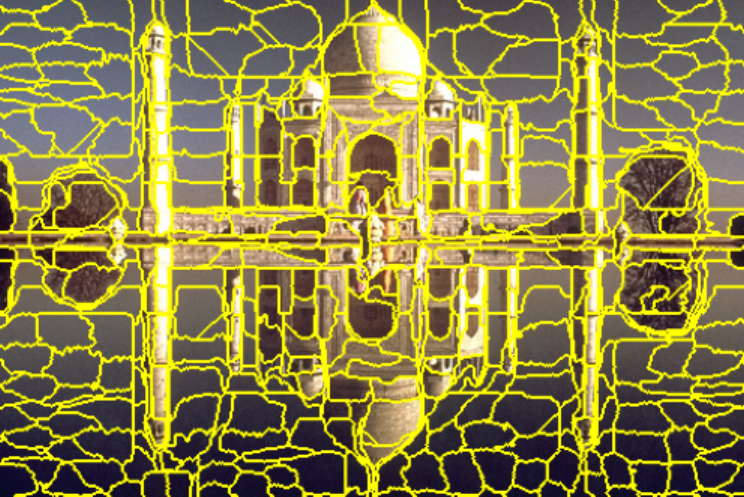}                                      \\
			Image                                                            &
			Accuracy:	 ASA = 0.9158                                          &
			Accuracy:	ASA = 0.9250                                           &
			\textbf{Accuracy:	 ASA = 0.9501}                                                                     \\
			                                                                 & \textbf{Regularity: GR =0.3978}
			                                                                 & Regularity: GR = 0.2477         &
			Regularity: GR = 0.3060                                                                              \\
			                                                                 &
			Color: EV = 0.8383                                               &
			\textbf{Color: EV = 0.9180}                                      &
			Color:  EV = 0.7948                                                                                  \\
		\end{tabular}
	}
	\caption{
		\textbf{Behavior trade-offs in superpixel methods} on a BSD image for $k=500$ superpixels.
		SLIC is the most regular method, while BASS focuses on color homogeneity and CDS on  accuracy.
	}
	\label{fig:example_metrics}
\end{figure}

Superpixel methods optimize different and sometimes conflicting properties. We
therefore group metrics into five families: accuracy, contour adherence, color
homogeneity, regularity, and topology. Fig.~\ref{fig:example_metrics}
illustrates such trade-offs on a representative image, and
Tab.~\ref{tab:metrics_summary} lists the metrics considered in the benchmark
together with their interpretation and reporting recommendation.

The accuracy family measures agreement with groundtruth regions after assigning each superpixel to annotated objects.
The contour family evaluates whether
superpixel boundaries accurately recover the annotated contours.
Color metrics quantify how
homogeneous the image appearance remains inside each superpixel. Regularity metrics capture geometric properties such as compactness, smoothness, and shape consistency. Finally,
topology metrics evaluate hierarchical consistency,
connectedness, and deviations from the requested number of superpixels.

\begin{wrapfigure}{r}{0.43\linewidth}
	\centering
	\vspace{-1.2em}
	\includegraphics[width=0.78\linewidth]{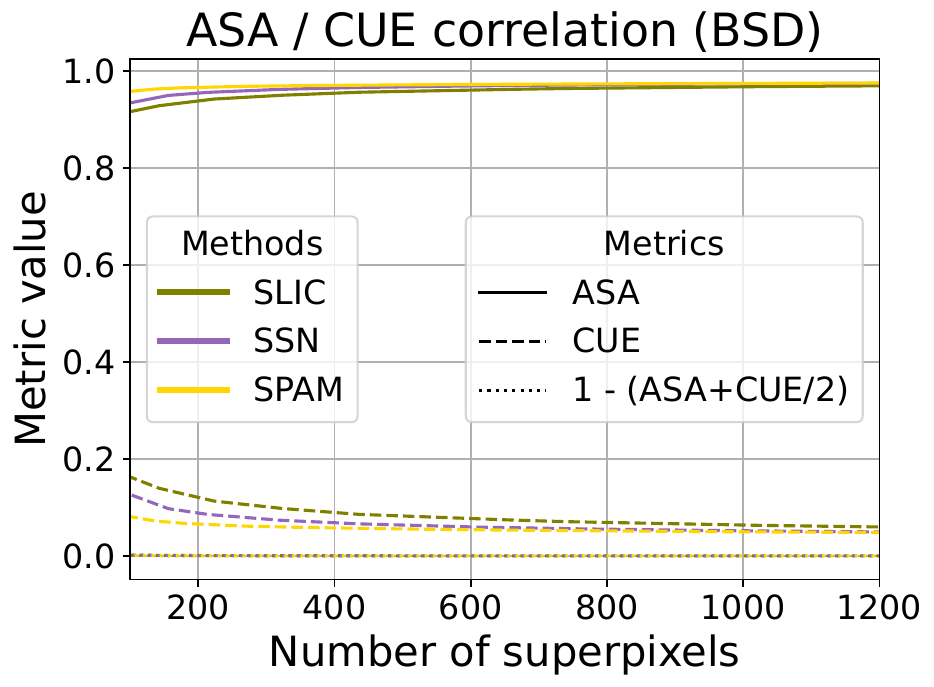}
	\caption{ASA, CUE and $1-(\text{ASA}+\text{CUE}/2)$ across scales for methods on BSD.}
	\label{fig:asa_cue}
	\vspace{-0.4em}
\end{wrapfigure}

\paragraph{Redundancy between ASA and CUE.}
We empirically analyze the relationship between Achievable Segmentation Accuracy (ASA) and Undersegmentation Error (CUE), two metrics that have long been identified as structurally linked in the superpixel literature \cite{giraud2017,stutz2018}.
As shown in Fig.~\ref{fig:asa_cue}, both quantities exhibit a deterministic coupling across methods and scales, following the relationship $\mathrm{ASA} = 1 - \text{CUE}/{2}$.
This direct affine dependency explains the observed near-perfect anti-correlation between the two metrics and confirms that they encode equivalent information about segmentation quality.
As a consequence, reporting both ASA and CUE is redundant, and one of them can be safely omitted without loss of information.

\begin{figure*}[t]
	\centering
	{\footnotesize
		\begin{tabular}{@{}c@{}c@{}c@{}c@{}}
			\centering
			\includegraphics[width=0.235\linewidth]{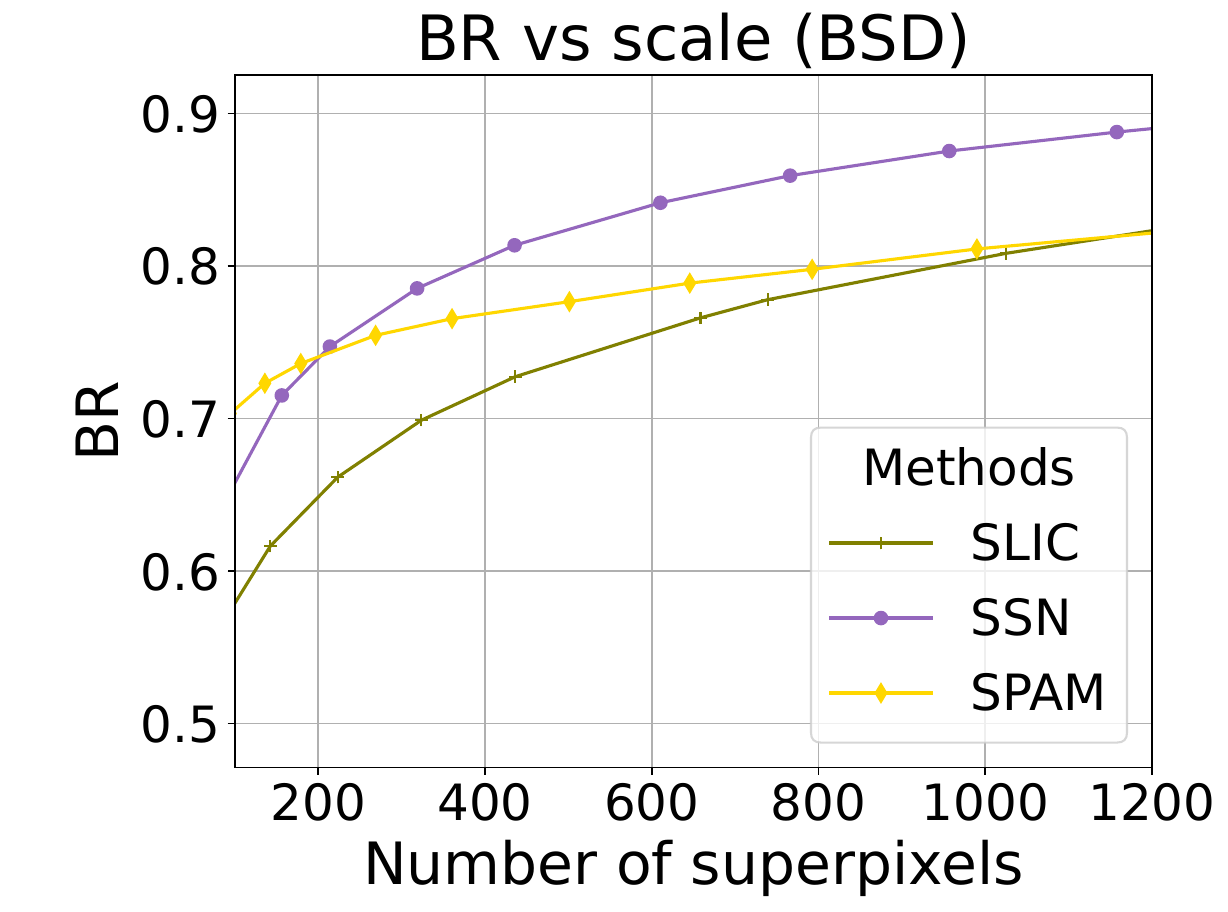} &
			\includegraphics[width=0.235\linewidth]{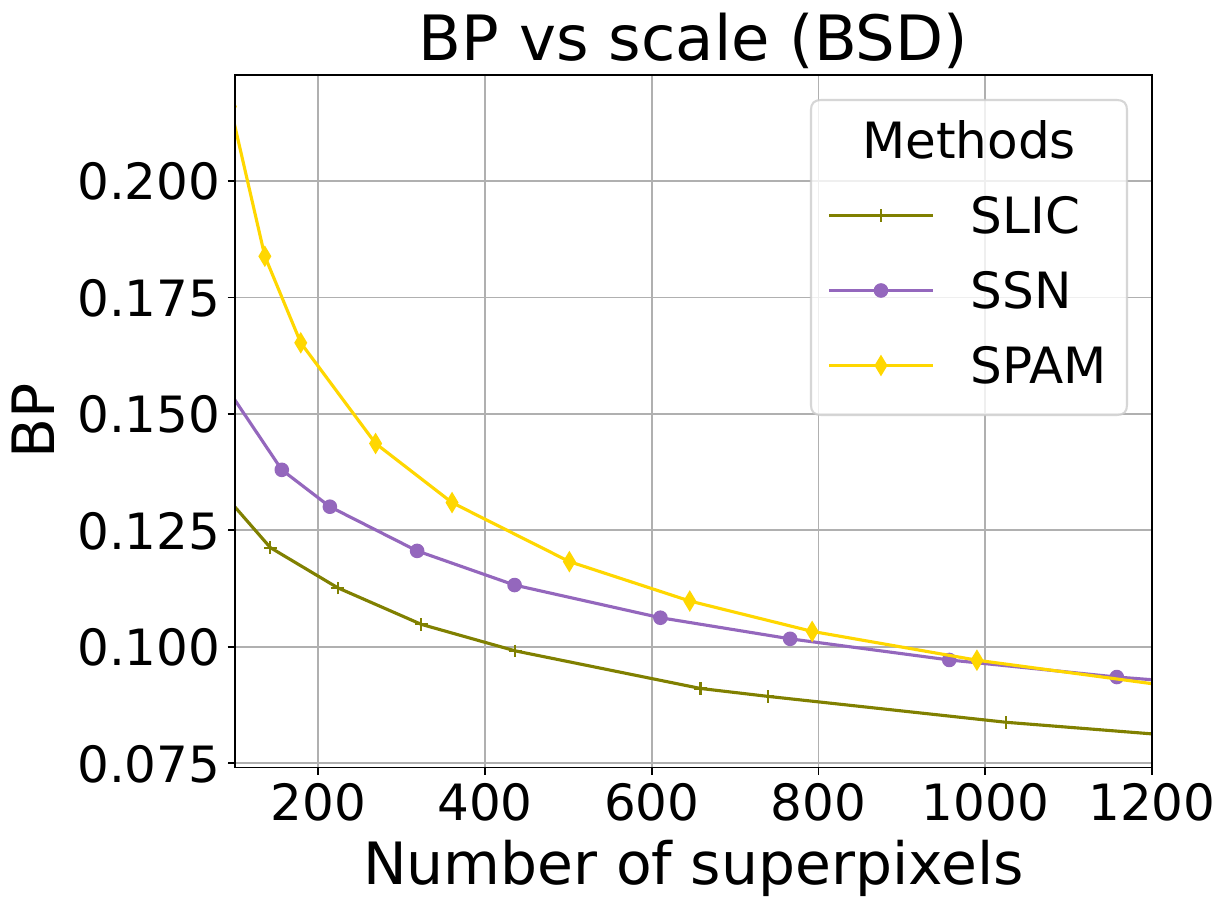} &
			\includegraphics[width=0.235\linewidth]{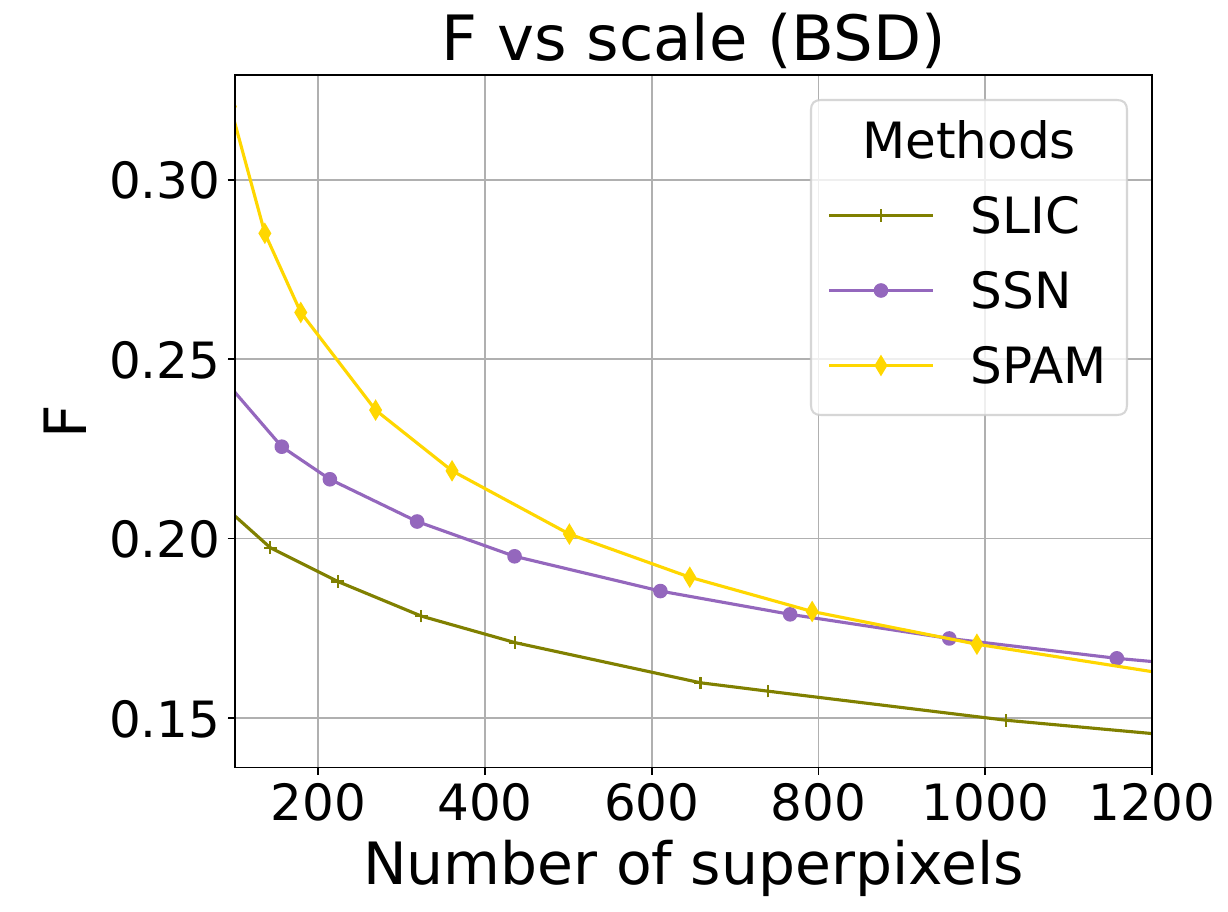}  &
			\includegraphics[width=0.235\linewidth]{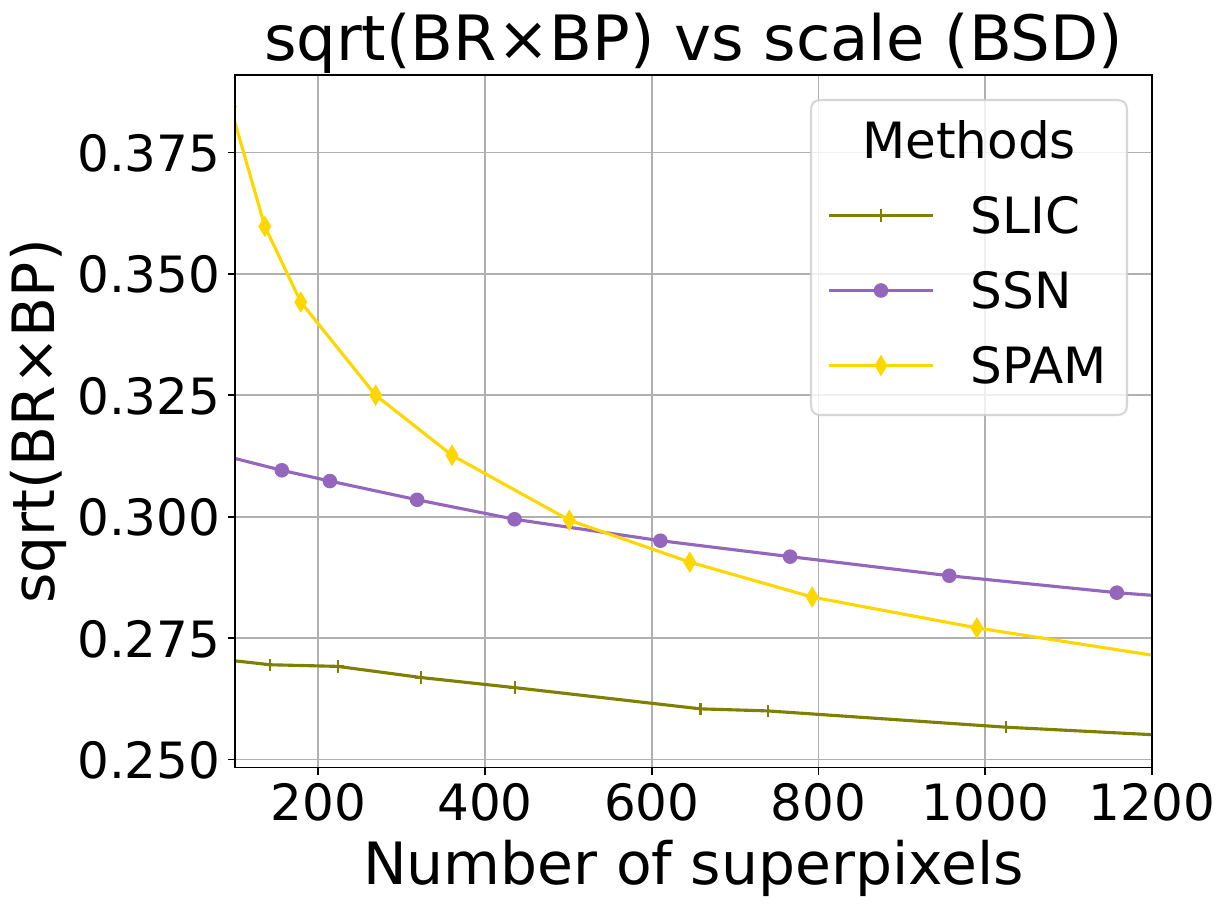}           \\
			BR                                                     & BP & F-measure &
			{\brbp{}}
		\end{tabular}
	}
	\caption{\textbf{Comparison of boundary-based metrics in superpixel evaluation} for methods on BSD. The proposed geometric mean provides a more balanced view of precision–recall trade-offs compared to individual BR, BP, and their harmonic aggregation in F-measure.}
	\label{fig:br_vs_bp}
\end{figure*}

\paragraph{Redundancy between BP and F.}
Regarding boundary-based metrics, Boundary Precision (BP) and the F-measure are tightly coupled due to the intrinsic structure of the task, \emph{i.e.}, providing an image oversegmentation with low BP and high Boundary Recall (BR).
As a result, reporting F and BP would be redundant.
Also reporting BP or BR in isolation naturally provides limited insight (see Fig.~\ref{fig:br_vs_bp}).
Reporting standard BR vs BP performance enables assessment of the method's behavior.
However, such representation leads to non-common range between methods, making AUC scores and related method ranking irrelevant.
To address this issue, we introduce the geometric mean of BR and BP as a balanced alternative as \brbp{}, providing a symmetric measure that better captures the precision–recall trade-off while reducing scale imbalance effects.

\begin{wrapfigure}{r}{0.37\linewidth}
	\centering
	\vspace{-1.1em}
	\includegraphics[width=0.8\linewidth]{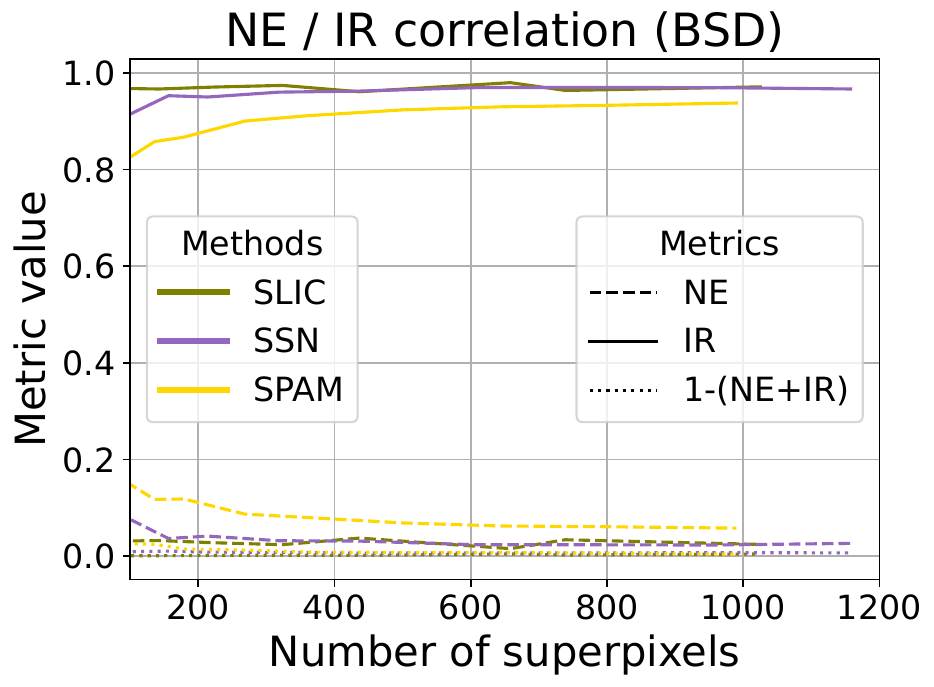}
	\caption{NE, IR, and $1-(\text{NE}+\text{IR})$ across scales for methods on BSD.}
	\label{fig:ne_ir}
	\vspace{-0.4em}
\end{wrapfigure}

\newpage
\paragraph{Redundancy between NE and IR.}
We empirically analyze the relationship between Nestedness (NE) and Inflation Ratio (IR) on BSD by plotting their evolution across scales for several representative methods (SLIC, SSN, SPAM). As shown in Fig.~\ref{fig:ne_ir}, both metrics exhibit a near-perfect anti-correlation over the full scale range, with $\mathrm{NE} + \mathrm{IR} \approx 1$. This behavior is consistent across methods and confirms that they capture complementary partitions of the same overlap structure. As a result, reporting both metrics is largely redundant, and one can be safely omitted without loss of information.

\begin{table}[t]
	\centering
	\caption{
		\textbf{Summary of the evaluation metrics considered in our framework.}
		This table consolidates interpretations and claims from prior works \cite{stutz2018,barcelos2024,giraud2024,wu2026survey}, which are revisited and empirically validated through our large-scale correlation analysis.
		It further provides recommendations (high/medium/low) toward a compact and representative metric subset based
		on metric intrinsic properties and redundancy compared to others.
		"High" indicates the least biased metric enabling relevant evaluation of a superpixel property, "medium" a good metric that has a better alternative in the category, and "low" a more strongly biased metric that could lead to misinterpretations.
	}
	{\footnotesize
		\setlength{\tabcolsep}{3.4pt}
		\renewcommand{\arraystretch}{1}
\begin{tabular}{@{}lp{0.24\linewidth}lp{0.44\linewidth}c@{}}
			\toprule
			\textbf{Metric}                                                                                                                                                                        & \textbf{Name} & \textbf{Ref.} & \textbf{What it measures. Comments} & \textbf{Recommendation} \\
			\midrule

			\multicolumn{4}{@{}l}{\textit{\textcolor{metricaccuracy}{Accuracy}}}                                                                                                                                                                                                                   \\
			\midrule
			\textcolor{metricaccuracy}{ASA} \upp                                                                                                                                                   &
			Achievable Segmentation Accuracy                                                                                                                                                       &
			\cite{liu2011}
			                                                                                                                                                                                       &
			Upper bound on segmentation accuracy obtained by assigning each superpixel to its dominant ground-truth region. Standard accuracy metric.                                              &
			high                                                                                                                                                                                                                                                                                   \\
\bottomrule
		\end{tabular}
	}
	\label{tab:metrics_summary}
\end{table}

\newcommand{\heee}{0.2\textwidth}

\begin{table*}[t]
	\centering
	\scriptsize
	\setlength{\tabcolsep}{2.3pt}
	\caption{\textbf{AUC values for all metrics on the BSD dataset}. Best, second-best, and third-best values per metric are highlighted.}
	\label{tab:auc-values}
	\resizebox{\textwidth}{!}{%
		\begin{tabular}{l>{\columncolor{aucaccuracy}}r>{\columncolor{aucaccuracy}}r>{\columncolor{aucaccuracy}}r>{\columncolor{auccontour}}r>{\columncolor{auccontour}}r>{\columncolor{auccontour}}r>{\columncolor{auccontour}}r>{\columncolor{auccolor}}r>{\columncolor{auccolor}}r>{\columncolor{auccolor}}r>{\columncolor{aucregularity}}r>{\columncolor{aucregularity}}r>{\columncolor{aucregularity}}r>{\columncolor{aucregularity}}r>{\columncolor{aucregularity}}r>{\columncolor{auctopology}}r>{\columncolor{auctopology}}r>{\columncolor{auctopology}}r>{\columncolor{auctopology}}r>{\columncolor{auctopology}}r}
			\toprule
			                                     & \multicolumn{3}{c}{\textcolor{metricaccuracy}{Accuracy}} & \multicolumn{4}{c}{\textcolor{metriccontour}{Contour}} & \multicolumn{3}{c}{\textcolor{metriccolor}{Color}} & \multicolumn{5}{c}{\textcolor{metricregularity}{Regularity}} & \multicolumn{5}{c}{\textcolor{metrictopology}{Topology}}                                                                                                                                                                                                                                                                                                                                                                                                                                                                                                                                                                                                                    \\
\bottomrule
		\end{tabular}%
	}
\end{table*}

\begin{table*}[t]
	\centering
	\scriptsize
	\setlength{\tabcolsep}{2.3pt}
	\caption{\textbf{Method rankings by AUC for all metrics on the BSD dataset}. The last column of each family is the family rank.}
	\label{tab:auc-ranking}
	\resizebox{\textwidth}{!}{%
		\begin{tabular}{l>{\columncolor{aucaccuracy}}r>{\columncolor{aucaccuracy}}r>{\columncolor{aucaccuracy}}r>{\columncolor{aucaccuracy}}r>{\columncolor{auccontour}}r>{\columncolor{auccontour}}r>{\columncolor{auccontour}}r>{\columncolor{auccontour}}r>{\columncolor{auccontour}}r>{\columncolor{auccolor}}r>{\columncolor{auccolor}}r>{\columncolor{auccolor}}r>{\columncolor{auccolor}}r>{\columncolor{aucregularity}}r>{\columncolor{aucregularity}}r>{\columncolor{aucregularity}}r>{\columncolor{aucregularity}}r>{\columncolor{aucregularity}}r>{\columncolor{aucregularity}}r>{\columncolor{auctopology}}r>{\columncolor{auctopology}}r>{\columncolor{auctopology}}r>{\columncolor{auctopology}}r>{\columncolor{auctopology}}r>{\columncolor{auctopology}}r}
			\toprule
			                                     & \multicolumn{4}{c}{\textcolor{metricaccuracy}{Accuracy}} & \multicolumn{5}{c}{\textcolor{metriccontour}{Contour}} & \multicolumn{4}{c}{\textcolor{metriccolor}{Color}} & \multicolumn{6}{c}{\textcolor{metricregularity}{Regularity}} & \multicolumn{6}{c}{\textcolor{metrictopology}{Topology}}                                                                                                                                                                                                                                                                                                                                                                                                                                                                                                                                                                                                                                                                                                                                                      \\
\bottomrule
		\end{tabular}%
	}
\end{table*}

\subsection{Comprehensive quantitative evaluation on the BSD}
\label{sec:app_bsd_result}

\begin{figure*}[t!]
	\centering
	\begingroup
	\setlength{\tabcolsep}{1.5pt}
	\renewcommand{\arraystretch}{0.85}
	{\scriptsize
		\begin{tabular}{ccccc}
			\includegraphics[width=0.19\textwidth]{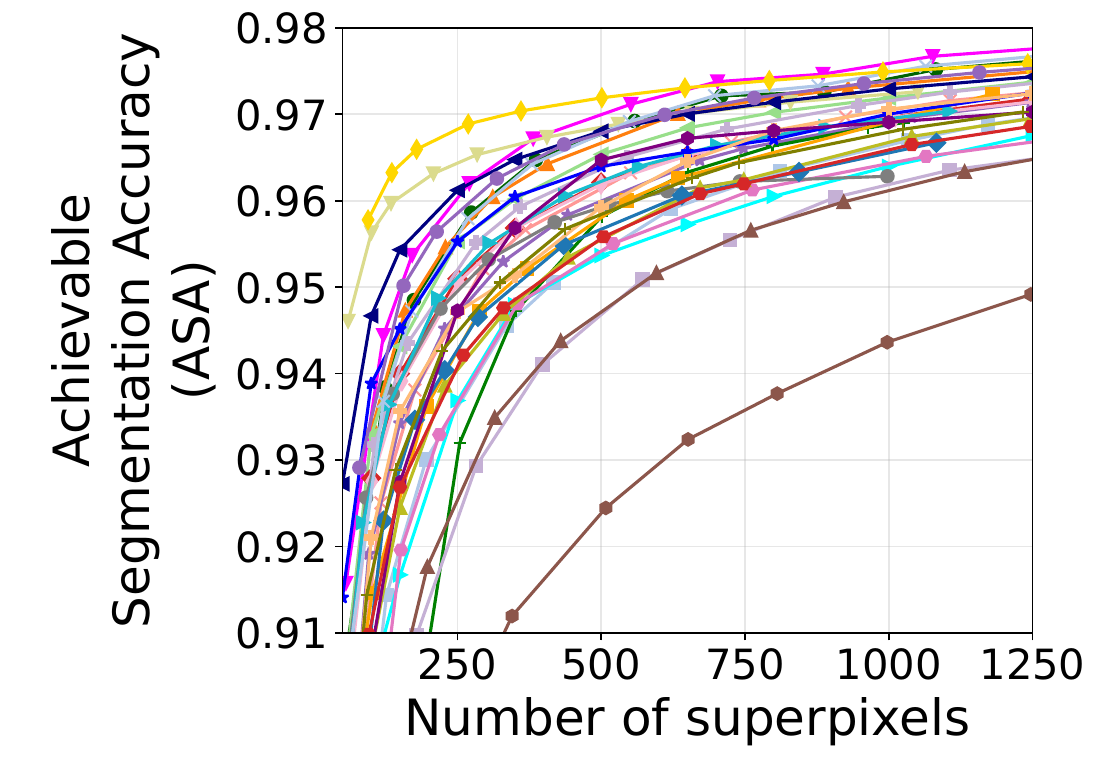}  &
			\includegraphics[width=0.19\textwidth]{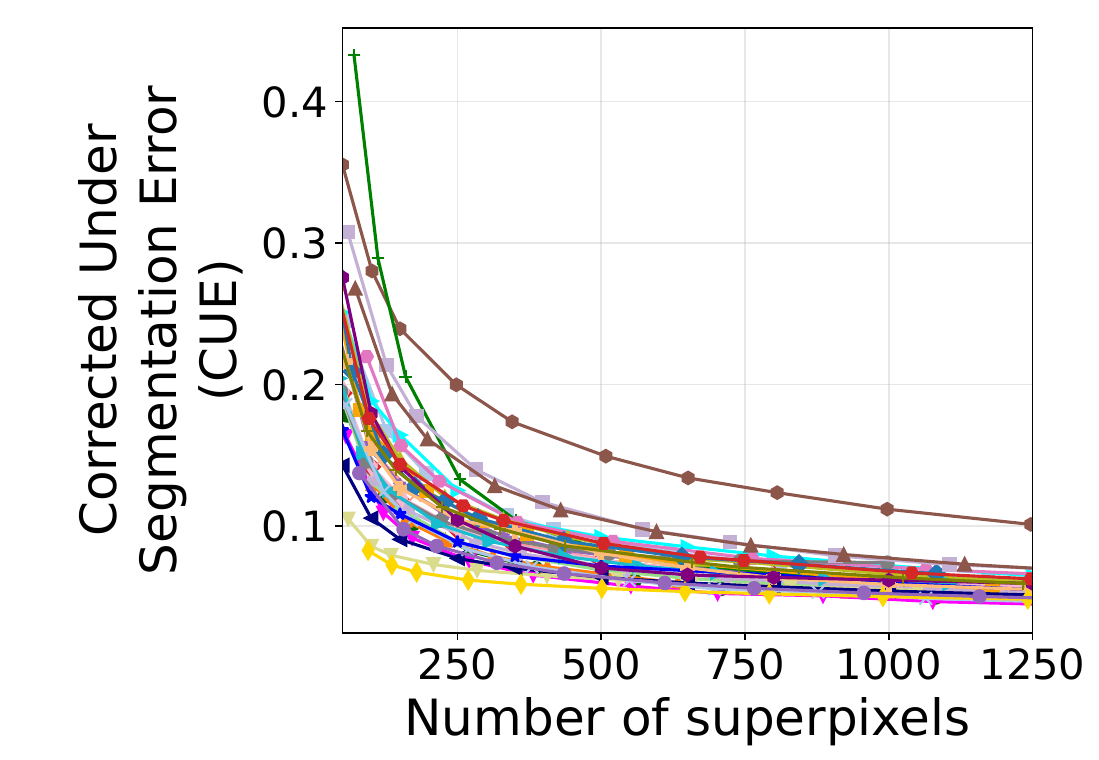}  &
			\includegraphics[width=0.19\textwidth]{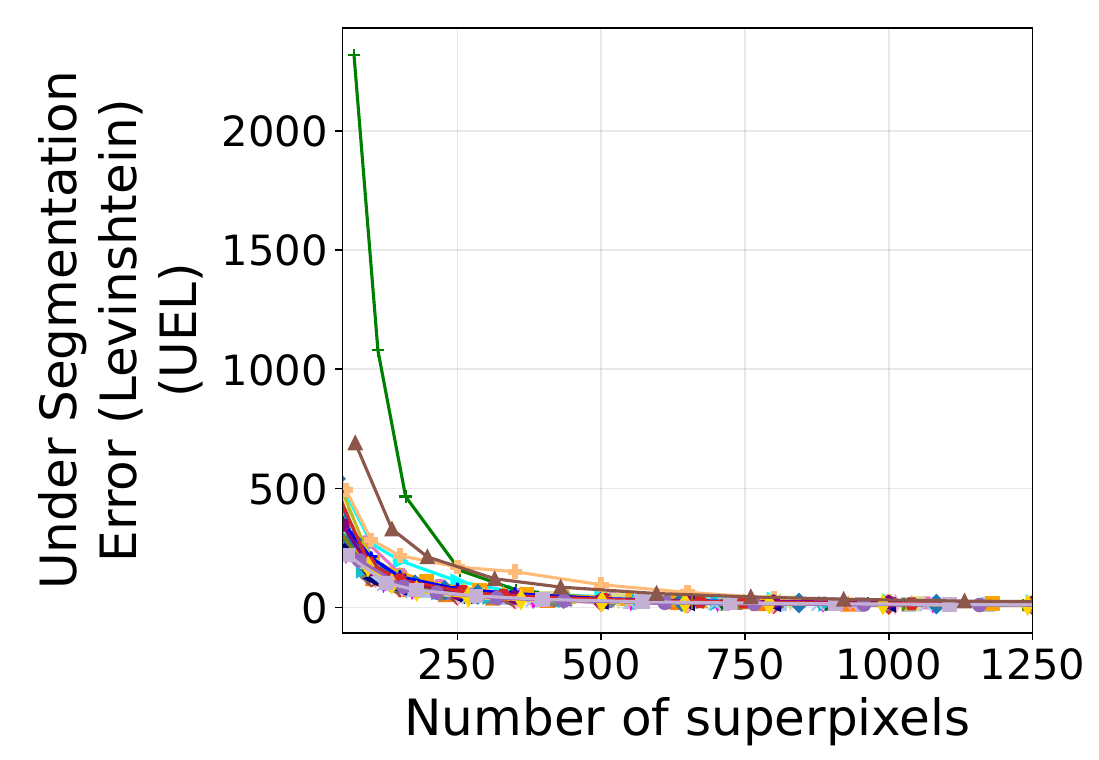}  &
			\includegraphics[width=0.19\textwidth]{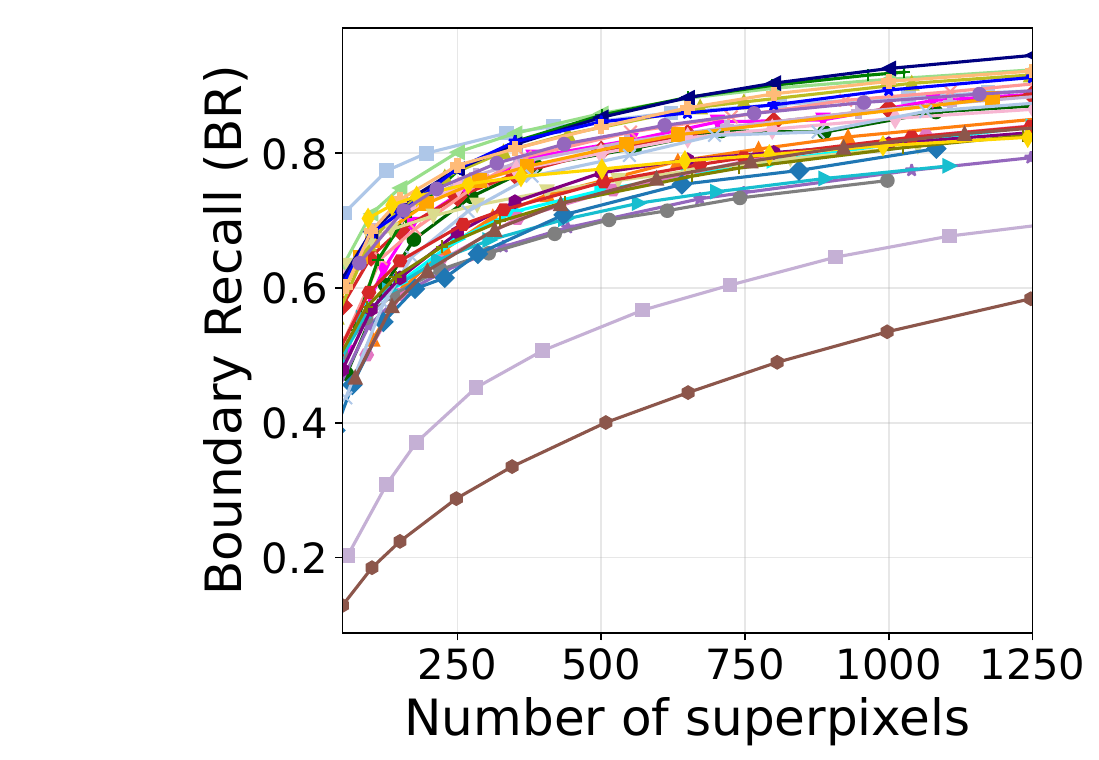}   &
			\includegraphics[width=0.19\textwidth]{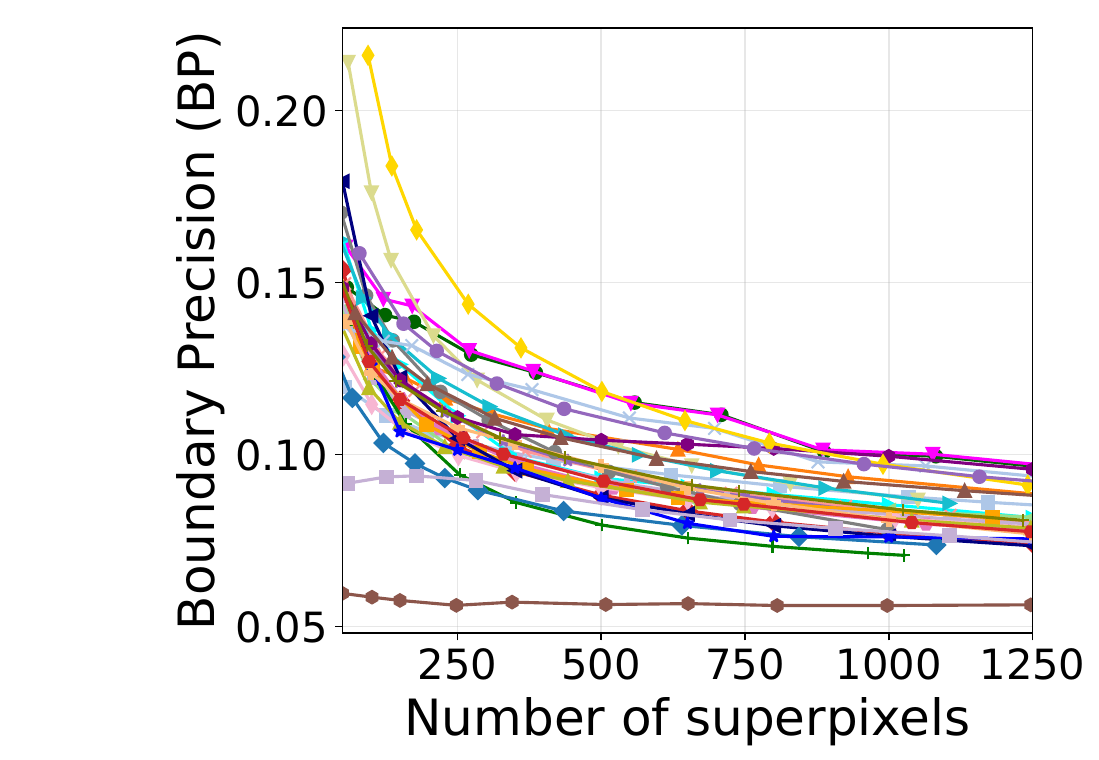}     \\
			\textcolor{metricaccuracy}{ASA}                                       &
			\textcolor{metricaccuracy}{CUE}                                       &
			\textcolor{metricaccuracy}{UEL}                                       &
			\textcolor{metriccontour}{BR}                                         &
			\textcolor{metriccontour}{BP}                                           \\
			\includegraphics[width=0.19\textwidth]{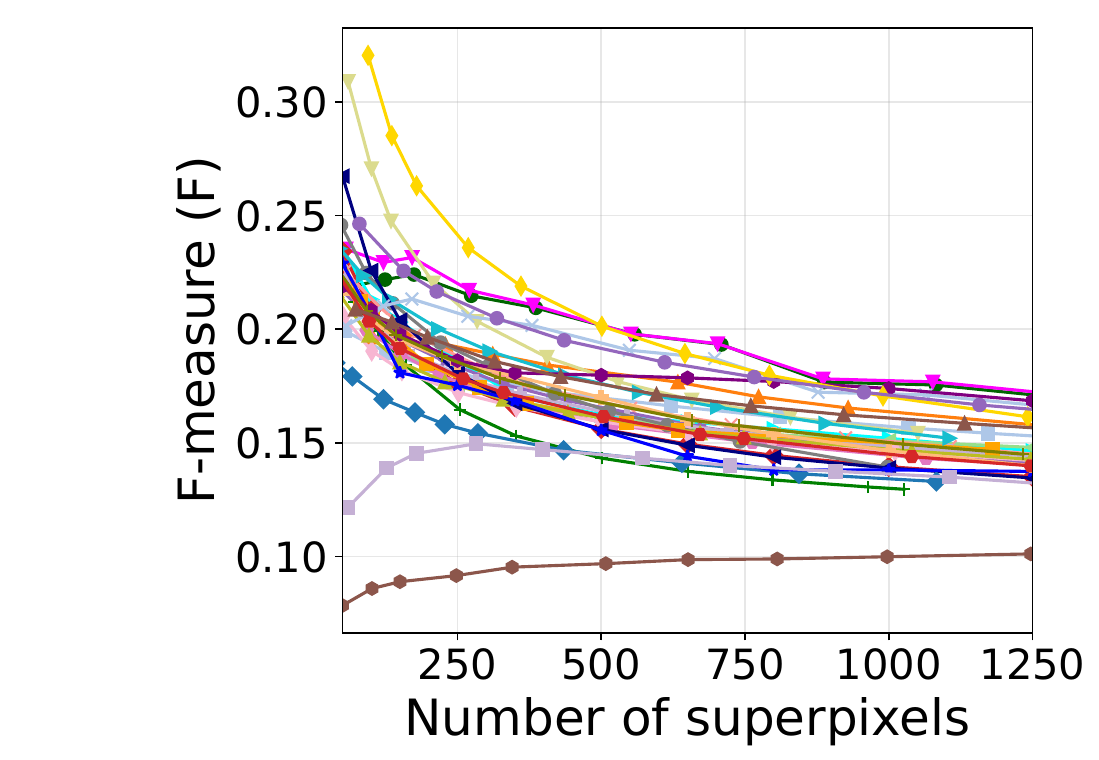}    &
			\includegraphics[width=0.19\textwidth]{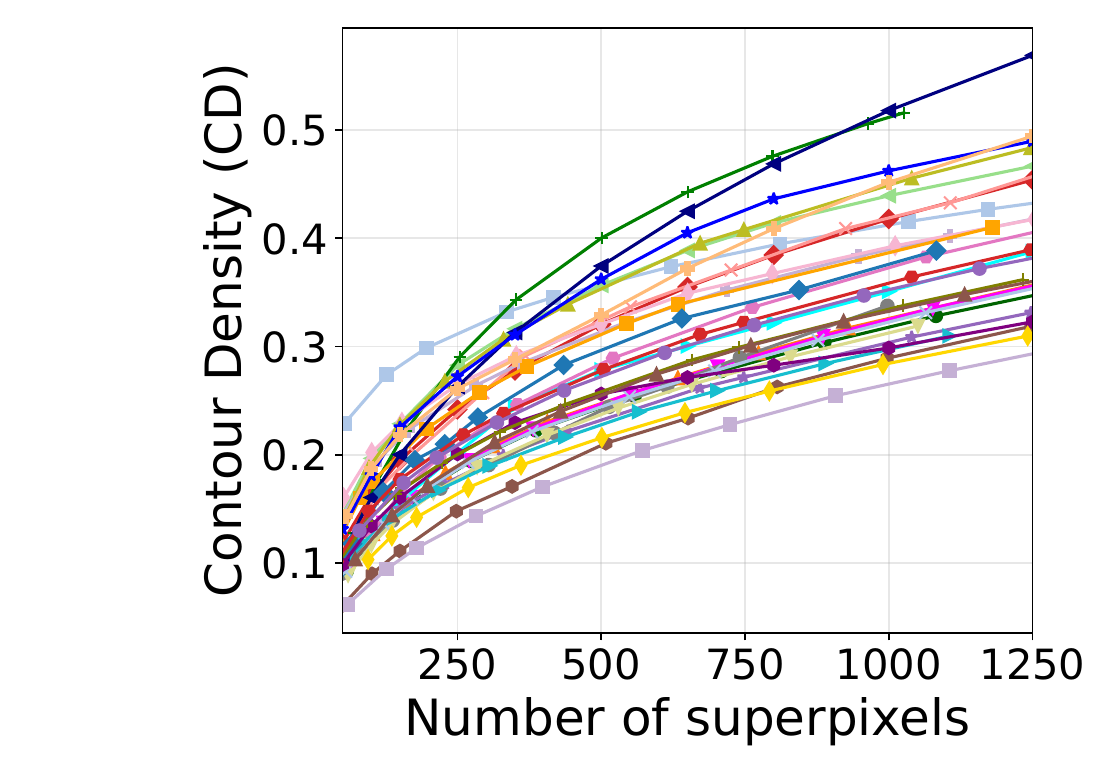}   &
			\includegraphics[width=0.19\textwidth]{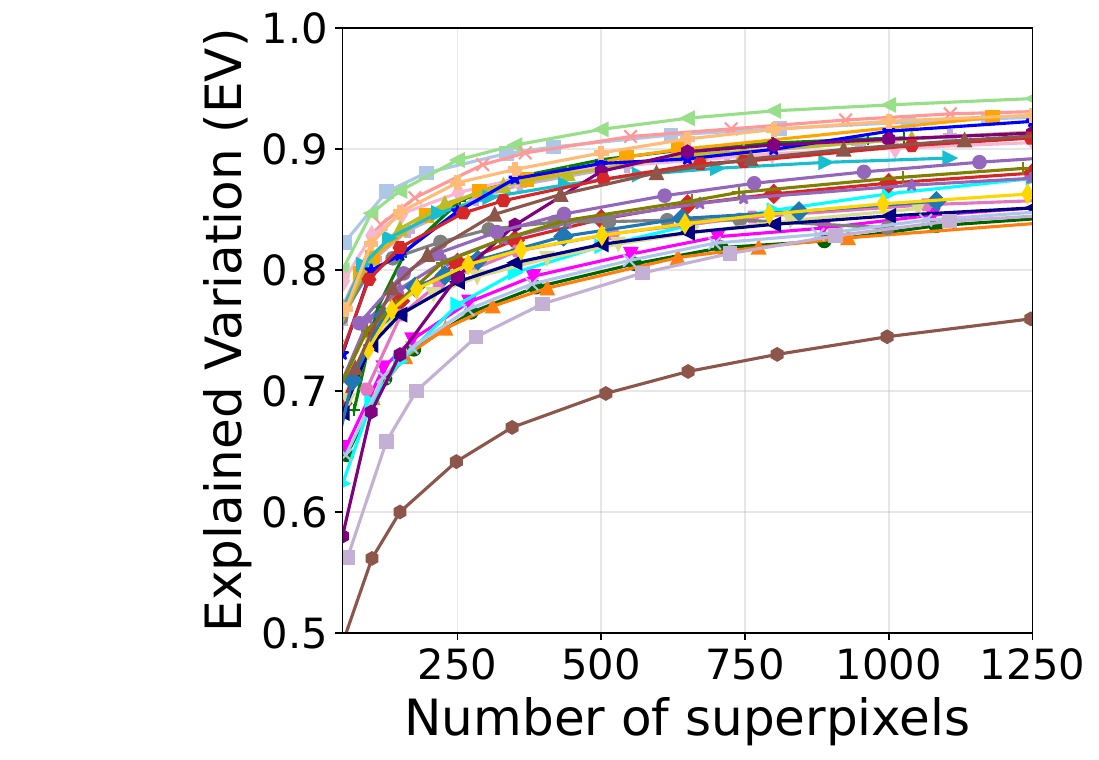}   &
			\includegraphics[width=0.19\textwidth]{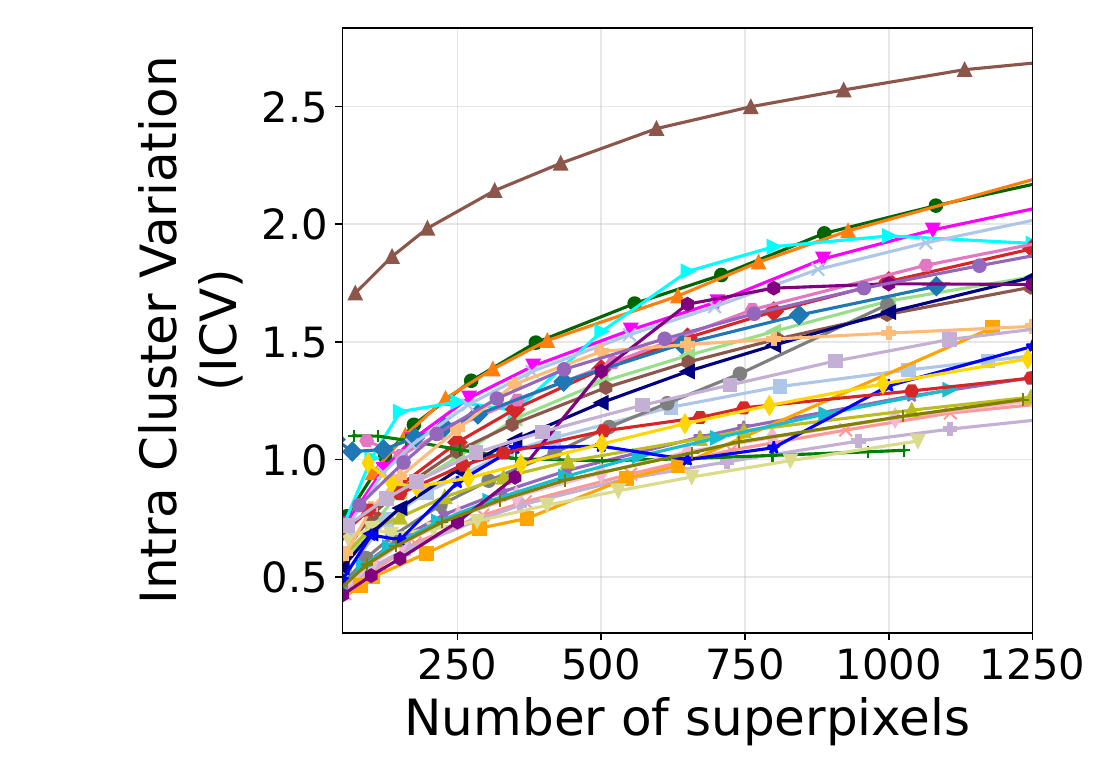}  &
			\includegraphics[width=0.19\textwidth]{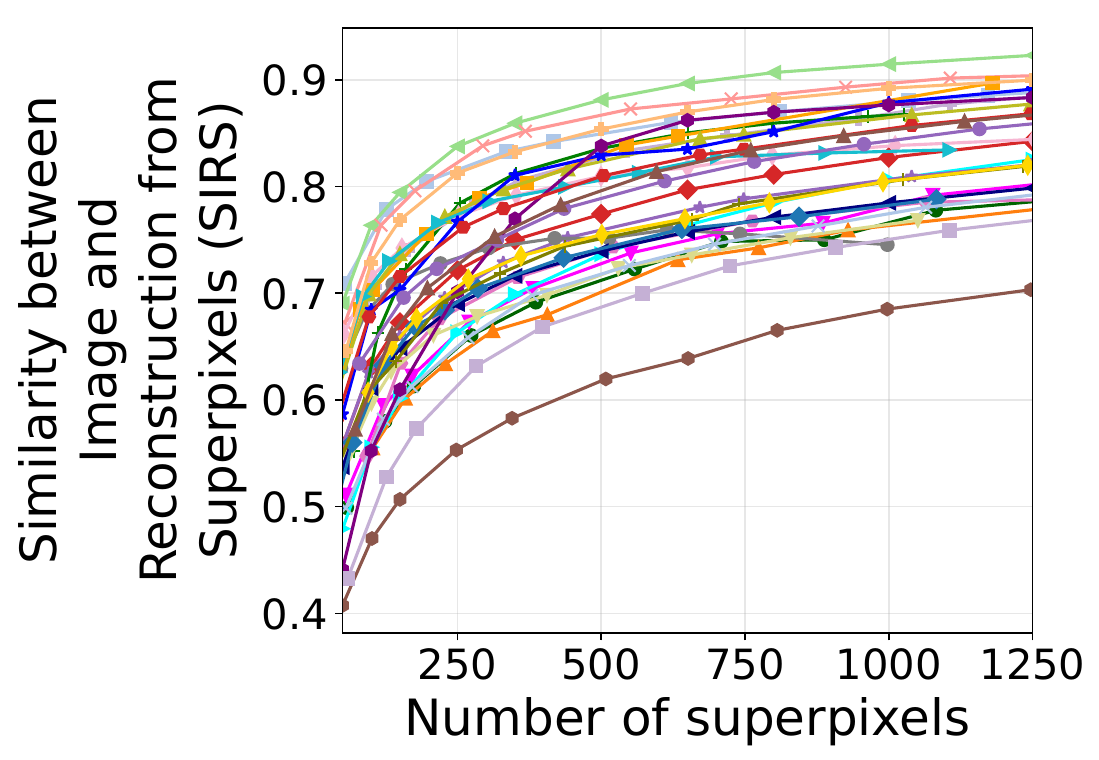}   \\
			\textcolor{metriccontour}{F}                                          &
			\textcolor{metriccontour}{CD}                                         &
			\textcolor{metriccolor}{EV}                                           &
			\textcolor{metriccolor}{ICV}                                          &
			\textcolor{metriccolor}{SIRS}                                           \\
\end{tabular}}
	\endgroup
	\caption{\textbf{All BSD metric visualization.}
		Each plot reports the evolution of one metric on BSD. }
	\label{fig:bsd_all_metric_profiles}
\end{figure*}

\begin{figure*}[t!]
	\centering
	{\footnotesize
		\begin{tabular}{@{}c@{\hspace{1mm}}c@{\hspace{1mm}}c@{}}
			\includegraphics[width=0.32\textwidth]{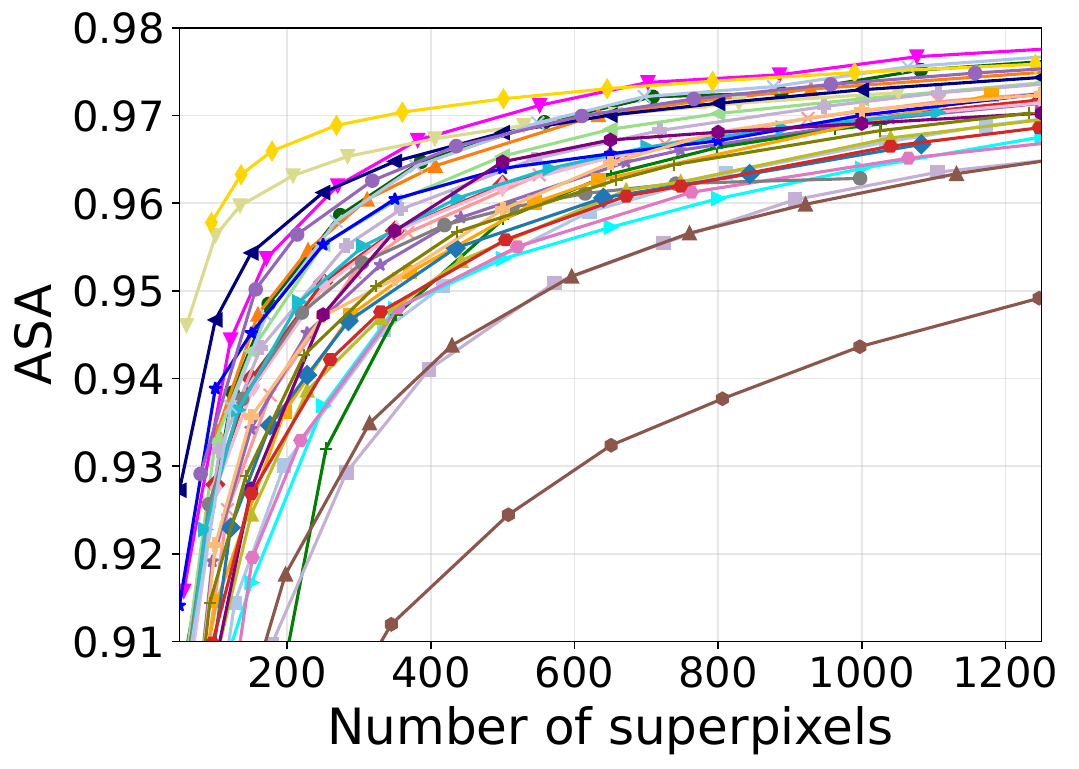}   &
			\includegraphics[width=0.32\textwidth]{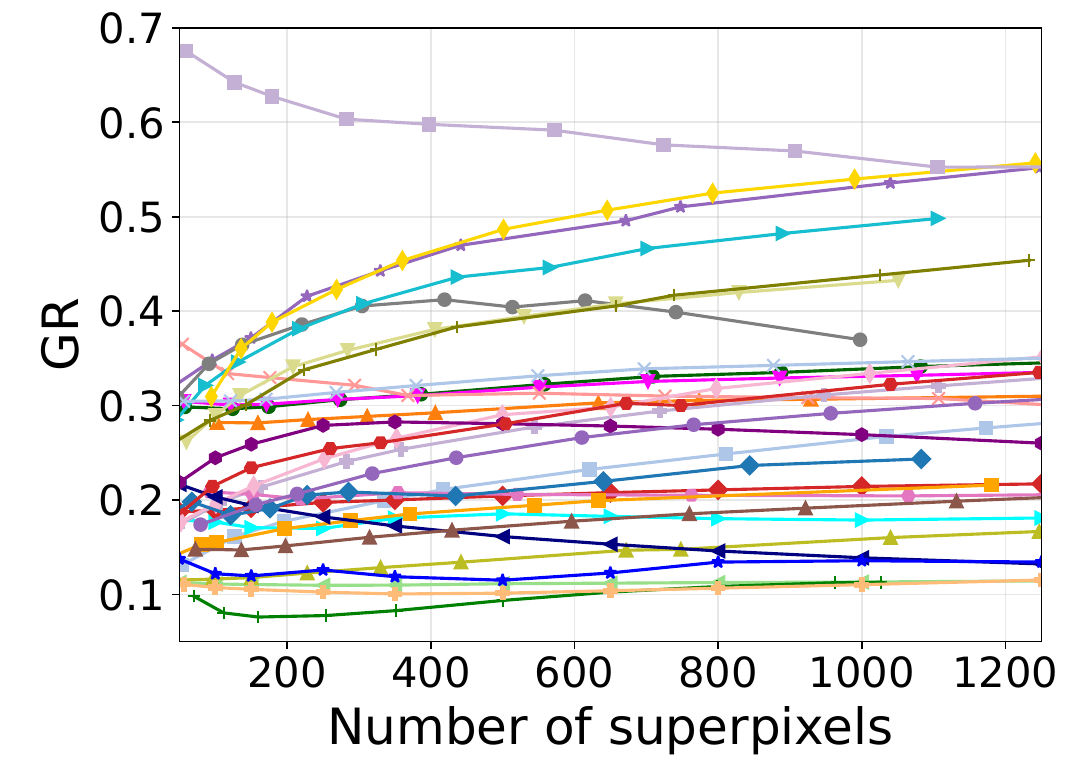}    &
			\includegraphics[width=0.32\textwidth]{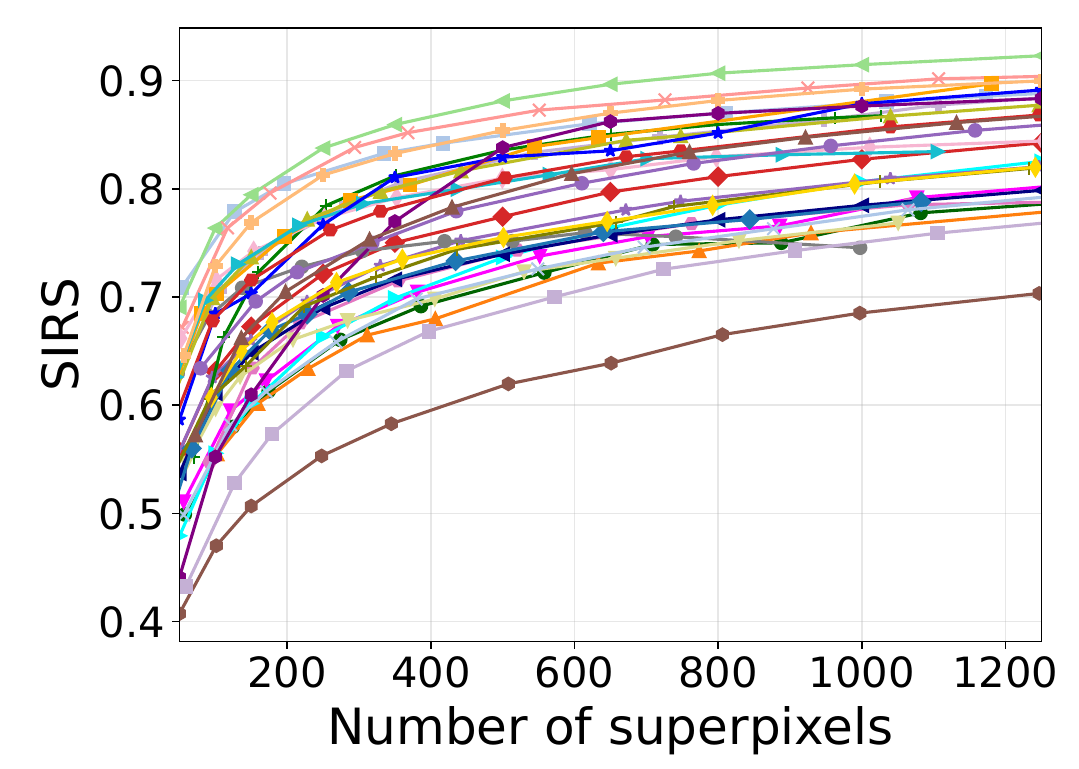} \\
			ASA                                                                     &
			GR                                                                      &
			SIRS                                                                                   \\
			\includegraphics[width=0.32\textwidth]{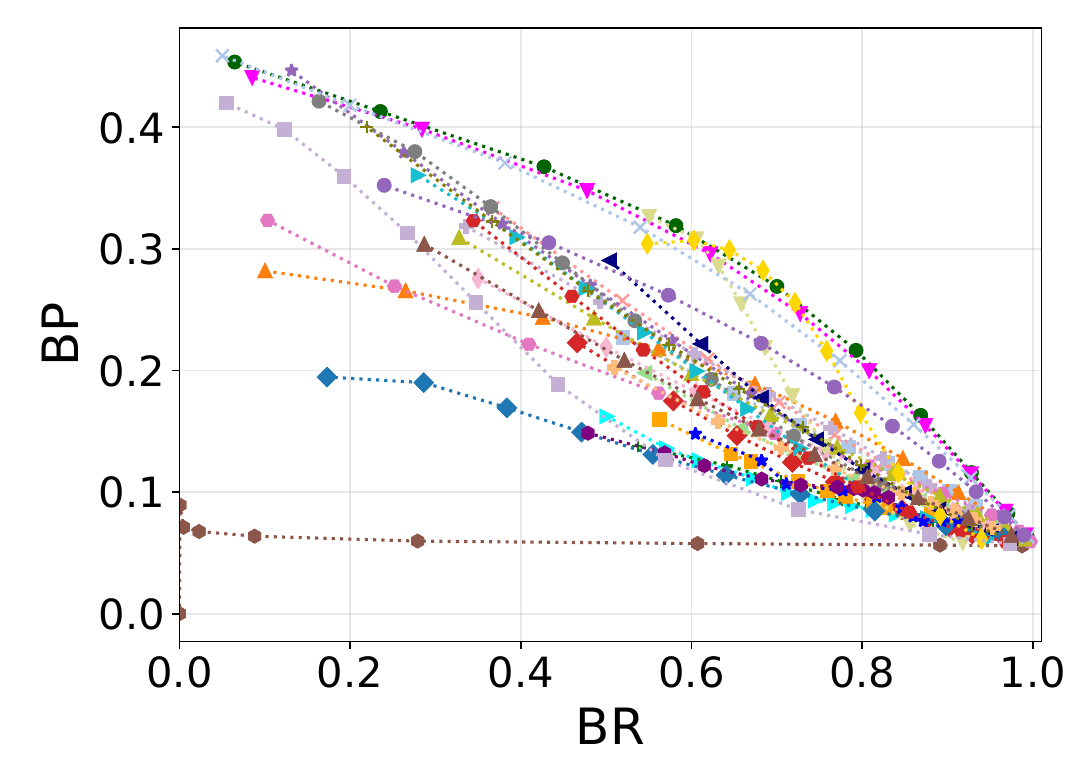} &
			\includegraphics[width=0.32\textwidth]{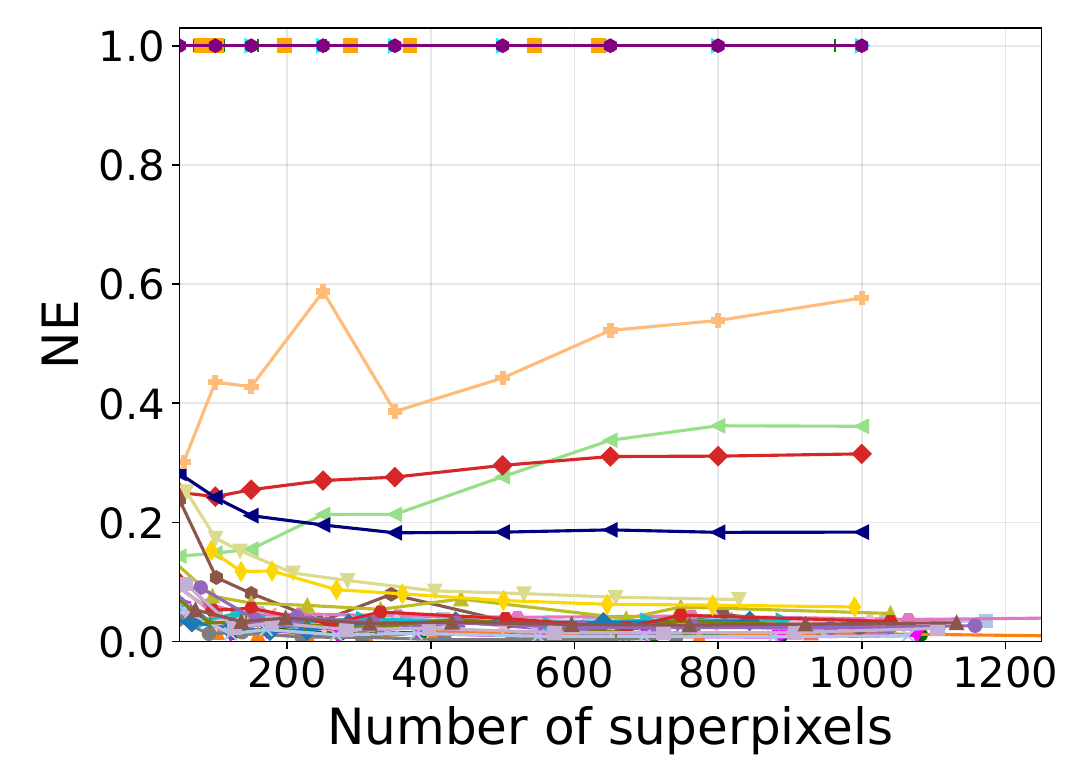}    &
			\includegraphics[width=0.32\textwidth]{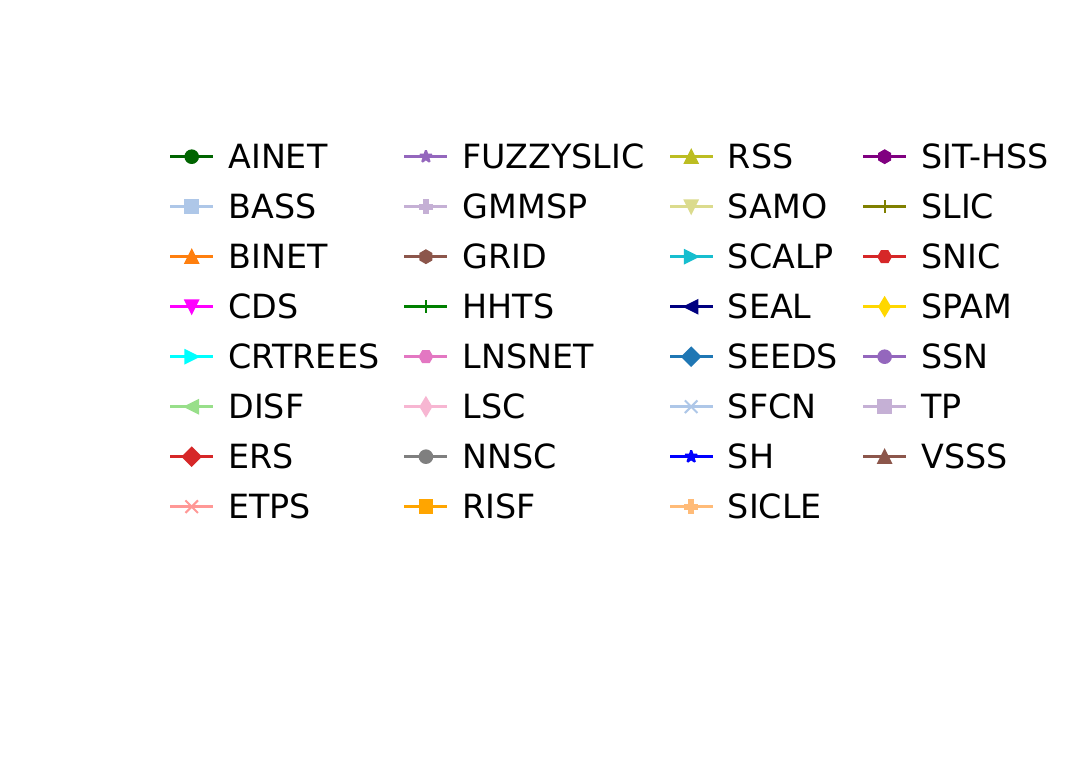}    \\
			BP / BR                                                                 &
			NE                                                                      &
			\\
	\end{tabular}}
	
	\caption{\textbf{BSD visualization profiles for the recommended  metric subset.}
		The plots summarize representative trends for accuracy (ASA), regularity
		(GR), color homogeneity (SIRS), contour adherence (BR/BP), and
		nestedness (NE).}
	\label{fig:bsd_selected_metric_profiles}
\end{figure*}

Fig.~\ref{fig:bsd_all_metric_profiles} reports the performance on the BSD dataset for all metrics in the SUPERVISE framework, while
Fig.~\ref{fig:bsd_selected_metric_profiles} selects only the metrics of the recommended subset as a standard validation example for newly introduced methods.
Results for each dataset can be directly displayed using our open source code.
Tabs.~\ref{tab:auc-values}
and~\ref{tab:auc-ranking} then provide the corresponding quantitative results:
the first reports AUC values for each metric, while the second converts these
values into ranks to make the ordering of methods easier to compare within and
across metric families. The final rank is obtained with the mean rank on the metric family.

\begin{figure*}[p]
	\centering
	{\scriptsize
		\setlength{\tabcolsep}{1pt}
		\renewcommand{\arraystretch}{0.85}
		\begin{tabular}{@{}cccc@{}}
			\includegraphics[width=0.245\textwidth]{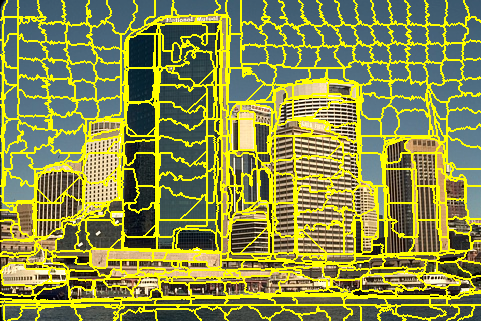}     &
			\includegraphics[width=0.245\textwidth]{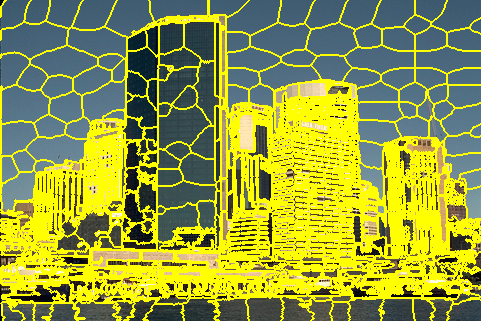}      &
			\includegraphics[width=0.245\textwidth]{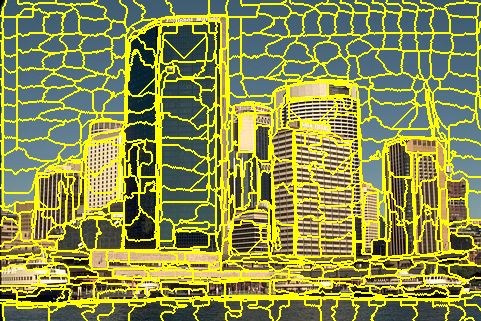}     &
			\includegraphics[width=0.245\textwidth]{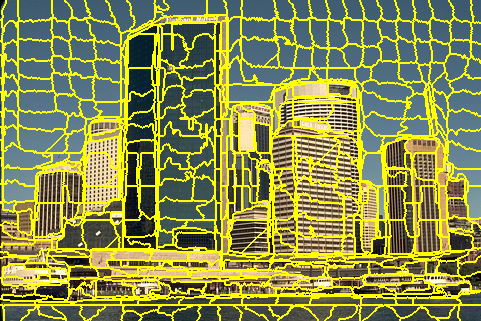}                                    \\
			AINET \cite{wang2021ainet}                                                                                         & BASS \cite{uziel2019bass}   & BINET \cite{zhao2024rethinking}     & CDS \cite{xu2024cds}   \\[0.4ex]
			\includegraphics[width=0.245\textwidth]{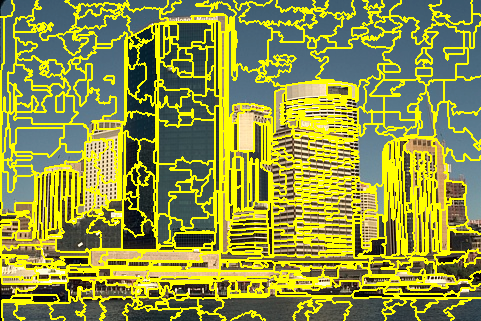}   &
			\includegraphics[width=0.245\textwidth]{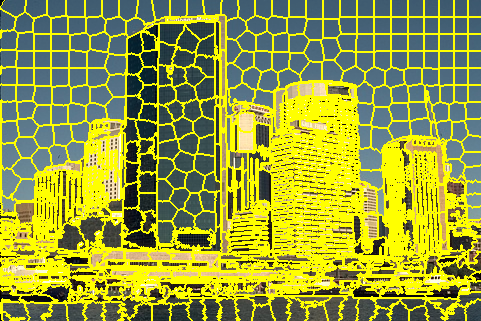}      &
			\includegraphics[width=0.245\textwidth]{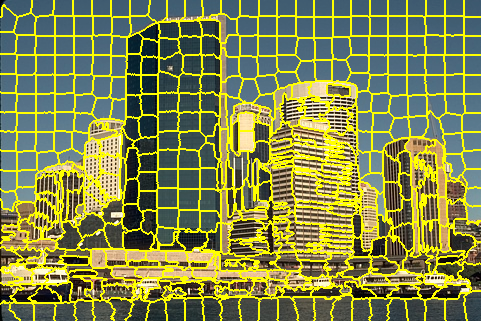} &
			\includegraphics[width=0.245\textwidth]{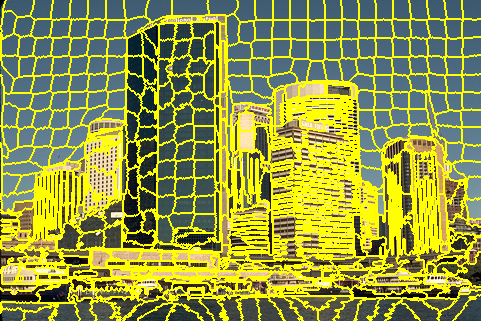}                                  \\
			CRTREES \cite{yan2022hierarchical}                                                                                       & ETPS \cite{yao2015}   & FUZZYSLIC \cite{wu2020fuzzy} & GMMSP \cite{Ban18} \\[0.4ex]
			\includegraphics[width=0.245\textwidth]{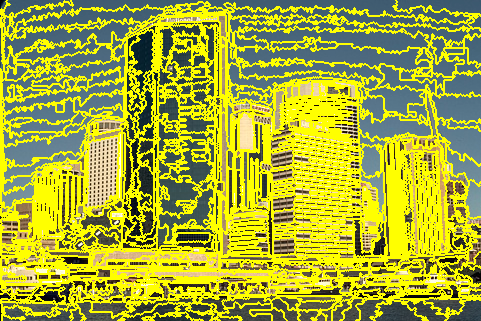}      &
			\includegraphics[width=0.245\textwidth]{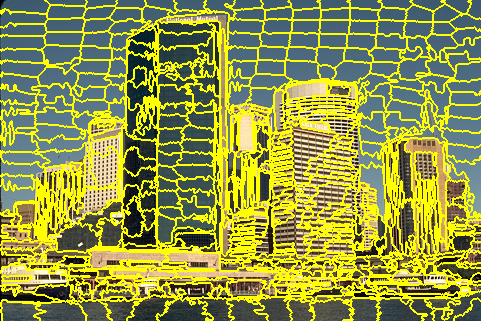}    &
			\includegraphics[width=0.245\textwidth]{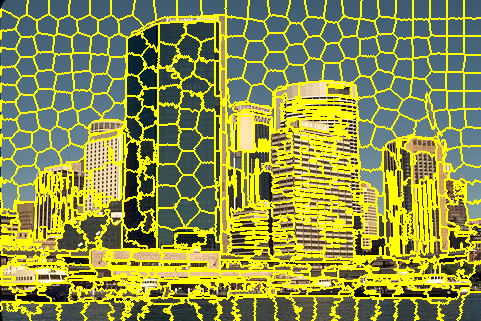}       &
			\includegraphics[width=0.245\textwidth]{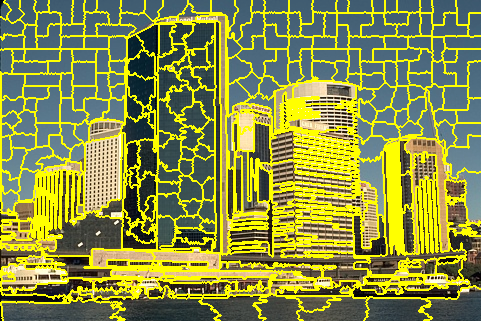}                                   \\
			HHTS \cite{chang2024hierarchical}                                                                                          & LNSNET \cite{zhu2021learning} & LSC \cite{chen2017}       & RISF \cite{galvao2020image}  \\
			\includegraphics[width=0.245\textwidth]{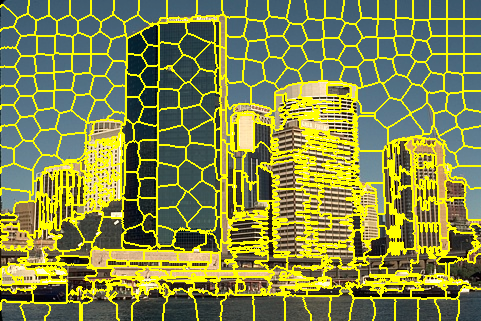}     &
			\includegraphics[width=0.245\textwidth]{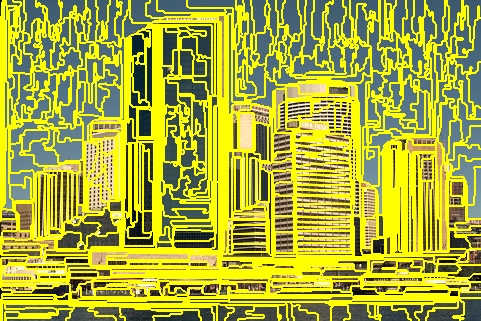}      &
			\includegraphics[width=0.245\textwidth]{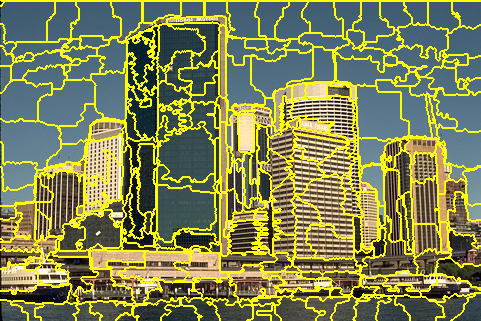}     &
			\includegraphics[width=0.245\textwidth]{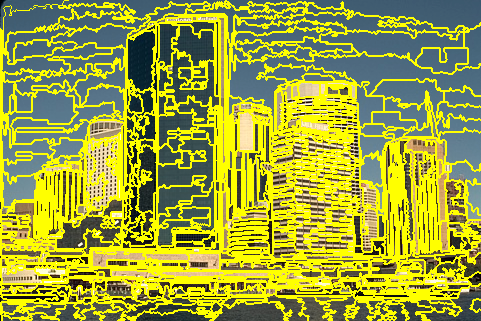}                                     \\
			SCALP \cite{giraud2018_scalp}                                                                                         & SEAL \cite{tu2018learning}   & SEEDS \cite{vandenbergh2012}     & SH \cite{wei2018sh}    \\[0.4ex]
			\includegraphics[width=0.245\textwidth]{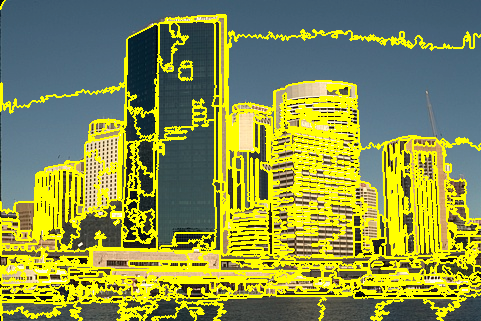}     &
			\includegraphics[width=0.245\textwidth]{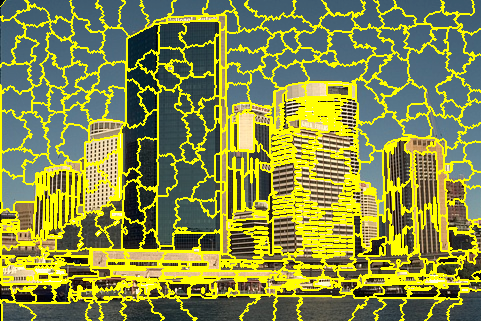}    &
			\includegraphics[width=0.245\textwidth]{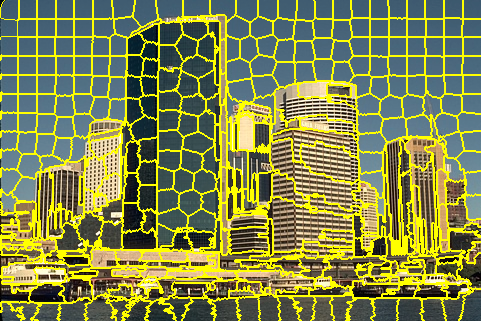}      &
			\includegraphics[width=0.245\textwidth]{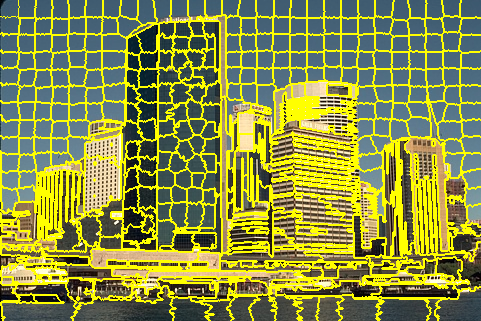}                                   \\
			SICLE \cite{belem2022fast}                                                                                         & SIT-HSS \cite{xie2025hierarchical} & SLIC \cite{achanta2012}      & SNIC \cite{achanta2017superpixels}  \\[0.4ex]
			\includegraphics[width=0.245\textwidth]{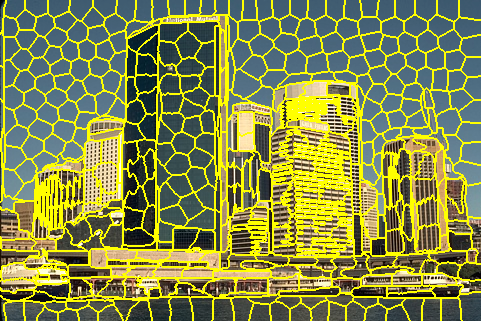} &
			\includegraphics[width=0.245\textwidth]{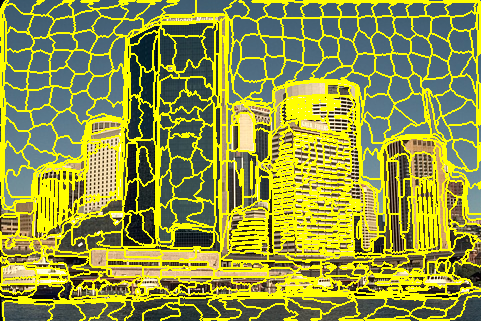}       &
			\includegraphics[width=0.245\textwidth]{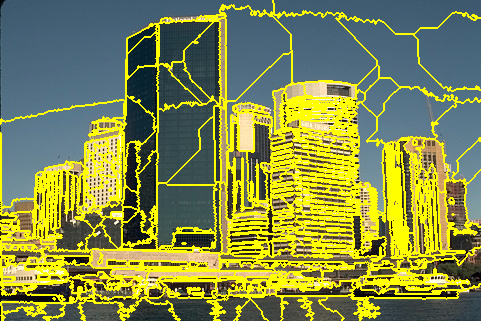}      &                            \\
			SPAM \cite{walther2025spam}                                                                                          & SSN \cite{jampani2018}    & VSSS \cite{zhou2023vine}      & {}
		\end{tabular}
	}
	\caption{\textbf{Qualitative superpixel segmentation results} on a BSD image for $k=500$ requested superpixels.}
	\label{fig:qualitative_bsd}
\end{figure*}

We report qualitative examples in Fig.~\ref{fig:qualitative_bsd}. These visual results illustrate the diversity of behaviors across methods: some produce noisy or fragmented decompositions, others favor highly regular superpixels, while some better adhere to object boundaries.

\subsection{Ranking stability and cross-dataset trends}
\label{sec:ranking_stability}

To assess how strongly method rankings depend on the evaluated superpixel-count
range, Tab.~\ref{tab:auc-interval-family-rank-summary-by-method} summarizes the
BSD rankings over four intervals: 50--500, 100--800, 200--900, and 400--1200
superpixels. These intervals cover both low and high superpixel budgets and are
intended to reveal stable trends as well as methods whose ranking changes with
scale. For each metric family, we report the mean rank, standard deviation, best
rank, and worst rank across the four intervals.
Complementing the recommended subset of Tab.~\ref{tab:auc-category-ranks-recommended},
Tab.~\ref{tab:auc-category-ranks-all} reports family-level rankings on all datasets,
where the per-metric AUC ranks of each family are averaged with equal weights.

\newpage

\subsection{Cross-dataset correlation}
\label{sec:cross_dataset_compact_panel}

\begin{wrapfigure}{r}{0.31\textwidth}
	\centering
	\vspace{-0.6cm}
	\definecolor{metricaccuracy}{RGB}{105,56,135}
	\definecolor{metriccontour}{RGB}{64,130,205}
	\definecolor{metricregularity}{RGB}{210,70,70}
	\definecolor{metriccolor}{RGB}{220,145,0}
	\definecolor{metrictopology}{RGB}{0,155,145}
	\setlength{\tabcolsep}{2pt}
	\includegraphics[width=0.265\textwidth]{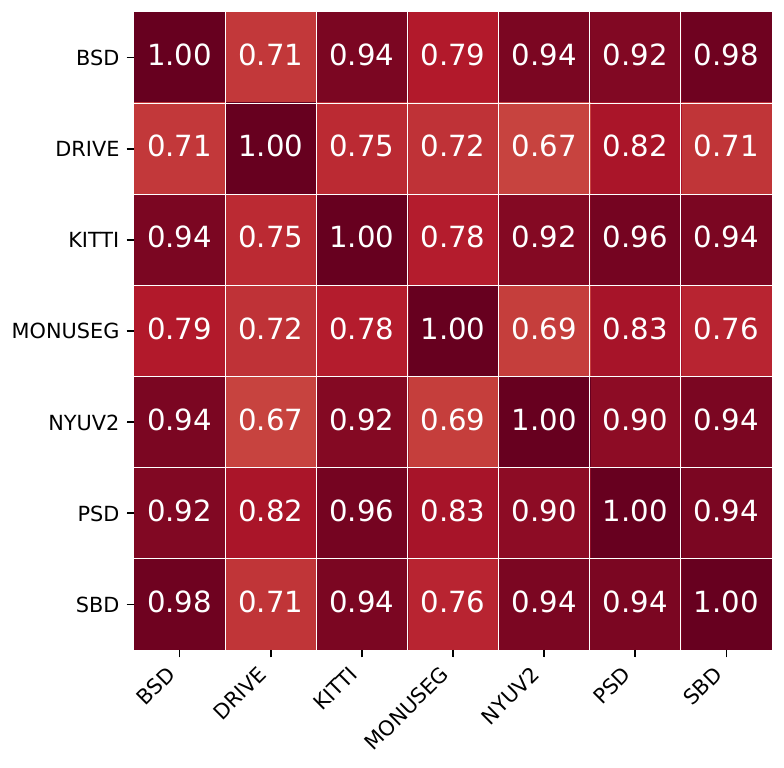}
	\caption{\textbf{Dataset correlation analysis.}
		Spearman correlations indicate higher similarity between natural image datasets.}
	\label{fig:stats_autres}
\end{wrapfigure}

Finally, we propose the first analysis of performance across datasets by comparing %
the vectors of method-level metric scores obtained after scale aggregation.
The cross-dataset correlation matrix in Fig.~\ref{fig:stats_autres} reveals high correlations ($0.67$--$0.98$), especially between BSD, SBD, NYUv2, KITTI, and PSD, indicating that similar metric structures organize the evaluation space across these datasets.

However, the datasets are not fully interchangeable.
DRIVE and MONUSEG exhibit lower correlations with other datasets, likely due to their specific image content and annotation topology (thin retinal vessels and dense nuclei instances, respectively).
If a compact dataset subset were to be selected, BSD appears particularly relevant due to its fine multi-annotations,
NYUv2 for its image diversity and lower correlation,
and DRIVE/MONUSEG as out-of-domain datasets stressing complementary segmentation properties.

\begin{figure*}[t!]
	\centering
	\definecolor{metricaccuracy}{RGB}{105,56,135}
	\definecolor{metriccontour}{RGB}{64,130,205}
	\definecolor{metricregularity}{RGB}{210,70,70}
	\definecolor{metriccolor}{RGB}{220,145,0}
	\definecolor{metrictopology}{RGB}{0,155,145}
	\setlength{\tabcolsep}{2pt}
	\renewcommand{\arraystretch}{0.8}

	\begin{tabular}{cccc}
		\includegraphics[width=0.24\textwidth]{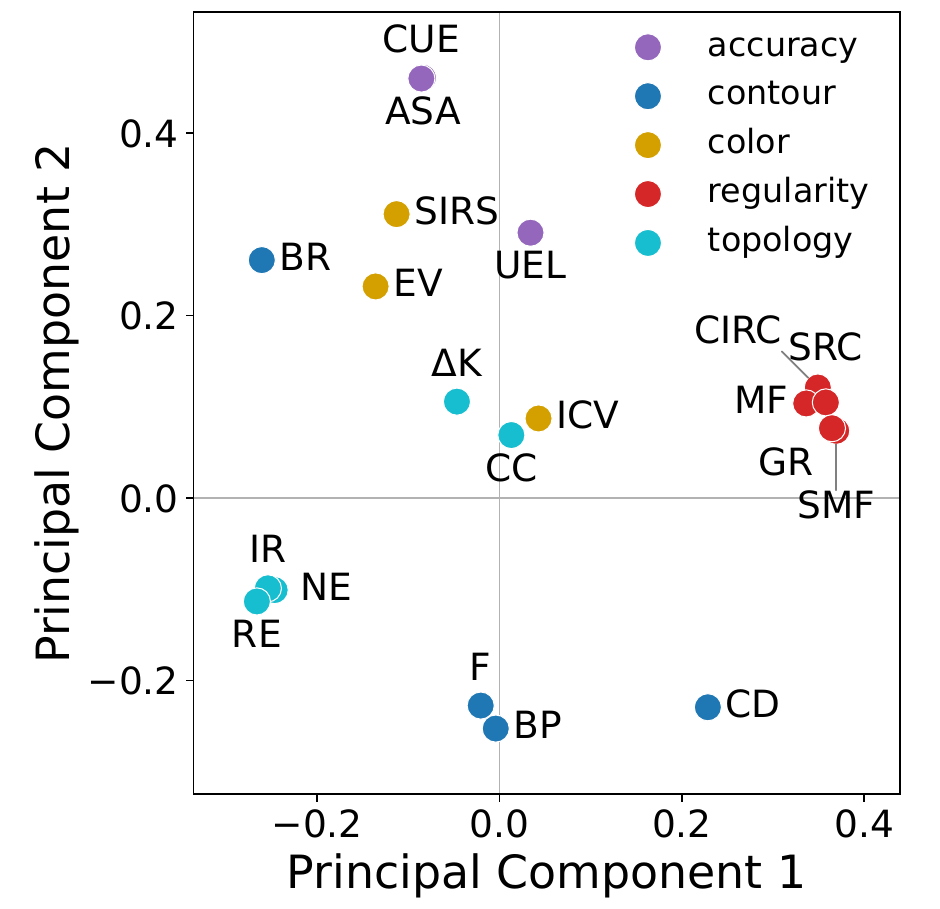}     &
		\includegraphics[width=0.24\textwidth]{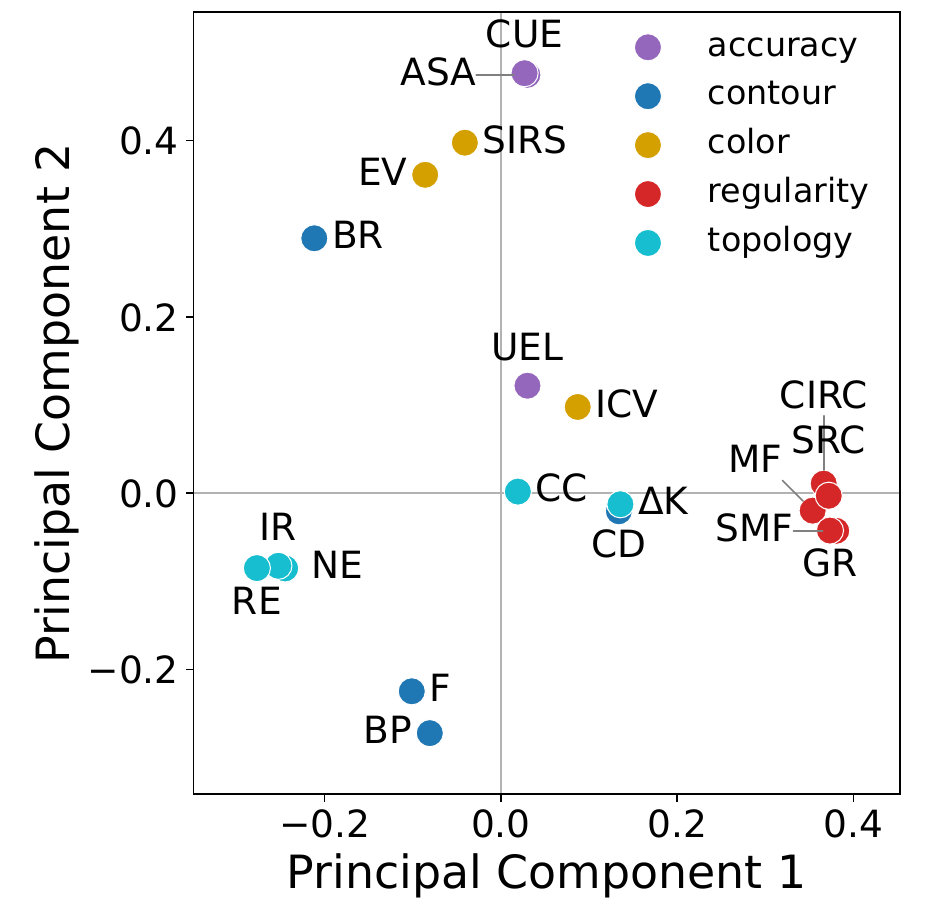}   &
		\includegraphics[width=0.24\textwidth]{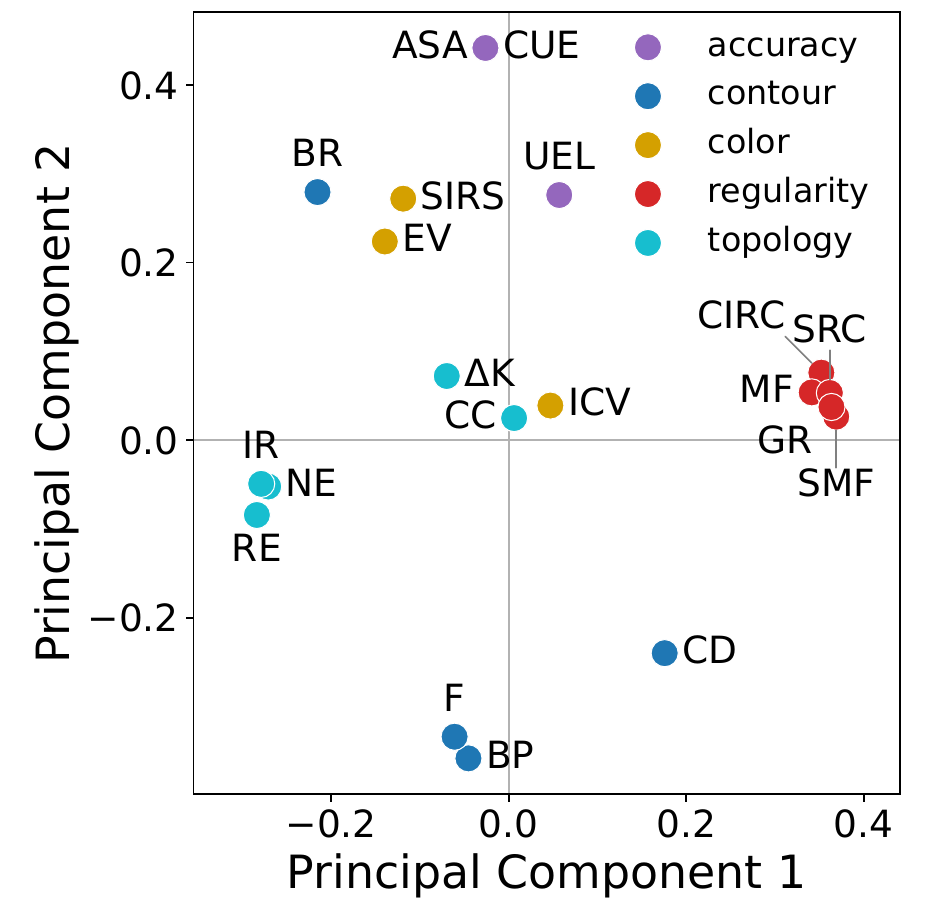}     &
		\includegraphics[width=0.24\textwidth]{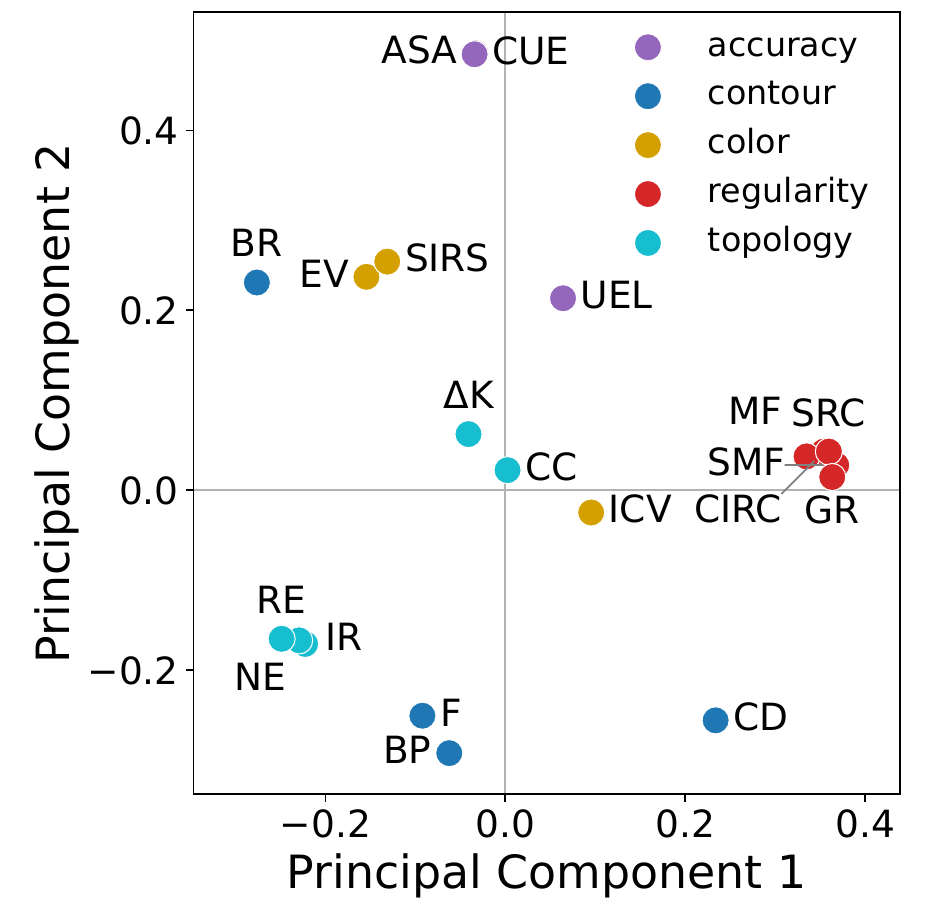}     \\
		\scriptsize {BSD}                                                &
		\scriptsize {NYUv2}                                              &
		\scriptsize {SBD}                                                &
		\scriptsize {KITTI}                                                \\
		\includegraphics[width=0.24\textwidth]{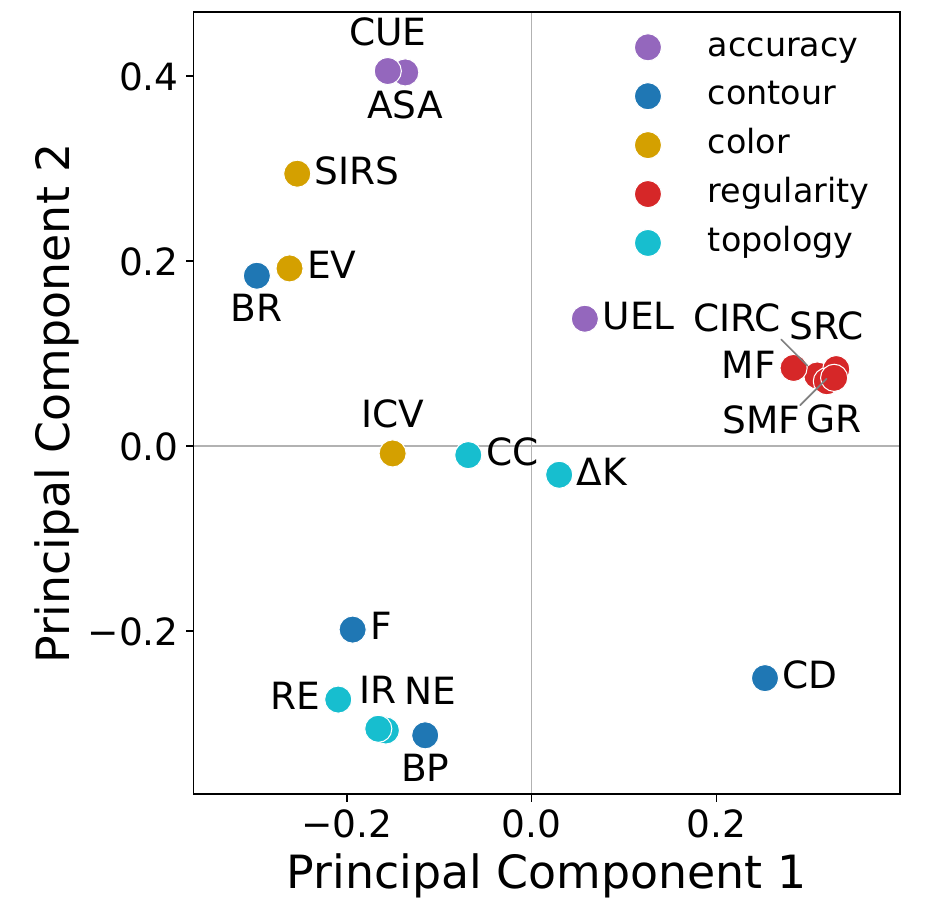} &
		\includegraphics[width=0.24\textwidth]{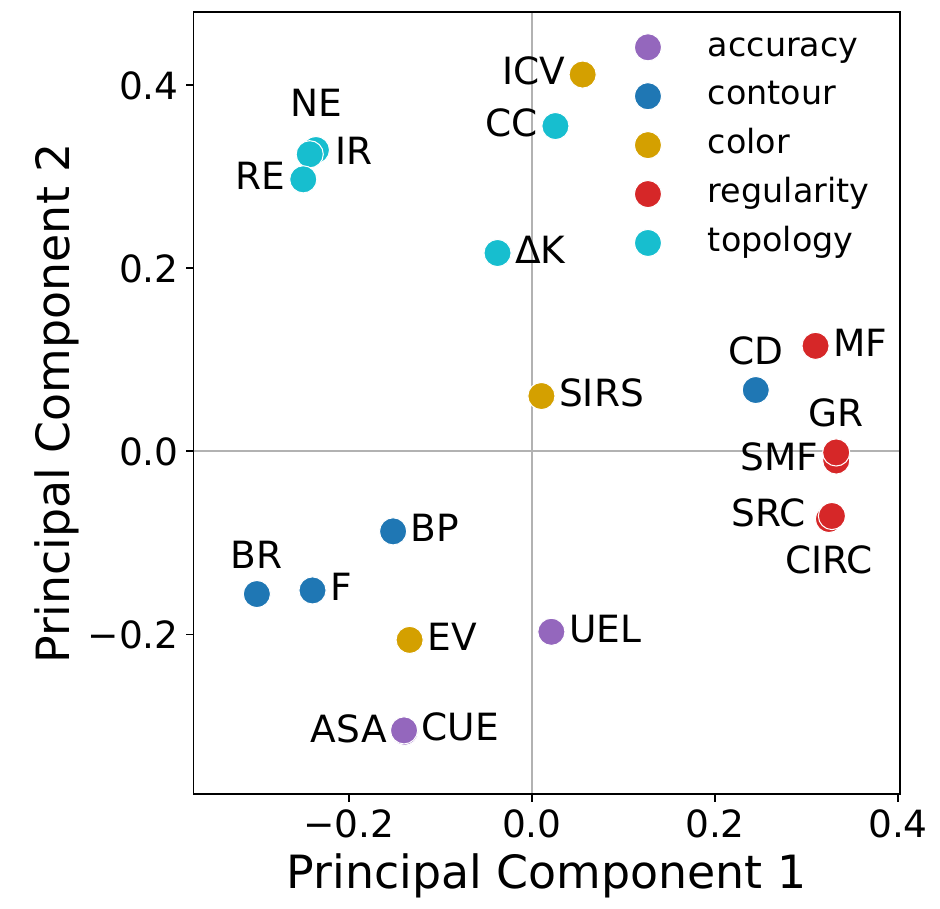}   &
		\includegraphics[width=0.24\textwidth]{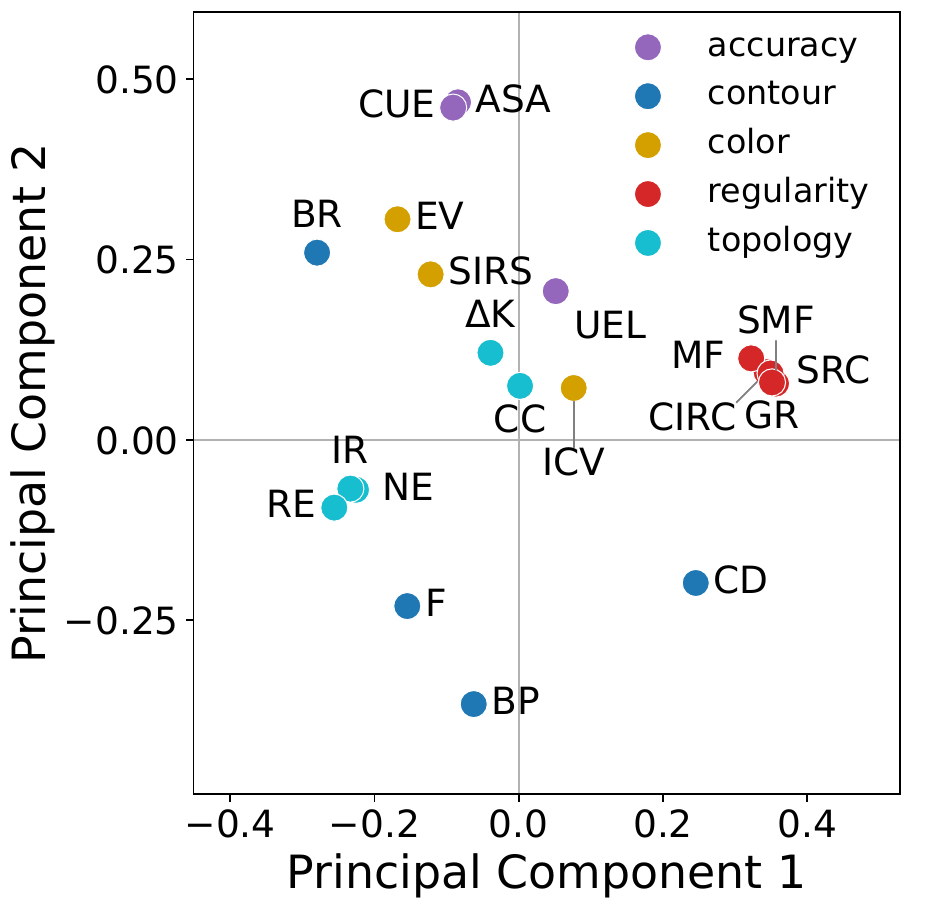}     & \\
		\scriptsize {MONUSEG}                                            &
		\scriptsize {DRIVE}                                              &
		\scriptsize {PSD}                                                & \\
	\end{tabular}

	\vspace{1.0ex}

	\caption{\textbf{PCA projection of the metric space across datasets.}
		The first two principal components show a consistent structure across datasets, with regularity metrics forming a distinct group and topology, contour, and accuracy metrics occupying complementary regions.}
	\label{fig:pca_projection}
\end{figure*}

\subsection{PCA projection cross-dataset metric space analysis}
\label{sec:appendix_pca}

Fig.~\ref{fig:pca_projection} shows that the projections of the two principal components have very similar trends across datasets. In particular,
regularity metrics remain separated from the other families in most cases,
indicating that they capture a geometric behavior that is not reducible to
accuracy, contour adherence, or color homogeneity. Accuracy and contour metrics
are usually closer to each other, which is consistent with the fact that
recovering object regions and recovering object boundaries are related
objectives. Color and topology metrics occupy more dataset-dependent positions,
but they do not collapse onto a single common direction either.

\begin{table*}[t]
	\centering
	\scriptsize
	\setlength{\tabcolsep}{2.3pt}
	\caption{\textbf{Ranking stability across interpolation ranges on the BSD dataset}. Rankings are computed independently over four intervals $[k_{\min},k_{\max}] \in \{[50,500], [100,800], [200,900], [400,1200]\}$ to evaluate how sensitive each method is to the chosen range. For each metric family, the table reports the mean rank, standard deviation, best rank, and worst rank across intervals. Lower ranks are better.}
	\label{tab:auc-interval-family-rank-summary-by-method}
	\resizebox{\textwidth}{!}{%
		\begin{tabular}{l>{\columncolor{aucaccuracy}}r>{\columncolor{aucaccuracy}}r>{\columncolor{aucaccuracy}}r>{\columncolor{aucaccuracy}}r>{\columncolor{auccontour}}r>{\columncolor{auccontour}}r>{\columncolor{auccontour}}r>{\columncolor{auccontour}}r>{\columncolor{auccolor}}r>{\columncolor{auccolor}}r>{\columncolor{auccolor}}r>{\columncolor{auccolor}}r>{\columncolor{aucregularity}}r>{\columncolor{aucregularity}}r>{\columncolor{aucregularity}}r>{\columncolor{aucregularity}}r>{\columncolor{auctopology}}r>{\columncolor{auctopology}}r>{\columncolor{auctopology}}r>{\columncolor{auctopology}}r}
			\toprule
			                                     & \multicolumn{4}{c}{\textcolor{metricaccuracy}{Accuracy}} & \multicolumn{4}{c}{\textcolor{metriccontour}{Contour}} & \multicolumn{4}{c}{\textcolor{metriccolor}{Color}} & \multicolumn{4}{c}{\textcolor{metricregularity}{Regularity}} & \multicolumn{4}{c}{\textcolor{metrictopology}{Topology}}                                                                                                                                                                                                                                                                                                                                                                                                                                                                                             \\
			\cmidrule(lr){2-5}\cmidrule(lr){6-9}\cmidrule(lr){10-13}\cmidrule(lr){14-17}\cmidrule(lr){18-21}
			\textbf{Method}                      & \textbf{Mean}                                            & \textbf{Std}                                           & \textbf{Best}                                      & \textbf{Worst}                                               & \textbf{Mean}                                            & \textbf{Std} & \textbf{Best}                      & \textbf{Worst}                     & \textbf{Mean}                         & \textbf{Std} & \textbf{Best}                      & \textbf{Worst}                     & \textbf{Mean}                         & \textbf{Std} & \textbf{Best}                      & \textbf{Worst}                     & \textbf{Mean}                         & \textbf{Std} & \textbf{Best}                      & \textbf{Worst}                     \\
			\midrule
			AINET \cite{wang2021ainet} & 6.25 & 0.83 & 5 & 7 & \textcolor{thirdgreen}{\textbf{3.00}} & 0.71 & \textcolor{secondblue}{\textbf{2}} & 4 & 30.25 & 0.43 & 30 & 31 & 11.75 & 0.83 & 11 & 13 & 28.25 & 3.03 & 23 & 30 \\
			BASS \cite{uziel2019bass} & 21.75 & 2.59 & 19 & 26 & 14.75 & 3.90 & 11 & 21 & 4.00 & 1.00 & \textcolor{thirdgreen}{\textbf{3}} & 5 & 19.75 & 1.92 & 18 & 23 & 22.25 & 2.77 & 20 & 27 \\
			BINET \cite{zhao2024rethinking} & 4.75 & 1.09 & \textcolor{thirdgreen}{\textbf{3}} & 6 & 8.25 & 1.09 & 7 & 10 & 30.75 & 0.43 & 30 & 31 & 15.25 & 0.83 & 14 & 16 & 25.00 & 1.00 & 24 & 26 \\
			CDS \cite{xu2024cds} & \textcolor{bestred}{\textbf{1.25}} & 0.43 & \textcolor{bestred}{\textbf{1}} & \textcolor{secondblue}{\textbf{2}} & \textcolor{secondblue}{\textbf{2.00}} & 0.71 & \textcolor{bestred}{\textbf{1}} & \textcolor{thirdgreen}{\textbf{3}} & 26.75 & 0.43 & 26 & 27 & 11.50 & 0.50 & 11 & 12 & 21.75 & 0.43 & 21 & 22 \\
			CRTREES \cite{yan2022hierarchical} & 29.50 & 0.50 & 29 & 30 & 17.00 & 1.41 & 15 & 19 & 25.75 & 1.09 & 24 & 27 & 24.00 & 0.00 & 24 & 24 & 9.00 & 0.00 & 9 & 9 \\
			DISF \cite{belem2020superpixel} & 13.25 & 0.83 & 12 & 14 & 18.50 & 2.96 & 14 & 22 & 4.75 & 1.09 & \textcolor{thirdgreen}{\textbf{3}} & 6 & 29.25 & 0.43 & 29 & 30 & 5.50 & 0.50 & 5 & 6 \\
			ERS \cite{liu2011} & 10.25 & 0.43 & 10 & 11 & 23.75 & 1.09 & 22 & 25 & 20.00 & 0.00 & 20 & 20 & 21.25 & 0.83 & 20 & 22 & 5.75 & 0.83 & 5 & 7 \\
			ETPS \cite{yao2015} & 13.00 & 1.00 & 12 & 14 & 11.50 & 1.66 & 10 & 14 & \textcolor{bestred}{\textbf{1.00}} & 0.00 & \textcolor{bestred}{\textbf{1}} & \textcolor{bestred}{\textbf{1}} & 10.25 & 2.28 & 8 & 14 & 19.00 & 0.71 & 18 & 20 \\
			FUZZYSLIC \cite{wu2020fuzzy} & 13.00 & 0.71 & 12 & 14 & 16.00 & 1.22 & 14 & 17 & 14.50 & 0.87 & 14 & 16 & 3.75 & 0.43 & \textcolor{thirdgreen}{\textbf{3}} & 4 & 23.00 & 2.92 & 18 & 25 \\
			GMMSP \cite{Ban18} & 9.00 & 0.00 & 9 & 9 & 16.75 & 0.83 & 16 & 18 & \textcolor{thirdgreen}{\textbf{3.00}} & 0.00 & \textcolor{thirdgreen}{\textbf{3}} & \textcolor{thirdgreen}{\textbf{3}} & 15.75 & 0.83 & 15 & 17 & 18.50 & 1.66 & 17 & 21 \\
			GRID & 22.75 & 1.48 & 21 & 25 & 30.00 & 0.00 & 30 & 30 & 28.25 & 0.43 & 28 & 29 & \textcolor{bestred}{\textbf{1.00}} & 0.00 & \textcolor{bestred}{\textbf{1}} & \textcolor{bestred}{\textbf{1}} & 15.00 & 1.00 & 14 & 16 \\
			HHTS \cite{chang2024hierarchical} & 27.25 & 4.32 & 20 & 31 & 29.00 & 0.00 & 29 & 29 & 6.50 & 4.39 & \textcolor{secondblue}{\textbf{2}} & 13 & 30.25 & 0.83 & 29 & 31 & \textcolor{thirdgreen}{\textbf{3.00}} & 0.00 & \textcolor{thirdgreen}{\textbf{3}} & \textcolor{thirdgreen}{\textbf{3}} \\
			LNSNET \cite{zhu2021learning} & 28.25 & 0.83 & 27 & 29 & 25.25 & 0.43 & 25 & 26 & 24.00 & 0.71 & 23 & 25 & 21.50 & 2.06 & 18 & 23 & 27.75 & 0.43 & 27 & 28 \\
			LSC \cite{chen2017} & 15.50 & 0.87 & 15 & 17 & 25.00 & 2.12 & 22 & 27 & 6.75 & 3.03 & \textcolor{thirdgreen}{\textbf{3}} & 11 & 13.50 & 2.18 & 10 & 16 & 14.75 & 0.83 & 14 & 16 \\
			NNSC \cite{giraud2019texture} & 17.25 & 2.17 & 14 & 20 & 16.00 & 6.96 & 8 & 27 & 15.75 & 2.86 & 11 & 18 & 6.50 & 1.12 & 5 & 8 & 25.75 & 2.17 & 23 & 29 \\
			RISF \cite{galvao2020image} & 23.50 & 0.50 & 23 & 24 & 19.25 & 0.43 & 19 & 20 & \textcolor{secondblue}{\textbf{2.25}} & 0.43 & \textcolor{secondblue}{\textbf{2}} & \textcolor{thirdgreen}{\textbf{3}} & 22.25 & 1.64 & 21 & 25 & 5.25 & 1.09 & 4 & 7 \\
			RSS \cite{chai2020rooted} & 25.25 & 1.30 & 24 & 27 & 22.25 & 1.92 & 20 & 25 & 7.25 & 0.43 & 7 & 8 & 26.75 & 0.43 & 26 & 27 & 12.75 & 0.43 & 12 & 13 \\
			SAMO \cite{giraud2024} & 7.50 & 0.50 & 7 & 8 & 5.50 & 2.18 & \textcolor{secondblue}{\textbf{2}} & 8 & 16.75 & 1.92 & 14 & 19 & 6.50 & 0.50 & 6 & 7 & 11.25 & 0.43 & 11 & 12 \\
			SCALP \cite{giraud2018_scalp} & 10.75 & 0.43 & 10 & 11 & 8.00 & 1.00 & 7 & 9 & 9.50 & 1.50 & 7 & 11 & 5.00 & 0.00 & 5 & 5 & 14.75 & 0.43 & 14 & 15 \\
			SEAL \cite{tu2018learning} & \textcolor{thirdgreen}{\textbf{3.75}} & 0.83 & \textcolor{thirdgreen}{\textbf{3}} & 5 & 20.00 & 4.64 & 12 & 23 & 22.00 & 0.00 & 22 & 22 & 24.00 & 1.73 & 21 & 25 & 8.00 & 0.00 & 8 & 8 \\
			SEEDS \cite{vandenbergh2012} & 21.75 & 3.03 & 20 & 27 & 31.00 & 0.00 & 31 & 31 & 23.25 & 0.43 & 23 & 24 & 20.50 & 0.87 & 20 & 22 & 27.00 & 1.22 & 26 & 29 \\
			SFCN \cite{yang2020} & 4.00 & 0.71 & \textcolor{thirdgreen}{\textbf{3}} & 5 & 4.75 & 0.83 & 4 & 6 & 28.00 & 0.00 & 28 & 28 & 9.50 & 0.50 & 9 & 10 & 23.25 & 0.43 & 23 & 24 \\
			SH \cite{wei2018sh} & 15.50 & 1.66 & 13 & 17 & 26.25 & 1.48 & 24 & 28 & 8.50 & 1.12 & 7 & 10 & 28.00 & 0.00 & 28 & 28 & \textcolor{bestred}{\textbf{1.00}} & 0.00 & \textcolor{bestred}{\textbf{1}} & \textcolor{bestred}{\textbf{1}} \\
			SICLE \cite{belem2022fast} & 21.25 & 0.83 & 20 & 22 & 12.50 & 0.87 & 11 & 13 & 9.50 & 1.12 & 8 & 11 & 30.25 & 0.43 & 30 & 31 & 4.75 & 1.30 & 4 & 7 \\
			SIT-HSS \cite{xie2025hierarchical} & 17.25 & 1.30 & 16 & 19 & 8.00 & 1.58 & 6 & 10 & 15.00 & 1.41 & 13 & 17 & 15.00 & 3.16 & 11 & 19 & \textcolor{bestred}{\textbf{1.00}} & 0.00 & \textcolor{bestred}{\textbf{1}} & \textcolor{bestred}{\textbf{1}} \\
			SLIC \cite{achanta2012} & 18.50 & 0.50 & 18 & 19 & 15.00 & 0.71 & 14 & 16 & 14.00 & 1.00 & 13 & 15 & 7.75 & 0.43 & 7 & 8 & 17.75 & 1.48 & 16 & 20 \\
			SNIC \cite{achanta2017superpixels} & 24.75 & 1.30 & 23 & 26 & 21.00 & 0.71 & 20 & 22 & 12.00 & 0.00 & 12 & 12 & 14.00 & 1.00 & 13 & 15 & 11.50 & 0.87 & 10 & 12 \\
			SPAM \cite{walther2025spam} & \textcolor{secondblue}{\textbf{1.75}} & 0.43 & \textcolor{bestred}{\textbf{1}} & \textcolor{secondblue}{\textbf{2}} & \textcolor{bestred}{\textbf{1.25}} & 0.43 & \textcolor{bestred}{\textbf{1}} & \textcolor{secondblue}{\textbf{2}} & 17.25 & 1.09 & 16 & 19 & \textcolor{thirdgreen}{\textbf{3.25}} & 0.43 & \textcolor{thirdgreen}{\textbf{3}} & 4 & 10.00 & 0.00 & 10 & 10 \\
			SSN \cite{jampani2018} & 5.75 & 1.30 & 4 & 7 & 4.25 & 0.43 & 4 & 5 & 18.25 & 0.83 & 17 & 19 & 17.75 & 0.43 & 17 & 18 & 19.25 & 1.30 & 18 & 21 \\
			TP \cite{levinshtein2009turbopixels} & 26.25 & 1.92 & 23 & 28 & 26.75 & 1.09 & 25 & 28 & 25.25 & 0.43 & 25 & 26 & \textcolor{secondblue}{\textbf{2.00}} & 0.00 & \textcolor{secondblue}{\textbf{2}} & \textcolor{secondblue}{\textbf{2}} & 31.00 & 0.00 & 31 & 31 \\
			VSSS \cite{zhou2023vine} & 30.75 & 0.43 & 30 & 31 & 9.75 & 0.83 & 9 & 11 & 21.00 & 0.00 & 21 & 21 & 25.75 & 0.43 & 25 & 26 & 28.25 & 1.30 & 27 & 30 \\
			\bottomrule
		\end{tabular}%
	}
\end{table*}

\begin{table*}[t]
    \centering
    \tiny
    \setlength{\tabcolsep}{1.6pt}
    \renewcommand{\arraystretch}{0.9}
    \caption{\textbf{Method rankings by AUC for metric families across datasets}. For each family and dataset, the per-metric AUC ranks are averaged with equal weights, then these means are ranked. Rank 1 is best. Columns follow NYUv2 (N), BSD (B), SBD (S), KITTI (K), DRIVE (D), MONUSEG (M), and PSD (P).}
    \label{tab:auc-category-ranks-all}
    \resizebox{\textwidth}{!}{%
        \begin{tabular}{l*{7}{>{\columncolor{aucaccuracy}}r}*{7}{>{\columncolor{auccontour}}r}*{7}{>{\columncolor{auccolor}}r}*{7}{>{\columncolor{aucregularity}}r}*{7}{>{\columncolor{auctopology}}r}}
            \toprule
             & \multicolumn{7}{c}{\textcolor{metricaccuracy}{Accuracy}} & \multicolumn{7}{c}{\textcolor{metriccontour}{Contour}} & \multicolumn{7}{c}{\textcolor{metriccolor}{Color}} & \multicolumn{7}{c}{\textcolor{metricregularity}{Regularity}} & \multicolumn{7}{c}{\textcolor{metrictopology}{Topology}} \\
            \cmidrule(lr){2-8} \cmidrule(lr){9-15} \cmidrule(lr){16-22} \cmidrule(lr){23-29} \cmidrule(lr){30-36}
            Method & N & B & S & K & D & M & P & N & B & S & K & D & M & P & N & B & S & K & D & M & P & N & B & S & K & D & M & P & N & B & S & K & D & M & P \\
            \midrule
            AINET \cite{wang2021ainet} & 8 & 6 & 6 & 6 & 19 & 20 & 12 & 5 & 3 & 2 & 5 & 20 & 22 & 24 & 29 & 30 & 31 & 27 & 28 & 25 & 30 & 13 & 11 & 12 & 13 & 14 & 4 & 11 & 25 & 30 & 23 & 23 & 24 & 25 & 25 \\
            BASS \cite{uziel2019bass} & 19 & 22 & 19 & 24 & 19 & 21 & 28 & 8 & 12 & 9 & 7 & 11 & 20 & 2 & 3 & 3 & 3 & 2 & 6 & 1 & 5 & 17 & 19 & 18 & 20 & 12 & 28 & 18 & 21 & 21 & 26 & 16 & 23 & 23 & 22 \\
            BINET \cite{zhao2024rethinking} & 2 & 5 & 4 & 9 & 18 & 31 & 26 & 2 & 8 & 7 & 3 & 8 & 26 & 25 & 24 & 31 & 29 & 30 & 22 & 30 & 31 & 18 & 15 & 16 & 17 & 24 & 13 & 18 & 27 & 26 & 21 & 31 & 31 & 30 & 30 \\
            CDS \cite{xu2024cds} & 3 & 1 & 2 & 2 & 2 & 26 & 5 & 4 & 2 & 5 & 4 & 4 & 27 & 18 & 24 & 27 & 25 & 24 & 29 & 28 & 25 & 13 & 11 & 13 & 11 & 20 & 5 & 14 & 23 & 22 & 22 & 22 & 25 & 26 & 24 \\
            CRTREES \cite{yan2022hierarchical} & 27 & 29 & 20 & 28 & 25 & 29 & 16 & 14 & 16 & 16 & 10 & 12 & 9 & 7 & 28 & 26 & 20 & 28 & 31 & 27 & 27 & 25 & 24 & 24 & 24 & 26 & 17 & 24 & 8 & 9 & 9 & 9 & 8 & 9 & 9 \\
            DISF \cite{belem2020superpixel} & 16 & 12 & 11 & 11 & 1 & 5 & 6 & 13 & 18 & 14 & 11 & 1 & 5 & 8 & 6 & 5 & 3 & 8 & 19 & 4 & 9 & 29 & 29 & 29 & 28 & 29 & 27 & 29 & 4 & 5 & 5 & 7 & 5 & 6 & 5 \\
            ERS \cite{liu2011} & 7 & 10 & 8 & 7 & 7 & 16 & 2 & 20 & 25 & 25 & 8 & 10 & 3 & 14 & 19 & 20 & 21 & 21 & 16 & 20 & 20 & 22 & 21 & 23 & 21 & 24 & 7 & 22 & 6 & 7 & 7 & 5 & 7 & 4 & 6 \\
            ETPS \cite{yao2015} & 15 & 12 & 17 & 18 & 28 & 11 & 22 & 11 & 10 & 12 & 12 & 27 & 17 & 5 & 1 & 1 & 1 & 1 & 9 & 2 & 1 & 6 & 10 & 15 & 9 & 5 & 15 & 6 & 17 & 19 & 20 & 18 & 21 & 17 & 18 \\
            FUZZYSLIC \cite{wu2020fuzzy} & 13 & 14 & 14 & 13 & 24 & 9 & 20 & 23 & 15 & 23 & 23 & 25 & 13 & 18 & 16 & 13 & 15 & 13 & 8 & 17 & 16 & 3 & 4 & 3 & 4 & 2 & 11 & 3 & 26 & 24 & 25 & 23 & 18 & 27 & 26 \\
            GMMSP \cite{Ban18} & 8 & 9 & 10 & 10 & 14 & 8 & 8 & 18 & 17 & 17 & 20 & 18 & 21 & 8 & 8 & 3 & 5 & 10 & 10 & 11 & 7 & 16 & 15 & 9 & 18 & 13 & 23 & 16 & 20 & 19 & 16 & 19 & 16 & 19 & 21 \\
            GRID & 22 & 22 & 20 & 24 & 26 & 25 & 23 & 29 & 30 & 31 & 29 & 31 & 31 & 31 & 30 & 28 & 30 & 28 & 30 & 31 & 29 & 1 & 1 & 1 & 1 & 1 & 1 & 1 & 14 & 16 & 19 & 14 & 13 & 21 & 15 \\
            HHTS \cite{chang2024hierarchical} & 31 & 30 & 31 & 31 & 30 & 23 & 31 & 28 & 29 & 28 & 15 & 22 & 18 & 20 & 23 & 8 & 11 & 19 & 2 & 9 & 23 & 31 & 31 & 30 & 31 & 31 & 31 & 31 & 3 & 3 & 3 & 2 & 3 & 7 & 3 \\
            LNSNET \cite{zhu2021learning} & 30 & 28 & 27 & 30 & 23 & 27 & 19 & 30 & 26 & 19 & 31 & 22 & 30 & 30 & 31 & 24 & 24 & 31 & 27 & 26 & 24 & 24 & 23 & 21 & 26 & 15 & 22 & 23 & 27 & 28 & 28 & 29 & 29 & 29 & 29 \\
            LSC \cite{chen2017} & 14 & 15 & 13 & 16 & 7 & 4 & 11 & 16 & 24 & 15 & 29 & 15 & 15 & 10 & 2 & 5 & 8 & 7 & 6 & 3 & 2 & 12 & 13 & 11 & 15 & 10 & 21 & 12 & 13 & 14 & 13 & 19 & 14 & 16 & 17 \\
            NNSC \cite{giraud2019texture} & 12 & 17 & 23 & 15 & 22 & 5 & 14 & 17 & 19 & 28 & 19 & 30 & 3 & 17 & 10 & 17 & 23 & 9 & 12 & 10 & 13 & 8 & 7 & 8 & 6 & 9 & 12 & 6 & 23 & 25 & 30 & 28 & 26 & 20 & 23 \\
            RISF \cite{galvao2020image} & 21 & 24 & 22 & 20 & 13 & 13 & 17 & 23 & 20 & 20 & 21 & 6 & 8 & 16 & 3 & 2 & 2 & 6 & 5 & 6 & 3 & 20 & 22 & 20 & 24 & 19 & 25 & 21 & 5 & 5 & 6 & 5 & 6 & 4 & 7 \\
            RSS \cite{chai2020rooted} & 26 & 25 & 28 & 24 & 12 & 19 & 29 & 21 & 22 & 22 & 27 & 5 & 23 & 26 & 5 & 7 & 6 & 4 & 1 & 14 & 7 & 27 & 26 & 24 & 26 & 23 & 29 & 27 & 15 & 13 & 14 & 13 & 15 & 13 & 14 \\
            SAMO \cite{giraud2024} & 6 & 8 & 9 & 8 & 27 & 2 & 9 & 3 & 6 & 11 & 6 & 22 & 2 & 13 & 13 & 16 & 18 & 22 & 21 & 18 & 15 & 8 & 6 & 7 & 7 & 8 & 13 & 9 & 11 & 11 & 11 & 12 & 11 & 11 & 12 \\
            SCALP \cite{giraud2018_scalp} & 11 & 11 & 12 & 13 & 10 & 14 & 13 & 10 & 8 & 13 & 14 & 19 & 7 & 6 & 7 & 10 & 12 & 4 & 13 & 7 & 6 & 4 & 5 & 4 & 5 & 3 & 9 & 5 & 15 & 15 & 15 & 16 & 22 & 15 & 16 \\
            SEAL \cite{tu2018learning} & 4 & 3 & 3 & 1 & 31 & 27 & 1 & 25 & 23 & 26 & 17 & 29 & 18 & 28 & 20 & 22 & 22 & 19 & 16 & 24 & 18 & 26 & 25 & 27 & 22 & 27 & 7 & 26 & 7 & 8 & 8 & 8 & 10 & 8 & 8 \\
            SEEDS \cite{vandenbergh2012} & 24 & 21 & 30 & 12 & 17 & 18 & 18 & 31 & 31 & 27 & 16 & 14 & 13 & 22 & 21 & 23 & 28 & 2 & 4 & 5 & 3 & 21 & 20 & 19 & 19 & 18 & 20 & 20 & 30 & 27 & 26 & 23 & 17 & 14 & 13 \\
            SFCN \cite{yang2020} & 5 & 4 & 4 & 4 & 6 & 22 & 9 & 6 & 5 & 8 & 9 & 17 & 25 & 29 & 26 & 28 & 26 & 25 & 24 & 22 & 27 & 11 & 9 & 10 & 10 & 16 & 3 & 10 & 22 & 23 & 23 & 23 & 27 & 23 & 26 \\
            SH \cite{wei2018sh} & 18 & 16 & 15 & 17 & 5 & 11 & 7 & 22 & 28 & 23 & 24 & 13 & 10 & 23 & 15 & 9 & 7 & 14 & 18 & 11 & 10 & 28 & 28 & 28 & 30 & 30 & 24 & 28 & 1 & 1 & 2 & 1 & 1 & 1 & 1 \\
            SICLE \cite{belem2022fast} & 28 & 20 & 18 & 18 & 3 & 10 & 14 & 12 & 12 & 6 & 26 & 6 & 6 & 3 & 10 & 10 & 9 & 10 & 10 & 8 & 11 & 30 & 30 & 31 & 29 & 28 & 30 & 30 & 10 & 4 & 4 & 4 & 4 & 3 & 4 \\
            SIT-HSS \cite{xie2025hierarchical} & 22 & 18 & 25 & 27 & 7 & 17 & 26 & 25 & 7 & 1 & 18 & 3 & 12 & 20 & 21 & 15 & 14 & 23 & 26 & 23 & 21 & 15 & 17 & 22 & 14 & 22 & 6 & 15 & 1 & 1 & 1 & 3 & 1 & 1 & 1 \\
            SLIC \cite{achanta2012} & 16 & 19 & 16 & 23 & 21 & 7 & 23 & 15 & 14 & 18 & 25 & 20 & 15 & 11 & 12 & 13 & 13 & 15 & 19 & 15 & 14 & 7 & 8 & 6 & 8 & 6 & 16 & 8 & 18 & 17 & 17 & 21 & 19 & 18 & 20 \\
            SNIC \cite{achanta2017superpixels} & 20 & 25 & 23 & 20 & 10 & 14 & 21 & 19 & 21 & 20 & 22 & 9 & 11 & 11 & 9 & 12 & 9 & 12 & 3 & 15 & 12 & 10 & 13 & 14 & 12 & 11 & 18 & 13 & 12 & 12 & 12 & 11 & 11 & 10 & 10 \\
            SPAM \cite{walther2025spam} & 1 & 2 & 1 & 3 & 29 & 1 & 4 & 1 & 1 & 3 & 1 & 28 & 1 & 3 & 18 & 17 & 19 & 18 & 13 & 13 & 18 & 5 & 3 & 4 & 3 & 7 & 9 & 4 & 9 & 10 & 10 & 10 & 9 & 11 & 10 \\
            SSN \cite{jampani2018} & 10 & 7 & 6 & 4 & 4 & 2 & 3 & 7 & 4 & 10 & 2 & 16 & 24 & 15 & 17 & 19 & 15 & 15 & 13 & 19 & 17 & 19 & 18 & 16 & 16 & 21 & 19 & 17 & 19 & 18 & 18 & 15 & 19 & 22 & 18 \\
            TP \cite{levinshtein2009turbopixels} & 25 & 27 & 26 & 22 & 16 & 24 & 23 & 27 & 26 & 28 & 28 & 26 & 27 & 27 & 27 & 25 & 27 & 26 & 22 & 29 & 26 & 2 & 2 & 2 & 2 & 4 & 2 & 2 & 31 & 31 & 31 & 30 & 30 & 31 & 30 \\
            VSSS \cite{zhou2023vine} & 28 & 31 & 28 & 29 & 15 & 30 & 30 & 8 & 11 & 3 & 13 & 2 & 29 & 1 & 13 & 21 & 17 & 15 & 25 & 21 & 21 & 23 & 26 & 26 & 23 & 17 & 26 & 25 & 27 & 28 & 29 & 27 & 28 & 28 & 28 \\
            \bottomrule
        \end{tabular}%
    }
\end{table*}

\end{document}